\documentclass[10pt,journal,compsoc]{IEEEtran}
\ifCLASSOPTIONcompsoc
  \usepackage[nocompress]{cite}
\else
  \usepackage{cite}
\fi
\usepackage{amsmath}
\usepackage{amssymb}
\usepackage{multirow}
\usepackage{booktabs}
\usepackage{caption}
\usepackage{bbding}
\usepackage{bm}
\usepackage[ruled]{algorithm2e}

\newtheorem{theorem}{\textbf{Theorem}}

\newtheorem{definition}{\textbf{Definition}}

\usepackage[detect-all]{siunitx}

\usepackage{tikz,xcolor,hyperref}
\usepackage[figure]{hypcap}
\definecolor{lime}{HTML}{A6CE39}
\DeclareRobustCommand{\orcidicon}{%
    \begin{tikzpicture}
    \draw[lime, fill=lime] (0,0)
    circle [radius=0.16]
    node[white] {{\fontfamily{qag}\selectfont \tiny ID}};    \draw[white, fill=white] (-0.0625,0.095)
    circle [radius=0.007];    \end{tikzpicture}
    \hspace{-2mm}}
\foreach \x in {A, ..., Z}{%
    \expandafter\xdef\csname orcid\x\endcsname{\noexpand\href{https://orcid.org/\csname orcidauthor\x\endcsname}{\noexpand\orcidicon}}
    }

\hypersetup{hidelinks}

\ifCLASSINFOpdf
\else
\fi
\begin{document}

%

\title{Self-Supervised Node Representation Learning via Node-to-Neighbourhood Alignment}
%
%
%
%

\author{Wei Dong\orcidA{},
        Dawei Yan\orcidC{},
        and Peng Wang\orcidB{}
\IEEEcompsocitemizethanks{\IEEEcompsocthanksitem Wei Dong~(E-mail: \href{DongWei156@outlook.com}{DongWei156@outlook.com}) and Dawei Yan~(E-mail: \href{ydw753@xauat.edu.cn}{ydw753@xauat.edu.cn}) were with the College of Information and Control Engineering, Xi'an University of Architecture and Technology, China. \protect
\IEEEcompsocthanksitem Peng Wang~(E-mail: \href{pengw@uow.edu.au}{pengw@uow.edu.au}) is with the School of Computing and Information Technology, University of Wollongong, Wollongong, NSW 2522, Australia.

\IEEEcompsocthanksitem Peng Wang is the corresponding author.

}
}

%
%

\markboth{Journal of \LaTeX\ Class Files,~Vol.~14, No.~8, August~2015}%
{Shell \MakeLowercase{\textit{et al.}}: Bare Demo of IEEEtran.cls for Computer Society Journals}
%



\IEEEtitleabstractindextext{%
\begin{abstract}
    Self-supervised node representation learning aims to learn node representations from unlabelled graphs that rival the supervised counterparts. The key towards learning informative node representations lies in how to effectively gain contextual information from the graph structure. In this work, we present simple-yet-effective self-supervised node representation learning via aligning the hidden representations of nodes and their neighbourhood. Our first idea achieves such node-to-neighbourhood alignment by directly maximizing the mutual information between their representations, which, we prove theoretically, plays the role of graph smoothing. Our framework is optimized via a surrogate contrastive loss and a Topology-Aware Positive Sampling (TAPS) strategy is proposed to sample positives by considering the structural dependencies between nodes, which enables offline positive selection. Considering the excessive memory overheads of contrastive learning, we further propose a negative-free solution, where the main contribution is a Graph Signal Decorrelation (GSD) constraint to avoid representation collapse and over-smoothing. The GSD constraint unifies some of the existing constraints and can be used to derive new implementations to combat representation collapse. By applying our methods on top of simple MLP-based node representation encoders, we learn node representations that achieve promising node classification performance on a set of graph-structured datasets from small- to large-scale.

\end{abstract}

}

\maketitle

\IEEEdisplaynontitleabstractindextext

%
\IEEEpeerreviewmaketitle

\IEEEraisesectionheading{\section{Introduction}\label{sec:introduction}}


\IEEEPARstart{G}{raph}-structured data is ubiquitous. As an effective graph modelling tool, Graph Neural Networks (GNNs) have gained increasing popularity in a wide range of domains such as computer vision~\cite{zhao2019multi}, natural language processing~\cite{wu2021graph}, knowledge representation~\cite{cui2018survey}, social networks~\cite{li2020graph}, and molecular property prediction~\cite{wieder2020compact}, just name a few.

In this work, we focus on the node classification task in graphs where the key is to learn informative structure-aware node representations by gaining contextual information from the graph topology. This motivates a massive proliferation of message passing techniques in GNNs. 
Among these methods, a dominant idea follows an AGGREGATION-COMBINE-UPDATE-PREDICTION pipeline, where the AGGREGATION step aggregates the neighbouring information into vectorized representations via various neighbourhood aggregators such as \emph{mean}~\cite{hamilton2017inductive}, \emph{max}~\cite{hamilton2017inductive}, \emph{attention}~\cite{velickovic2019graph}, and \emph{ensemble}~\cite{corso2020principal}, which are COMBINED with the node representations via sum or concatenation to realize neighbourhood information fusion. To model multi-hop message passing, the AGGREGATION and COMBINE operations tend to be repeated before the ultimate node representations are obtained for label prediction, wherein the UPDATE step iteratively updates each node representation by using parameterized networks. In other words, the information exchange in this pipeline is driven by a node classification loss in the PREDICTION phase.
Under the umbrella of supervised learning, this line of methods unifies node representation learning and classification. A potential problem is that they may suffer from the scalability issue due to the expensive labelling cost for large-scale graph data.

As a remedy to expensive human annotation, Self-Supervised Learning~(SSL) has achieved proven success in various domains such as computer vision~\cite{multiview,chen2020simple} and natural language processing~\cite{devlin2019bert}, with the aim of learning meaningful representations from unlabelled data. In SSL, Graph-based Self-Supervised Learning~(GSSL) is a relatively new task that remains nascent, where the aim is to learn graph or node representations without human annotation. The key challenge thereof lies in how to design suitable \textit{pretext} task from non-Euclidean graph-structured data to learn informative node representations. Inheriting the idea of learning augmentation-invariant representations in computer vision, many recent attempts approach GSSL by designing heuristic graph augmentations based on the characteristics of the graph data to generate multi-view graphs for contrastive loss, \emph{a.k.a.} Graph Contrastive Learning~(GCL), assuming the graph or node representations should stay consistent under minor graph distortions such as dropping nodes, edge perturbation, and attribute masking~\cite{liu2022graph}. However, given the complex topological structure of the graph data, it remains unclear what is an ideal graph augmentation.


In this work, we propose an augmentation-free self-supervised node representation scheme via aligning the hidden representations of nodes and their neighbourhood. Intuitively, neighbouring nodes tend to share similar information and by aligning the node representation to the neighbourhood representation, we can distill useful contextual information from the surrounding, analogous to knowledge distillation~\cite{park2019relational,cho2019efficacy}.  Our first idea achieves such node-to-neighbourhood alignment by directly maximizing the mutual information between the hidden representations of nodes and their neighbourhood. Theoretically, the proposed Node-to-Neighbourhood (N2N) mutual information maximization essentially encourages graph smoothing based on a quantifiable graph smoothness metric. Following InfoNCE~\cite{oord2018representation}, the mutual information can be optimized by a surrogate contrastive loss, where the key boils down to positive sample definition and selection. To improve the efficiency and scalability of our N2N network as well as ensuring the quality of selected positives, we propose a Topology-Aware Positive Sampling (TAPS) strategy, which samples positives for a node from the neighbourhood by considering the structural dependencies between nodes. This enables offline positive selection.



The aforementioned N2N mutual information maximization strategy is formulated as a contrastive learning framework, which is notorious for the excessive memory overheads resulted from large negative size. This prohibits the employment of GCL to large-scale graph data especially under limited memory budgets. To this end, we propose a negative-free solution by directly maximizing the similarity between the representations of nodes and their neighbourhood, \emph{a.k.a} Negative-Free N2N (NF-N2N). However, directly pulling neighbouring nodes together may lead to a trivial solution and consequently meaningless node representations, \emph{i.e.}, all representations collapsing to constant. To avoid such representation collapse, we propose a new Graph Signal Decorrelation (GSD) constraint, which is inspired from the connection between maximizing the similarity between neighbouring representations and Graph Total Variation (GTV) minimization. GSD can be used as a general principle to explain some of the existing anti-collapse constraints such as BARLOW TWINS~\cite{zbontar2021barlow} in the context of GSSL and derive new solutions to combat representation collapse.

Different from most of the existing GSSL variants that adopt off-the-shelf GNNs~\cite{kipf2017semi,xu2019how} as node encoders, we apply our two versions of N2N loss functions on top of MLP encoders. Experiments on both attributed graph datasets and pure-structured graph datasets~\cite{yue2022survey} of varying scales show that we can achieve promising node classification performance. Our contributions are four-folds:
\begin{itemize}
    \item We propose a simple-yet-effective self-supervised node representation framework via aligning the representations of nodes and their neighbourhood. The framework is augmentation-free and is based on simple MLP node encoders.
    \item We achieve the node-to-neighbourhood (N2N) alignment by maximizing the mutual information between the hidden representations of nodes and their neighbourhood and we also present its connection to graph smoothing. A Topology-Aware Positive Sampling (TAPS) is proposed to enable efficient and quality positive sampling.
    \item A negative-free node-to-neighbourhood (NF-N2N) alignment is proposed to avoid negative sampling in N2N and thus significantly alleviates the memory overheads. The main contribution is a newly proposed Graph Signal Decorrelation (GSD) constraint which unifies some of the existing constraints to combat representation collapse in negative-free similarity maximization learning and can also be used as a principle to inspire new implementations.
    \item We conduct comprehensive experiments on different types of graph datasets with varying sizes. The results demonstrate the effectiveness of the proposed methods in node representation learning.
\end{itemize}

Note that the preliminary version of our N2N model was published in~\cite{dong2022node}. In this paper, we have made significant extensions. Firstly, we propose the GSD-based NF-N2N, which enables negative-free node-to-neighbourhood alignment and significantly reduces the memory overheads of the proposed framework. The GSD constraint can be used as a general principle to explain or derive other solutions used to combat representation collapse in negative-free similarity maximization learning. Secondly, we conducted new experiments on larger-scale pure-structured graph datasets to evaluate the effectiveness of our methods while in~\cite{dong2022node} experiments were only conducted on smaller-scale attributed graph datasets.

The rest of the paper is organized as follows. Section~\ref{sec:related} briefly presents related work. Section~\ref{sec:n2n-methodology} and~\ref{sec:nf-n2n-methodology} introduces our N2N mutual information maximization and NF-N2N framework respectively. In Section~\ref{sec:mlp-mased}, the training strategies are given. Experimental results are presented in Section~\ref{sec:experiments}, which is followed by the conclusion in Section~\ref{sec:conclusions}. Our codes are available at \url{https://github.com/dongwei156/n2n}.


\section{Related Works}\label{sec:related}

In this section, we provide brief overview of existing works on (1)~Graph Neural Networks (GNNs), (2)~Graph Contrastive Learning (GCL), (3)~MLP-based encoder in GCL, and (4)~strategies to avoid representation collapse in similarity-maximization driven learning in SSL or GSSL.

\textbf{Graph Neural Networks.}~GNNs target to learn structure-aware node/graph representations based on the topology structure of the graph data~\cite{wu2019comprehensive}. A main focus thereof is to design effective message passing strategies to encourage information propagation in the graph. Early attempts~\cite{scarselli2008graph} learn node representations by following a recurrent fashion, where the node states are updated by applying a propagation function iteratively until equilibrium is reached. Inspired by convolutional neural networks that were proposed originally for grid-like topology, such as images, convolution-like propagation was introduced into graph data~\cite{defferrard2016convolutional}. As a prevailing variant of this line of GNN works, Graph Convolutional Network (GCN)~\cite{kipf2017semi} stacks a set of $1$-hop spectral filters~\cite{defferrard2016convolutional} and nonlinear activation function to learn node representations. GCN ignites a wave of following-up work with the aim of improving the efficiency or effectiveness of information exchange in graphs. Simplifying Graph Convolutional Network (SGC)~\cite{wu2019simplifying} reduces the excessive complexity of GCN by removing the nonlinear activation function to obtain collapsed aggregation matrix, in which the expensive operation of aggregation is applied upfront before training phase. GraphSAGE-Mean~\cite{hamilton2017inductive} applies \emph{mean} aggregator to fixed number of randomly sampled neighbours to reduce the computational cost and adopts concatenation to merge the node and neighbourhood information. Another line of work aims to design sophisticated neighbourhood aggregation strategies. Graph Attention Network (GAT)~\cite{velickovic2019graph} stacks masked self-attention layers in order that the nodes can adaptively attend their neighbours. In the work~\cite{hou2019measuring}, CS-GNN is designed to understand and improve the use of graph information in GNNs by leveraging feature and label based smoothness metrics. The aforementioned GNNs are under the umbrella of supervised learning, showing proven success in various applications by virtue of sufficient labeled data.

\textbf{Graph Contrastive Learning.}~~As a prevailing paradigm in GSSL, GCL methods are usually developed based on the concept of Mutual Information (MI) maximization which works by encouraging the similarity of the positive views derived from nodes~\cite{peng2020graph}, sub-graphs~\cite{peng2020graph}, or graphs~\cite{velivckovic2018deep}. In this line of work, the key boils down to defining the positive pairs. Inspired by Deep InfoMax~\cite{hjelm2018learning} in which the image representation is learned by maximizing the mutual information between the learned embeddings and the input image, several variants are derived from applying this mutual information maximization idea in graphs. Deep Graph InfoMax (DGI)~\cite{velivckovic2018deep} maximizes the agreement between the global graph representation and its hidden node representations by pushing away the node representations derived from a corrupted graph. Graphical Mutual Information (GMI)~\cite{peng2020graph} aligns the input sub-graph to the output node representation to avoid the corrupted operation in~\cite{velivckovic2018deep}. Multi-View Graph Representation Learning (MVGRL)~\cite{hassani2020contrastive} maximizes mutual information between node representations of one view and graph representations of another graph diffusion view to learn node and graph representation. InfoGraph~\cite{sun2020infograph} views graph and patch representations as pairs and determine whether they are from the same graph. Graph Contrastive learning with Adaptive (GCA) augmentation~\cite{zhu2021graph} considers important topology and attribute information by using adaptive augmentation. GRACE~\cite{zhu2020deep} learns node representations by maximizing the agreement of node representations from two corrupted graph views generated on both structure and attribute levels. Augmentation-Free Graph Contrastive Learning (AF-GCL)~\cite{wang2022augmentation} avoids graph augmentation which is proven to perturb the middle and high-frequency components of the graph and hinder GCL application on heterophilic graphs.
These GCL methods need careful treatment of negative pairs by either relying on large batch sizes, memory banks, or customized mining strategies to retrieve the negative pairs, since such cumbersome-yet-effective negative pairs can prevent the representation collapse.

\textbf{MLP-based Encoder in GCL.}~~Prevailing GCL methods usually employ GNN-based encoders to capture the topology-aware information, but the encoding overheads in the training process are exponentially increasing with the number of hidden layers due to message passing. Graph-MLP~\cite{hu2021graph} pioneers the use of the MLP-based encoder in GCL, which postpones the expensive neighbourhood aggregation to the contrastive objective function. Simple Contrastive Graph Clustering (SCGC)~\cite{liu2022simple} treats the information aggregation as an independent pre-processing step, which is followed by MLP-based encoder.
However, these methods still preserve the information aggregation operation either before~\cite{hu2021graph} or after~\cite{liu2022simple} the node encoding.

\textbf{Strategies to Avoid Representation Collapse.}~~Learning augmentation-invariant representations by maximizing the similarities between two views derived from different augmentations is a predominant paradigm in SSL. Considering directly maximizing similarity between representations can lead to trivial solutions, \emph{i.e.}, representations collapsing to constant vectors, contrastive learning is normally employed by contrasting the similarity of positive pairs against those from negative pairs. Recently, there is a trend in computer vision to develop SSL methods that are free of negative samples. Two major lines of methods toward this goal include asymmetric learning and explicit constraint. Asymmetric learning strategies work by introducing asymmetry into the structure design or optimization of the two networks responsible for anchor sample and its positive view. Popular solutions include prediction head~\cite{grill2020bootstrap}, momentum encoder~\cite{he2020momentum} or stop-gradient~\cite{chen2021exploring}. Another line of methods avoid trivial solution in similarity maximization through designing explicit regularizations such as the redundancy reduction mechanism in ~\cite{zbontar2021barlow} and variance-invariance-convariance regularization in~\cite{bardes2022vicreg}. Existing work in GSSL mainly follows the line of asymmetric learning to prevent node or graph representations from being collapsed to constant vectors. Bootstrapped Graph Latent (BGRL)~\cite{thakoor2021bootstrapped} designs an asymmetric architecture by maintaining two distinct graph encoders and learning node representations from an online encoder to predict those from a target encoder. Self-supervised Graph Neural Networks (SelfGNN)~\cite{kefato2021self} leverages Batch Normalization as implicit contrastive terms. AF-GRL~\cite{lee2021augmentation} avoids negative samples by directly adopting BYOL~\cite{grill2020bootstrap} as the backbone.

\section{Node-to-Neighbourhood Alignment via Mutual Information Maximization}\label{sec:n2n-methodology}

In this section, we firstly introduce notations, symbols, and necessary background about GNN models and GCL schemes. We then present our idea of N2N mutual information maximization and its link to graph smoothing. Finally, we elaborate the proposed TAPS strategy.

\subsection{Preliminary Knowledge on GNN and GCL}\label{sec:preliminary-gnn-gcl}

We denote a graph $\mathcal{G}=(\mathcal{V}, \mathcal{E}, \mathbf{A}, \mathbf{X})$, which is composed of a set of nodes $\mathcal{V}$ with the number of nodes $N=|\mathcal{V}|$, a set of edges $\mathcal{E}$, an adjacency matrix $\mathbf{A} \in \mathbb{R}^{N \times N}$, and node feature matrix $\mathbf{X} \in \mathbb{R}^{N \times D}$. Each node $v_{i} \in \mathcal{V}$ has a feature vector $\vec{\boldsymbol x}_{i} \in \mathbb{R}^{D}$ and all these node feature vectors form the aforementioned node feature matrix $\mathbf{X}=[\vec{\boldsymbol x}_{1}, \cdots, \vec{\boldsymbol x}_{N}]^{{\rm T}}$. From the perspective of graph signals~\cite{shuman2013emerging}, the node feature matrix can also be perceived as $\mathbf{X}=[\vec{\boldsymbol s}_{1}, \cdots, \vec{\boldsymbol s}_{D}]$, with each $N$-dimensional column vector as a graph signal $\vec{\boldsymbol s}_{d} \in \mathbb{R}^{N}$.
In the supervised learning task for the graph $\mathcal{G}$, GNNs utilize a neighbourhood aggregation scheme to learn latent node embedding $\vec{\boldsymbol h}^{(l)} \in \mathbb{R}^{D^{(l)}}$ in $(l)$-th layer for each node $v$, and a prediction function is applied to the node representations of final hidden layer to predict the class label $y_{v}$ of each node $v$.

Based on such notations, a commonly adopted ${\rm AGGREGATION}$-${\rm COMBINE}$-${\rm UPDATE}$-${\rm PREDICTION}$ pipeline for supervised GNNs can be defined as:

\begin{equation}\label{eq:pipeline}
\begin{split}
    \vec{\boldsymbol a}_{i}^{(l-1)} &= {\rm AGGREGATION}(\{\vec{\boldsymbol h}_{j}^{(l-1)}:v_{j}\in \mathcal{N}_{i}\}), \\
    \vec{\boldsymbol c}_{i}^{(l)} &= {\rm COMBINE}(\{\vec{\boldsymbol a}_{i}^{(l-1)}, \vec{\boldsymbol h}_{i}^{(l-1)}\}), \\
    \vec{\boldsymbol h}_{i}^{(l)} &= {\rm UPDATE}(\vec{\boldsymbol c}_{i}^{(l)}), \\
    \mathcal{L}_{{\rm CE}} &= {\rm PREDICTION}(\{\vec{\boldsymbol h}_{i}^{(L)}, y_{v_{i}}\}), \\
\end{split}
\end{equation}

\noindent where the ${\rm AGGREGATION}$ function can be any form of aggregators such as \emph{mean}~\cite{hamilton2017inductive}, \emph{max}~\cite{hamilton2017inductive}, \emph{sum}~\cite{xu2019how}, \emph{attention}~\cite{velickovic2019graph}, and \emph{ensemble}~\cite{corso2020principal} that learn the neighbourhood representation $\vec{\boldsymbol a}_{i}^{(l-1)}$ from the set $\{\vec{\boldsymbol h}_{j}^{(l-1)}:v_{j}\in \mathcal{N}_{i}\}$ based on the neighbourhood $\mathcal{N}_{i}$ and the neighbouring node embedding $\vec{\boldsymbol h}_{j}^{(l-1)}$ with $(l-1)$-th layer, and the ${\rm COMBINE}$ function updates $\vec{\boldsymbol h}_{i}^{(l-1)}$ to a new representation $\vec{\boldsymbol c}_{i}^{(l)}$ in $(l)$-th layer by combining $\vec{\boldsymbol a}_{i}^{(l-1)}$ with $\vec{\boldsymbol h}_{i}^{(l-1)}$. The Following ${\rm UPDATE}$ function learns a new node representation $\vec{\boldsymbol h}_{i}^{(l)}$ by leveraging the learnable parameters $\theta$ onto $\vec{\boldsymbol c}_{i}^{(l)}$. A $L$-layer GNN iterates the above three operations $L$ times and the ${\rm PREDICTION}$ function is applied in the output layer for node classification. A de-facto loss function for the ${\rm PREDICTION}$ layer is Cross-Entropy (CE) loss.

Mutual Information (MI) maximization is employed as a de-facto objective function to develop graph contrastive learning~\cite{velivckovic2018deep}, where the MI between representations of nodes~\cite{peng2020graph}, subgraphs~\cite{peng2020graph}, or graphs~\cite{velivckovic2018deep} under different graph augmentations is maximized. A typical node-based contrastive learning framework is formulated as:

\begin{equation}\label{eq:node-scale-contrastive-learning}
    \theta^{\ast} = \mathop{\arg\max}\limits_{\theta} I\left(f_{\theta}\left(\mathcal{T}\left(\vec{\boldsymbol x}_{i}, \mathcal{G}\right)\right); f_{\theta}\left(\mathcal{T}^{'}\left(\vec{\boldsymbol x}_{i}, \mathcal{G}\right)\right)\right), \\
\end{equation}

\noindent where the $f_{\theta}(\cdot)$ denotes node encoder with learnable parameters $\theta$; $\mathcal{T}(\cdot, \cdot)$ and $\mathcal{T}^{'}(\cdot, \cdot)$ represent different augmentation strategies applied to the same node feature vector $\vec{\boldsymbol x}_{i}$ relying on the topological context of graph $\mathcal{G}$.


\subsection{N2N Mutual Information Maximization}\label{sec:n2n}

We learn topology-aware node representation by maximizing the mutual information between the hidden representations of nodes and their neighbourhood, which is partially motivated by knowledge distillation~\cite{park2019relational}. In this section, we firstly present the definition of the N2N mutual information, which is followed by the optimization of the mutual information and its link to graph smoothing.

We denote the Probability Density Function (PDF) of the node representation $\vec{\boldsymbol h}_{i}^{(l)}$ over the feature space $\mathcal{X}^{D^{(l)}}$ in $[0,1]^{D^{(l)}}$ as $p(H(\bm{x})^{(l)})$ , where $\bm{x}\in \mathcal{X}^{D^{(l)}}$ and $H(\cdot)^{(l)}$ is a mapping function from $\bm{x}$ to $\vec{\boldsymbol h}_{i}^{(l)}$; the PDF of the neighbourhood representation $\vec{\boldsymbol a}_{i}^{(l)}$ as $p(S(\bm{x})^{(l)})$ with the mapping function $S(\cdot)^{(l)}$ from $\bm{x}$ to $\vec{\boldsymbol a}_{i}^{(l)}$; and the joint PDF between node and neighbourhood is $p(S(\bm{x})^{(l)}, H(\bm{x})^{(l)})$. We define the mutual information between the node representations and their corresponding neighbourhood representation as:

\begin{equation}\label{eq:mutual-information}
\begin{split}
    &I(S(\bm{x})^{(l)};H(\bm{x})^{(l)})= \\
    &\underset{\mathcal{X}^{D^{(l)}}}{\int}p(S(\bm{x})^{(l)},H(\bm{x})^{(l)})\cdot\log\frac{p(S(\bm{x})^{(l)},H(\bm{x})^{(l)})}{p(S(\bm{x})^{(l)})\cdot p(H(\bm{x})^{(l)})}d\bm{x}.
\end{split}
\end{equation}

This operation encourages each node representation to distill the contextual information presented in its neighbourhood representation. However, mutual information is notoriously difficult to compute, particularly in continuous and high-dimensional space. Fortunately, scalable estimation enabling efficient computation of mutual information is made possible through Mutual Information Neural Estimation (MINE)~\cite{belghazi2018mutual}, which converts mutual information maximization into minimizing the InfoNCE loss~\cite{oord2018representation}. The surrogate InfoNCE loss function of the N2N mutual information in Eq.~(\ref{eq:mutual-information}) is defined as:

\begin{equation}\label{eq:infonce}
\begin{split}
    &\mathcal{L}_{{\rm InfoNCE}}=\\
    &-\mathbb{E}_{v_{i}\in \mathcal{V}}\left[\log\frac{\exp({\rm sim}(\vec{\boldsymbol a}_{i}^{(l)}, \vec{\boldsymbol h}_{i}^{(l)})/\tau)}{\sum_{v_{k}\in \mathcal{V}}\exp({\rm sim}(\vec{\boldsymbol h}_{k}^{(l)}, \vec{\boldsymbol h}_{i}^{(l)})/\tau)}\right], \\
\end{split}
\end{equation}

\noindent which estimates the mutual information via node sampling, where the ${\rm sim}(\cdot, \cdot)$ function denotes the cosine similarity, the ${\rm exp}(\cdot)$ function implies the exponential function, and $\tau$ is the temperature parameter. The positive pair is $(\vec{\boldsymbol a}_{i}^{(l)}, \vec{\boldsymbol h}_{i}^{(l)})$ and the negative pair is $(\vec{\boldsymbol h}_{k}^{(l)}, \vec{\boldsymbol h}_{i}^{(l)})_{i\neq k}$.

In essence, maximizing $I(S(\bm{x})^{(l)};H(\bm{x})^{(l)})$ can play the role of graph smoothing, which has proven to be able to benefit node/graph prediction. To elaborately prove this point, we resort to a feature smoothness metric in~\cite{hou2019measuring}:

\begin{equation}\label{eq:feature-smoothness}
    \delta_{f}^{(l)}=\frac{\|\sum_{v_{i}\in \mathcal{V}}(\sum_{v_{j}\in \mathcal{N}_{i}}(\vec{\boldsymbol h}_{i}^{(l)}-\vec{\boldsymbol h}_{j}^{(l)}))^{2}\|_{1}}{|\mathcal{E}|\cdot D^{(l)}},
\end{equation}

\noindent where $\|\cdot\|_{1}$ is the Manhattan norm. The work~\cite{hou2019measuring} further proposes that the information gain from the neighbourhood representation $\vec{\boldsymbol a}_{i}^{(l)}$ is defined as the Kullback-Leibler divergence:

\begin{equation}\label{eq:kl-divergence}
\begin{split}
    &D_{KL}(S(\bm{x})^{(l)}\|H(\bm{x})^{(l)})= \\ &\underset{\mathcal{X}^{D^{(l)}}}{\int}p(S(\bm{x})^{(l)})\cdot\log\frac{p(S(\bm{x})^{(l)})}{p(H(\bm{x})^{(l)})}d\bm{x}, \\
\end{split}
\end{equation}

\noindent which is positively correlated to the feature smoothness metric $\delta_{f}^{(l)}$, \textit{i.e.}, $D_{KL}(S(\bm{x})^{(l)}\|H(\bm{x})^{(l)})\sim \delta_{f}^{(l)}$. This standpoint implies that a large feature smoothness value $\delta_{f}^{(l)}$ means significant disagreement between node representations $\{\vec{\boldsymbol h}_{i}^{(l)}\}$ and their corresponding neighbourhood representations $\{\vec{\boldsymbol a}_{i}^{(l)}\}$. This inspires the following theorem (Ref. Appendix~\ref{appe:theo:kl-mut} for proof):

\begin{theorem}\label{theo:kl-mut}
    For a graph $\mathcal{G}$ with the set of features $\mathcal{X}^{D^{(l)}}$ in space $[0,1]^{D^{(l)}}$, the information gain represented by $D_{KL}(S(\bm{x})^{(l)}\|H(\bm{x})^{(l)})$ is negatively correlated to the mutual information $I(S(\bm{x})^{(l)};H(\bm{x})^{(l)})$ and thus maximizing $I(S(\bm{x})^{(l)};H(\bm{x})^{(l)})$ essentially minimizes $D_{KL}(S(\bm{x})^{(l)}\|H(\bm{x})^{(l)})$ and $\delta_{f}^{(l)}$, which attains the goal of graph smoothing:
    \vspace{-0.2cm}
    \begin{equation}\label{eq:kl-mut}
    \begin{split}
        I(S(\bm{x})^{(l)};H(\bm{x})^{(l)})&\sim \frac{1}{D_{KL}(S(\bm{x})^{(l)}\|H(\bm{x})^{(l)})} \\
        &\sim \frac{1}{\delta_{f}^{(l)}}. \\
    \end{split}
    \end{equation}
\end{theorem}

\subsection{Topology-Aware Positive Sampling (TAPS)}\label{sec:taps}

Up to now, we obtain the neighbourhood representation $\vec{\boldsymbol a}_{i}^{(l)}$ of a node by applying the ${\rm AGGREGATION}$ function to all neighbours of the node. This solution may suffer from two issues. Firstly, the whole neighbourhood can include redundant or even noisy information. Secondly, the aggregation operation is computationally expensive. To address these two problems, we propose a TAPS strategy for self-supervised node representation learning. The basic idea is that we measure the topological dependencies between a node and its neighbours and sample positives of the node based on the ranked dependency values.

For a node $v_i$, we use a variable $X_i$ to represent its topological information. $X_i$ can take the value of either $\mathcal{N}_i$ or $\overline{\mathcal{N}_i}=\mathcal{V}-\mathcal{N}_i$, where the former corresponds to the neighbourhood information and the latter is the contextual information complementary to the neighbourhood. Based on the definition of $X_i$, we define $p(X_{i}=\mathcal{N}_{i})=\frac{|\mathcal{N}_{i}|}{|\mathcal{V}|}$ and probability $p(X_{i}=\overline{\mathcal{N}_{i}})=\frac{|\mathcal{V}-\mathcal{N}_{i}|}{|\mathcal{V}|}$, where $|\cdot|$ is the cardinality function. Basically $p(X_{i}=\mathcal{N}_{i})$ indicates when we sample a node randomly on the graph, the probability that the node will fall into the neighbourhood of $v_{i}$. Furthermore, for two neighbouring nodes $v_i$ and $v_j$, we can define the following joint probabilities:

\begin{equation}\label{eq:dist}
\begin{split}
    p(X_{i}=\mathcal{N}_{i}, X_{j}=\mathcal{N}_{j})&=\frac{|\mathcal{N}_{i}\cap \mathcal{N}_{j}|}{|\mathcal{V}|}, \\
    p(X_{i}=\mathcal{N}_{i}, X_{j}=\overline{\mathcal{N}_{j}})&=\frac{|\mathcal{N}_{i}\cap (\mathcal{V}-\mathcal{N}_{j})|}{|\mathcal{V}|}, \\
    p(X_{i}=\overline{\mathcal{N}_{i}}, X_{j}=\mathcal{N}_{j})&=\frac{|(\mathcal{V}-\mathcal{N}_{i})\cap \mathcal{N}_{j}|}{|\mathcal{V}|}, \\
    p(X_{i}=\overline{\mathcal{N}_{i}}, X_{j}=\overline{\mathcal{N}_{j}})&=\frac{|(\mathcal{V}-\mathcal{N}_{i})\cap (\mathcal{V}-\mathcal{N}_{j})|}{|\mathcal{V}|},
\end{split}
\end{equation}

\noindent where $p(X_{i}=\mathcal{N}_{i}, X_{j}=\mathcal{N}_{j})$ is the probability that the randomly selected node will fall into the intersected neighbours of $v_{i}$ and $v_{j}$. Motivated by mutual information, we define the graph-structural dependency between $v_i$ and $v_j$ as:

\begin{definition}\label{definition:graph-structural-dependency}
    Graph-structural dependency between neighbouring node $v_{i}$ and $v_{j}$ is defined as:

    \begin{equation}\label{eq:tami}
    \begin{split}
        I(X_{i};X_{j})=&\sum_{X_{i}} \sum_{X_{j}} p(X_{i}, X_{j})\cdot\log\frac{p(X_{i}, X_{j})}{p(X_{i})\cdot p(X_{j})}, \\
        &s.t.\;v_{j}\in \mathcal{N}_{i}. \\
    \end{split}
    \end{equation}

\end{definition}

\noindent The graph-structural dependency value above basically measures the topological similarity of two nodes. A large value suggests the strong dependency between two nodes.

In TAPS strategy, we select positives of $v_{i}$ by ranking the dependency values between the neighbouring nodes and $v_i$ and then obtain the neighbourhood representation $\vec{\boldsymbol a}_{i}^{(l)}$ by applying aggregator, \textit{e.g.}, \emph{mean}, to the selected positives. In particular, when only one positive is selected, we directly select the node $v_j$ with maximum dependency value to $v_i$ and thus avoid the expensive aggregation operation. Meanwhile, because the topology structure of a graph relies on the adjacency matrix only, TAPS allows us to perform positive sampling upfront, which can avoid the positive sampling overhead during training phase.

\section{Negative-Free Node-to-Neighbourhood Alignment via Graph Signal Decorrelation}\label{sec:nf-n2n-methodology}

In the previous section, we learn topology-aware node representations by aligning nodes and their neighbourhood through a contrastive learning framework, i.e., pulling a node and its neighbourhood close in the embedding space while pushing the node apart from other negative nodes. However, the treatment of negative pairs in contrastive learning needs to be careful because it requires either large memory consumption to hold sufficient negative samples or customized mining strategies to retrieve effective negatives~\cite{zbontar2021barlow,grill2020bootstrap,bardes2022vicreg}. This tends to be problematic when dealing with large-scale data or under limited computational budget.


In this section, we aim to design a negative-free node-to-neighbourhood alignment strategy to learn node representations. However, directly aligning the nodes and their neighbourhood without resorting to the contrastive term can lead to a trivial solution, i.e., all nodes collapse to the same constant representations.
To address the representation collapse problem and consequently learn meaningful node representations, we inspect the collapse from the lens of graph signal processing and propose a general constraint based on Graph Signal Decorrelation (GSD).



\subsection{Self-Supervised Node Representation Learning via Graph Signal Decorrelation}\label{sec:nf-n2n-gsd}

\begin{figure*}[!t]
    \centering
    \includegraphics[scale=0.3]{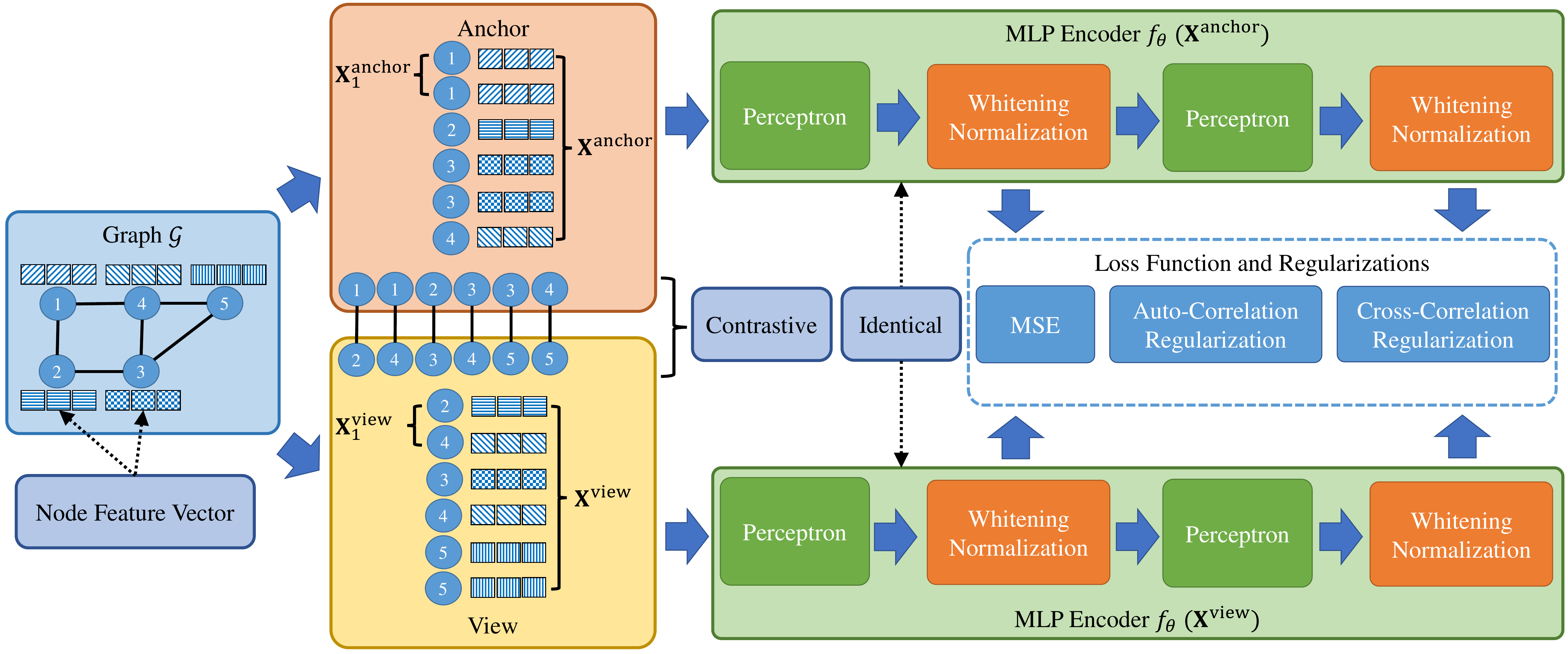}
    \caption{Negative-Free Node-to-Neighbourhood (NF-N2N) graph contrastive learning via graph signal decorrelation. We directly regard two nodes linked at one edge as a positive pair in which one node is an anchor and another is a view, \emph{i.e.}, each anchor node has multi-view nodes. We then feed these multi-view positive pairs into the MLP encoder for GCL, with three decorrelation strategies (whitening normalization, auto-correlation, and cross-correlation) to prevent over-smoothing and representation collapse.}\label{fig:MLP-based-training}
\end{figure*}

Following previous section, in graph $\mathcal{G}$, neighbouring nodes have higher chances to share similar properties such as node labels\footnote{In this work, we focus on homogeneous graph. We will investigate if our strategy applies to heterogeneous graph as future work.}. Motivated by this assumption, we treat a node $v_{i}$ as an anchor and its neighbouring node $v_{j}\in \mathcal{N}_{i}$ as an alternative view of $v_{i}$. If node $v_{i}$ has $J$ neighbours, we can directly derive the multiple views of $v_{i}$ and form a set of positive pairs as $\{(\vec{\boldsymbol x}_{i}^{{\rm anchor}}, \vec{\boldsymbol x}_{1 \in \mathcal{N}_{i}}^{{\rm view}}),\cdots,(\vec{\boldsymbol x}_{i}^{{\rm anchor}}, \vec{\boldsymbol x}_{J \in \mathcal{N}_{i}}^{{\rm view}})\}$ as shown in Fig.~\ref{fig:MLP-based-training}. The samples within each positive pair are fed into identical MLP-based encoder $f_{\theta}(\cdot)$ to derive corresponding embeddings $(\mathbf{H}_{i}^{{\rm anchor}},\mathbf{H}_{i}^{{\rm view}})$ with $\mathbf{H}_{i}^{{\rm anchor}}=f_{\theta}(\mathbf{X}_{i}^{{\rm anchor}})$ and $\mathbf{H}_{i}^{{\rm view}}=f_{\theta}(\mathbf{X}_{i}^{{\rm view}})$, where:

\begin{equation}\label{eq:embeddings}
\begin{split}
    \mathbf{X}_{i}^{{\rm anchor}}&=[\vec{\boldsymbol x}_{i}, \cdots, \vec{\boldsymbol x}_{i}, \cdots, \vec{\boldsymbol x}_{i}]^{{\rm T}}, \\
    \mathbf{X}_{i}^{{\rm view}}&=[\vec{\boldsymbol x}_{1 \in \mathcal{N}_{i}}, \cdots, \vec{\boldsymbol x}_{j \in \mathcal{N}_{i}}, \cdots, \vec{\boldsymbol x}_{J \in \mathcal{N}_{i}}]^{{\rm T}}, \\
    \mathbf{H}_{i}^{{\rm anchor}}&=[\vec{\boldsymbol h}_{i}^{(l)}, \cdots, \vec{\boldsymbol h}_{i}^{(l)}, \cdots, \vec{\boldsymbol h}_{i}^{(l)}]^{{\rm T}}, \\
    \mathbf{H}_{i}^{{\rm view}}&=[\vec{\boldsymbol h}_{1 \in \mathcal{N}_{i}}^{(l)}, \cdots, \vec{\boldsymbol h}_{j \in \mathcal{N}_{i}}^{(l)}, \cdots, \vec{\boldsymbol h}_{J \in \mathcal{N}_{i}}^{(l)}]^{{\rm T}}. \\
\end{split}
\end{equation}
\noindent All these feature and embedding matrices are mapped into the space $\mathbb{R}^{J \times D^{(l)}}$. Then our goal is to learn view-invariant node representations by maximizing the similarities between such positive pairs of embeddings. Note that the number of positive pairs from graph $\mathcal{G}$ equals to the number of edges $|\mathcal{E}|$.

Following the work~\cite{grill2020bootstrap}, we maximize the similarity between a positive pair $v_i$ and $v_{j\in \mathcal{N}_{i}}$ by minimizing the Mean Squared Error (MSE) between their embeddings $\vec{\boldsymbol h}_{i}^{(l)}$ and $\vec{\boldsymbol h}_{j}^{(l)}$ with $(l)$-th layer. Applying the MSE loss function to all the positive pairs, we get:

\begin{equation}\label{eq:MSE}
\begin{split}
    \mathcal{L}_{{\rm MSE}}^{(l)}&=\sum_{i=1}^{N}\sum_{j\in \mathcal{N}_{i}}\|\vec{\boldsymbol h}_{i}^{(l)}-\vec{\boldsymbol h}_{j}^{(l)}\|_{2}^{2} \\
    &=\sum_{i=1}^{N}\sum_{j\in \mathcal{N}_{i}}\left(2-2\cdot \vec{\boldsymbol h}_{i}^{(l)}, \vec{\boldsymbol h}_{j}^{(l)}\right), \\
\end{split}
\end{equation}


\noindent where $\vec{\boldsymbol h}_{i}^{(l)}$ is the intermediate $L_2$-norm node embedding in layer $l$ of the node encoder $f_{\theta}(\vec{\boldsymbol x}_{i})$. If we view graph as a set of graph signals, Eq.~(\ref{eq:MSE}) essentially represents Graph Total Variation (GTV)~\cite{shuman2013emerging} that can be used as an indicator of the smoothness level of the graph. The smaller the GTV value is, the smoother the graph is. GTV is defined as:

\begin{equation}\label{eq:MSE-GTV}
\begin{split}
   \sum_{i=1}^{N}&\sum_{j\in \mathcal{N}_{i}}\|\vec{\boldsymbol h}_{i}^{(l)}-\vec{\boldsymbol h}_{j}^{(l)}\|_{2}^{2} \\
   &=\sum_{e_{i,j} \in \mathcal{E}}\|\vec{\boldsymbol h}_{i}^{(l)}-\vec{\boldsymbol h}_{j}^{(l)}\|_{2}^{2} \\
   &=\sum_{e_{i,j} \in \mathcal{E}}\sum_{d^{(l)}=1}^{D^{(l)}}\left(\vec{\boldsymbol h}_{i}^{(l)}[d^{(l)}]-\vec{\boldsymbol h}_{j}^{(l)}[d^{(l)}]\right)^{2} \\
   &=\sum_{d^{(l)}=1}^{D^{(l)}}\sum_{e_{i,j} \in \mathcal{E}}\left(\vec{\boldsymbol h}_{i}^{(l)}[d^{(l)}]-\vec{\boldsymbol h}_{j}^{(l)}[d^{(l)}]\right)^{2} \\
   &=\sum_{d^{(l)}=1}^{D^{(l)}}\vec{\boldsymbol s}_{d^{(l)}}^{{\rm T}} \cdot \mathbf{L} \cdot \vec{\boldsymbol s}_{d^{(l)}}, \\
\end{split}
\end{equation}

\noindent where $\vec{\boldsymbol s}_{d^{(l)}}$ is a graph signal, $\mathbf{L}=\mathbf{D}-\mathbf{A}$ is the Graph Laplacian matrix~\cite{kipf2017semi} with the adjacency matrix $\mathbf{A}$ and the degree matrix $\mathbf{D}_{[i,i]}=\sum_{j \in \mathcal{N}_{i}}\mathbf{A}_{[i,j]}$; $\sum_{d^{(l)}=1}^{D^{(l)}}\vec{\boldsymbol s}_{d^{(l)}}^{{\rm T}} \cdot \mathbf{L} \cdot \vec{\boldsymbol s}_{d^{(l)}}$ denotes GTV.

Following Theorem~\ref{theo:kl-mut}, minimizing the MSE loss function $\mathcal{L}_{{\rm MSE}}^{(l)}$ essentially maximizes the mutual information between the neighbourhood representation $\vec{\boldsymbol a}_{i}^{(l)}$ and the embedding $\vec{\boldsymbol h}_{i}^{(l)}$ of the node $v_i$. Theorem~\ref{theo:info-MSE} theorizes the relationship between such mutual information and MSE (GTV), proved in Appendix~\ref{app:theo:info-MSE}:

\begin{theorem}\label{theo:info-MSE}
    Minimizing the MSE loss function $\mathcal{L}_{{\rm MSE}}^{(l)}$ essentially maximizes $I(S(\bm{x})^{(l)};H(\bm{x})^{(l)})$, which attains the goal of graph smoothing:

    \begin{equation}
        \mathcal{L}_{{\rm MSE}}^{(l)}({\rm GTV}) \sim \frac{1}{I(S(\bm{x})^{(l)};H(\bm{x})^{(l)})},
    \end{equation}

    when $\vec{\boldsymbol s}_{d^{(l)}}\to \hat{\sigma} \cdot \vec{\boldsymbol 1}$ with a scalar $\hat{\sigma}$.
\end{theorem}

However, under the positive semi-definite property of Graph Laplacian $\mathbf{L}$~\cite{shuman2013emerging}, directly minimizing the MSE loss in the above equations can push GTV approaching its minimum value $0$ and consequently results in over-smoothing. In other words, all the node embeddings $\{\vec{\boldsymbol h}_{i}^{(l)}\}$ will collapse to constant. From the graph signal angle, minimizing the GTV will force each graph signal $\vec{\boldsymbol s}_{d^{(l)}}$ to be a constant vector, \emph{i.e.}, $\vec{\boldsymbol s}_{d^{(l)}}=\hat{\sigma} \cdot \vec{\boldsymbol 1}$ in accordance with Theorem~\ref{theo:info-MSE}. When all $\{\vec{\boldsymbol s}_{d^{(l)}}\}$ are a vector whose elements are all $1$ with the scalar $\hat{\sigma}$, the embedding matrix $\mathbf{H}^{(l)}=[\vec{\boldsymbol s}_{1},\cdots,\vec{\boldsymbol s}_{d^{(l)}},\cdots,\vec{\boldsymbol s}_{D^{(l)}}]$ has only one non-zero eigenvalue, whose embedding space collapses to only one dimension. All equal elements of graph signals $\{\vec{\boldsymbol s}_{d^{(l)}}\}$ also cause graph over-smoothing~\cite{li2018deeper,xu2018representation}.

Furthermore, we assume Graph Laplacian matrix $\mathbf{L}$ has eigenvalues $\lambda_{1}<\lambda_{2}<\cdots<\lambda_{N}$ and their corresponding eigenvectors $[\vec{\boldsymbol v}_{1},\cdots,\vec{\boldsymbol v}_{N}]$. By virtue of the property of Graph Laplacian~\cite{shuman2013emerging}, the smallest eigenvalue $\lambda_{1}=0$ and its eigenvector $\vec{\boldsymbol v}_{1}=\frac{\vec{\boldsymbol 1}}{\|\vec{\boldsymbol 1}\|_{2}}$. From Theorem~\ref{theo:info-MSE}, since MSE minimization leads to $\vec{\boldsymbol s}_{d^{(l)}}=\hat{\sigma} \cdot \vec{\boldsymbol 1}=\hat{\sigma} \cdot \|\vec{\boldsymbol 1}\|_{2} \cdot \frac{\vec{\boldsymbol 1}}{\|\vec{\boldsymbol 1}\|_{2}}=\sigma \cdot \vec{\boldsymbol v}_{1}$ with $\sigma=\hat{\sigma} \cdot \|\vec{\boldsymbol 1}\|_{2}$, we obtain:

\begin{equation}\label{eq:eigenvalues-eigenvectors}
\begin{split}
    \vec{\boldsymbol s}_{d^{(l)}}^{{\rm T}} \cdot \mathbf{L} \cdot \vec{\boldsymbol s}_{d^{(l)}} &= \sigma \cdot \vec{\boldsymbol v}_{1}^{{\rm T}} \cdot \mathbf{L} \cdot \sigma \cdot \vec{\boldsymbol v}_{1} \\
    &= \sigma^{2} \cdot \vec{\boldsymbol v}_{1}^{{\rm T}} \cdot \mathbf{L} \cdot \vec{\boldsymbol v}_{1} \\
    &= \sigma^{2} \cdot \lambda_{1} \cdot \vec{\boldsymbol v}_{1}^{{\rm T}} \cdot \vec{\boldsymbol v}_{1} \\
    &= \sigma^{2} \cdot \lambda_{1} \\
    &=0, \\
\end{split}
\end{equation}

\noindent where $\mathbf{L} \cdot \vec{\boldsymbol v}_{i}=\lambda_{i} \cdot \vec{\boldsymbol v}_{i}$ and $\vec{\boldsymbol v}_{1}^{{\rm T}} \cdot \vec{\boldsymbol v}_{1}=1$.

To avoid such trivial solution and consequently learn meaningful node representations, we explicitly require any two graph signals being orthogonal to each other, \emph{i.e.}, $\vec{\boldsymbol s}_{i} \perp \vec{\boldsymbol s}_{j}\ s.t.\ i\neq j$ with $1 \leq i \leq D^{(l)}, 1\leq j \leq D^{(l)}$. Under this explicit orthogonality constraint, due to the semi-definite nature of $\mathbf{L}$, GTV minimization will ideally push one term $\vec{\boldsymbol s}_{d^{(l)}}^{{\rm T}} \cdot \mathbf{L} \cdot \vec{\boldsymbol s}_{d^{(l)}} $ approaching $0$, \emph{i.e.} $\vec{\boldsymbol s}_{d^{(l)}}=\hat{\sigma} \cdot \vec{\boldsymbol 1}$, but the remaining terms $\vec{\boldsymbol s}_{i}^{{\rm T}} \cdot \mathbf{L} \cdot \vec{\boldsymbol s}_{i}$ with $i\neq d^{(l)}$ will be positive. Specifically, Theorem~\ref{theo:graph-signal-decoorelation} describes such orthogonality constraint mentioned and we prove it in Appendix~\ref{app:theo:graph-signal-decoorelation}:

\begin{theorem}\label{theo:graph-signal-decoorelation}
    Given $D^{(l)}<N$, eigenvalues $0=\lambda_{1}<\lambda_{2}<\cdots<\lambda_{N}$ of Graph Laplacian matrix $\mathbf{L}$ and their corresponding eigenvectors $[\vec{\boldsymbol v}_{1},\cdots,\vec{\boldsymbol v}_{N}]$, and $\vec{\boldsymbol s}_{i} \perp \vec{\boldsymbol s}_{j}\ s.t.\ i\neq j$ with $1\leq i\leq D^{(l)}, 1 \leq j \leq D^{(l)}$. When MSE minimization makes any graph signal, for example $\vec{\boldsymbol s}_{1}$, equal to $\sigma_{1} \cdot \vec{\boldsymbol v}_{1}$ resulting in $\vec{\boldsymbol s}_{1}^{{\rm T}} \cdot \mathbf{L} \cdot \vec{\boldsymbol s}_{1}=0$ in terms of Eq.~(\ref{eq:eigenvalues-eigenvectors}), the GTV is:

    \begin{equation}\label{eq:decorrelation-GTV}
        \sum_{i=1}^{D^{(l)}}\vec{\boldsymbol s}_{i}^{{\rm T}} \cdot \mathbf{L} \cdot \vec{\boldsymbol s}_{i}=\sigma_{1}^{2} \cdot \lambda_{1}+\sigma_{2}^{2} \cdot \lambda_{2} + \cdots + \sigma_{D^{(l)}}^{2} \cdot \lambda_{D^{(l)}}>0.
    \end{equation}
\end{theorem}

\noindent Eq.~(\ref{eq:decorrelation-GTV}) prevents GTV from converging to $0$ and can thus avoid the representation collapse and over-smoothing.

\subsection{Three Constraints Derived from Graph Signal Decorrelation}\label{sec:three}

In this section, we design three constraints to avoid representation collapse based on the graph signal decorrelation principle, where the basic idea is to push the graph signals to be orthogonal to each other. Before delving into the details of the constraints, we start with the definition of auto-covariance matrix:

\begin{equation}\label{eq:auto-covariance-matrix}
\begin{split}
    \mathbf{C}_{{\rm auto}}^{(l)}=&\frac{1}{N}\left(\mathbf{H}^{(l)}-\vec{\boldsymbol 1}\cdot\left(\frac{1}{N}\vec{\boldsymbol 1}^{{\rm T}}\cdot\mathbf{H}^{(l)}\right)\right)^{{\rm T}}\cdot \\
    &\left(\mathbf{H}^{(l)}-\vec{\boldsymbol 1}\cdot\left(\frac{1}{N}\vec{\boldsymbol 1}^{{\rm T}}\cdot\mathbf{H}^{(l)}\right)\right)+\epsilon\cdot\mathbf{I}, \\
\end{split}
\end{equation}

\noindent in which off-diagonal element describes the correlation between two graph signals $\vec{\boldsymbol s}_{i}$ and $\vec{\boldsymbol s}_{j}$ with $i \neq j$, where a positive value indicates positive correlation and value $0$ means two signals are not correlated. $\epsilon\cdot\mathbf{I}$ with an identity matrix $\mathbf{I}$ and a small number $\epsilon$ prevents invalid operation in the algorithm program of $\mathbf{C}_{{\rm auto}}^{(l)}$ (Not a Number, NaN). Two strategies, whitening normalization and auto-correlation regularization, can make the off-diagonal elements in matrix $\mathbf{C}_{{\rm auto}}^{(l)}$ as close to $0$ as possible.

\textbf{Whitening Normalization.}~~Whitening normalization used in self-supervised computer vision tasks can scatter the batch samples to avoid degenerating embedding solutions collapsed onto a few dimensions or into a single point~\cite{ermolov2021whitening}, with more feature or embedding decorrelation effect than Batch Normalization (BN)~\cite{ioffe2015batch}, even though BN also boosts the whitening effectiveness additionally~\cite{huang2018decorrelated}. In this work, we design a whitening normalization by employing Zero-phase Component Analysis (ZCA) sphering~\cite{huang2018decorrelated,huang2019iterative} to decorrelate any two graph signals $\vec{\boldsymbol s}_{i}$ and $\vec{\boldsymbol s}_{j}$ in the node embeddings $\mathbf{H}^{(l)}$. Generally, ZCA whitening operation involves computation-intensive Singular Value Decomposition (SVD) that is not GPU friendly:

\begin{equation}\label{eq:ZCA}
    \mathbf{H}_{{\rm ZCA}}^{(l)}=\left(\mathbf{H}^{(l)}-\vec{\boldsymbol 1}\cdot\left(\frac{1}{N}\vec{\boldsymbol 1}^{{\rm T}}\cdot\mathbf{H}^{(l)}\right)\right)\mathbf{U}^{(l)}\mathbf{\Lambda}^{(l)-\frac{1}{2}}\mathbf{U}^{(l){\rm T}},
\end{equation}

\noindent where $\mathbf{H}_{{\rm ZCA}}^{(l)}$ is the whitened embeddings, $\mathbf{\Lambda}^{(l)}={\rm diag}(\lambda_{1}, \cdots, \lambda_{D^{(l)}})^{(l)}$ and $\mathbf{U}^{(l)}=[\vec{\boldsymbol u}_{1},\cdots, \vec{\boldsymbol u}_{D^{(l)}}]^{(l)}$ are the eigenvalues and associated eigenvectors of the auto-covariance matrix $\mathbf{C}_{{\rm auto}}^{(l)}$ under $\mathbf{C}_{{\rm auto}}^{(l)-\frac{1}{2}}=\mathbf{U}^{(l)}\mathbf{\Lambda}^{(l)-\frac{1}{2}}\mathbf{U}^{(l){\rm T}}$. Fortunately, an iterative ZCA normalization enables more efficient whitening by using Newton's iterations~\cite{huang2019iterative}:

\begin{equation}\label{eq:Newton-ZCA}
\begin{split}
    \mathbf{P}_{p}=&\frac{1}{2}\left(3\cdot\mathbf{P}_{p-1}-\mathbf{P}_{p-1}^{3}\left(\frac{\mathbf{C}_{{\rm auto}}^{(l)}}{{\rm tr}\left(\mathbf{C}_{{\rm auto}}^{(l)}\right)}\right)\right), \\
    \mathbf{H}_{{\rm ZCA}}^{(l)}=&\left(\mathbf{H}^{(l)}-\vec{\boldsymbol 1}\cdot\left(\frac{1}{N}\vec{\boldsymbol 1}^{{\rm T}}\cdot\mathbf{H}^{(l)}\right)\right)\left(\frac{\mathbf{P}_{P}}{\sqrt{{\rm tr}(\mathbf{C}_{{\rm auto}}^{(l)})}}\right), \\
    & s.t.\ \mathbf{P}_{0}=\mathbf{I},\ 1 \leq p \leq P, \\
\end{split}
\end{equation}

\noindent where ${\rm tr}(\cdot)$ computes the trace of matrix $\mathbf{C}_{{\rm auto}}^{(l)}$, $\mathbf{I}$ is an identity matrix, and $\frac{\mathbf{P}_{p}}{\sqrt{{\rm tr}(\mathbf{C}_{{\rm auto}}^{(l)})}} \to \mathbf{C}_{{\rm auto}}^{(l)-\frac{1}{2}}$ when $p\to +\infty$. However, this iterative strategy using the Newton method may cause the gradient vanishing or exploding problem in the training pipeline due to the larger exponent $3$ to matrix $\mathbf{P}_{p}$. We hence use a Schur–Newton scheme~\cite{guo2010newton} instead of the traditional Newton method to compute ZCA sphering:

\begin{equation}\label{eq:Schur–Newton-ZCA}
\begin{split}
    \mathbf{P}_{p}=&\mathbf{P}_{p-1}\left(\frac{3\cdot\mathbf{I}-\mathbf{N}_{p-1}}{2}\right), \\
    \mathbf{N}_{p}=&\left(\frac{3\cdot\mathbf{I}-\mathbf{N}_{p-1}}{2}\right)^{2}\mathbf{N}_{p-1}, \\
    & s.t.\ \mathbf{P}_{0}=\mathbf{I},\ \mathbf{N}_{0}=\frac{\mathbf{C}_{{\rm auto}}^{(l)}}{{\rm tr}\left(\mathbf{C}_{{\rm auto}}^{(l)}\right)}, \ 1 \leq p \leq P, \\
\end{split}
\end{equation}

\noindent where $\mathbf{N}_{p}\to \mathbf{I}$ and $\frac{\mathbf{P}_{p}}{\sqrt{{\rm tr}(\mathbf{C}_{{\rm auto}}^{(l)})}} \to \mathbf{C}_{{\rm auto}}^{(l)-\frac{1}{2}}$ when $p\to +\infty$. Therefore, Eq.~(\ref{eq:Schur–Newton-ZCA}) employs Schur–Newton iterations with the smaller exponent of matrix $\mathbf{P}_{p}$ to compute ZCA whitening normalization to decorrelate graph signals. The whitening normalization pseudocode is shown in Appendix~\ref{app:pseudocodes}.

\textbf{Auto-Correlation Regularization.}~~Auto-correlation regularization employs the auto-covariance matrix $\mathbf{C}_{{\rm auto}}^{(l)}$ as an objective regularization introduced into the loss function:

\begin{equation}\label{eq:auto-correlation}
\begin{split}
   \mathbf{C}_{{\rm auto}[i,j]}^{(l)} =& \frac{\widehat{\vec{\boldsymbol s}}_{i}^{{\rm T}} \cdot \widehat{\vec{\boldsymbol s}}_{j}}{\|\widehat{\vec{\boldsymbol s}}_{i}\|_{2} \cdot \|\widehat{\vec{\boldsymbol s}}_{j}\|_{2}}, \\
   \mathcal{L}_{{\rm auto-reg}}^{(l)} =& \sum_{i}\left(1-\mathbf{C}_{{\rm auto}[i,i]}^{(l)}\right)^{2} + \beta \cdot \sum_{i}\sum_{j \neq i}\mathbf{C}_{{\rm auto}[i,j]}^{(l)2}, \\
   & s.t. \ 1 \leq i \leq D^{(l)}, 1 \leq j \leq D^{(l)}, \\
\end{split}
\end{equation}

\noindent where $\vec{\boldsymbol s}_{i} \in \mathbf{H}^{(l)}$, $\vec{\boldsymbol s}_{j} \in \mathbf{H}^{(l)}$, the centralized graph signal $\widehat{\vec{\boldsymbol s}}_{i}=\vec{\boldsymbol s}_{i}-\vec{\boldsymbol 1}\cdot \left(\frac{1}{N}\vec{\boldsymbol 1}^{{\rm T}}\cdot\vec{\boldsymbol s}_{i}\right)$,  and $\beta$ is a positive constant trading off the importance of the invariance term $\sum_{i}\left(1-\mathbf{C}_{{\rm auto}[i,i]}^{(l)}\right)^{2}$ and the redundancy reduction term $\sum_{i}\sum_{j \neq i}\mathbf{C}_{{\rm auto}[i,j]}^{(l)2}$~\cite{zbontar2021barlow}. Intuitively, the invariance term tries to
force the diagonal elements of the auto-correlation matrix $\mathbf{C}_{{\rm auto}}^{(l)}$ to $1$ and makes the graph signal invariant to the distortions applied. The redundancy reduction term tries to push the off-diagonal elements thereof to $0$ and decorrelates the different graph signal vectors to each other. Appendix~\ref{app:pseudocodes} shows the pseudocode of auto-correlation regularization.

\textbf{Cross-Correlation Regularization.}~~The definition of cross-correlation regularization is similar to Eq.~(\ref{eq:auto-correlation}) of auto-correlation regularization and the only difference is the cross covariance matrix $\mathbf{C}_{{\rm cross}}$ defined as:

\begin{equation}\label{eq:cross-covariance-matrix}
\begin{split}
    \mathbf{C}_{{\rm cross}}=&\frac{1}{N}\left(\mathbf{H}^{{\rm anchor}}-\vec{\boldsymbol 1}\cdot\left(\frac{1}{N}\vec{\boldsymbol 1}^{{\rm T}}\cdot\mathbf{H}^{{\rm anchor}}\right)\right)^{{\rm T}}\cdot \\
    &\left(\mathbf{H}^{{\rm view}}-\vec{\boldsymbol 1}\cdot\left(\frac{1}{N}\vec{\boldsymbol 1}^{{\rm T}}\cdot\mathbf{H}^{{\rm view}}\right)\right)+\epsilon\cdot\mathbf{I}, \\
\end{split}
\end{equation}

\noindent whose pseudocode is shown in Appendix~\ref{app:pseudocodes}.

\section{MLP-Based Training Framework}\label{sec:mlp-mased}

According to the relationship among GNN encoders, \emph{pretext} tasks, and \emph{downstream} tasks, there are three types of graph-based self-supervised training schemes~\cite{liu2022graph}. The first type is Pre-training and Fine-tuning (PT\&FT). The pre-training stage firstly initializes the parameters of the GNN encoder with \emph{pretext} tasks. After this, this pre-trained GNN encoder is fine-tuned under the supervision of specific \emph{downstream} tasks. The second is Joint Learning (JL) scheme, where the GNN encoder, \emph{pretext} and \emph{downstream} tasks are trained jointly. The last type is Unsupervised Representation Learning (URL). Akin to PT\&FT, URL also follows a two-stage training scheme, where the first stage trains the GNN encoder based on the \emph{pretext} task but in the second \emph{downstream} task stage, the GNN encoder is frozen. In our work, we adopt both JL and URL pipelines to train and evaluate our N2N network, and only URL to our NF-N2N strategy with GSD constraint.

\begin{figure*}[!t]
    \centering
    \includegraphics[scale=0.4]{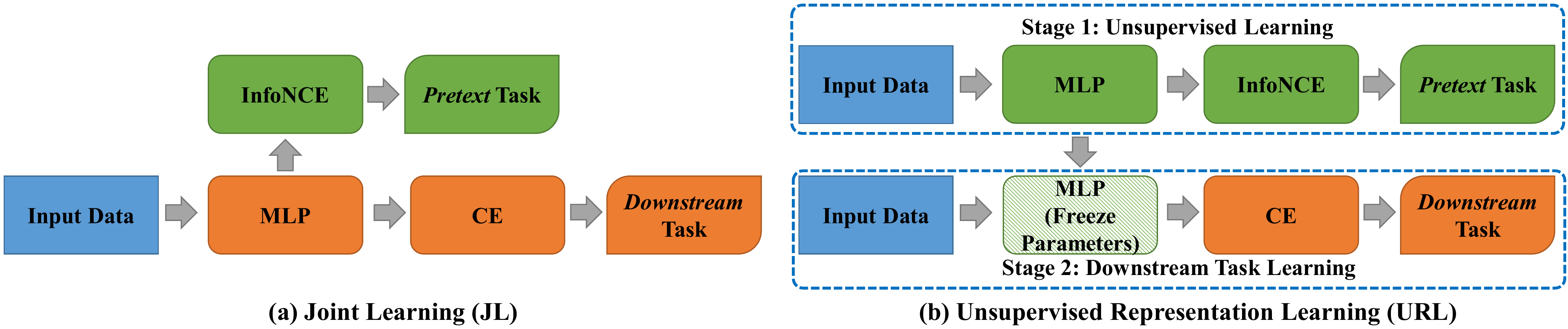}
    \caption{Illustration of two training pipelines adopted for the proposed model.}\label{fig:jl-url}
\end{figure*}

\textbf{JL Training Framework.}~~Fig.~\ref{fig:jl-url}~(a) illustrates our JL training pipeline to N2N strategy. As can be seen, unlike most existing GSSL work that uses GNN as node/graph encoder, we simply use a shallow MLP as encoder, which is more efficient. In JL scheme of N2N, we apply InfoNCE loss and Cross-Entropy loss jointly on top of the node representations obtained as output of the MLP encoder:

\begin{align}
\mathcal{L}=(1-\alpha)\mathcal{L}_{{\rm CE}} + \alpha\mathcal{L}_{{\rm InfoNCE}},
\end{align}

\noindent where $\alpha$ is a trade-off parameter used to balance the two loss functions.

\textbf{URL Training Framework.}~~The URL framework of N2N network, as shown in Fig.~\ref{fig:jl-url}~(b), involves two training stages: the pre-training \emph{pretext} task trains the MLP encoder using InfoNCE loss $\mathcal{L}_{{\rm InfoNCE}}$, and the \emph{downstream} task learns linear node classifiers using Cross-Entropy loss $\mathcal{L}_{{\rm CE}}$.

In URL training framework underpinning NF-N2N, Fig.~\ref{fig:MLP-based-training} shows the URL-based training pipeline of our augmentation-free graph method. In Fig.~\ref{fig:MLP-based-training}, we firstly input a positive pair into the MLP-based encoder of the pipeline, producing node representations. This pair includes an anchor node and a view node linked by one edge, and each anchor node has multi-view nodes. We then design the whitening normalization to decorrelate the graph signals of node representations in each MLP-based encoding layer. Finally, the whitened node representations are fed into MSE loss function and our designed auto- or cross-correlation regularization. The reason why JL is not used in NF-N2N is that two losses $\mathcal{L}_{{\rm CE}}$ and $\mathcal{L}_{{\rm MSE}}$ under JL pipeline require more memory consumption violating the original target of NF-N2N that is to reduce memory budgets.

\section{Experiments}\label{sec:experiments}

In this section, we start by introducing the experimental setups including datasets, existing methods, baselines, our models, and implementation details. We then show the performance comparison for existing methods. Finally, we present ablation studies to demonstrate other appealing properties of the proposed methods.

\subsection{Experimental Setups}\label{sec:setups}

In this section we introduce our experimental setups and begin by briefly describing two types of graph datasets: attributed and pure-structured graph datasets~\cite{yue2022survey}.

\textbf{Attributed Graph Datasets}~~We run experiments on seven prevailing node classification datasets, all of which are connected, undirected, and attributed graphs: Cora~\cite{fey2018splinecnn, hou2019measuring}, Pubmed~\cite{hou2019measuring}, Citeseer~\cite{hou2019measuring}, Cora Full~\cite{shchur2018pitfalls}, Amazon Photo~\cite{shchur2018pitfalls}, Coauthor CS~\cite{shchur2018pitfalls}, and Coauthor Physics~\cite{shchur2018pitfalls}. The first four datasets are constructed as citation networks. Amazon Photo is for the Amazon co-purchase graph, and Coauthor CS \& Physics are for co-authorship graphs. For Cora, Pubmed, and Citeseer datasets, we use the train/val/test splits in~\cite{fey2018splinecnn,hou2019measuring}. Table~\ref{table:dataset-statistics} shows the statistics of aforementioned attributed graph datasets\footnote{In this work, we did not use the prevalent Open Graph Benchmark~(OGB) datasets to evaluate our models, because the node features therein are mapped to the the space of $[-1, 1]^{D^{(l)}}$, while according to Theorem~\ref{theo:kl-mut}, the features are required to be in $[0, 1]^{D^{(l)}}$ as in ~\cite{hou2019measuring}.}.

\textbf{Pure-Structured Graph Datasets.}~~We further run node classification tasks on other seven pure-structured graph datasets, all of which only involve graph topology without node features: Feather-Lastfm~\cite{rozemberczki2020feather}, Musae-Github~\cite{rozemberczki2019multiscale}, Feather-Deezer~\cite{rozemberczki2020feather}, Twitch-Gamers~\cite{rozemberczki2021twitch}, Com-Amazon~\cite{yang2012defining}, Com-DBLP~\cite{yang2012defining}, and Com-Youtube~\cite{yang2012defining}. The first four datasets are constructed as social networks in which Feather-Lastfm, Musae-Github, and Feather-Deezer datasets only have incomplete node features. Hence, we use their adjacency matrix as a substitute for the node feature matrix. In this feature alternative, the number of feature dimension $D$ for a node $v_i$ feature vector equals the number of nodes $N$ and each element can be either $1$ or $0$ denoting $v_i$ connected or disconnected to another node $v_j$. Com-Amazon community dataset is the Amazon co-purchase network; Com-DBLP community dataset is a co-authorship network; Com-Youtube community dataset is a video-sharing network. All of the last three community datasets only include the graph structure without node features and their adjacency matrices are used as the node feature matrix in our experiments. The possible reason why they lack node features is the restriction of privacy policy on social or community networks or the expensive overheads of node feature collection. Table~\ref{table:dataset-statistics} shows the statistics of aforementioned pure-structured graph datasets.

\begin{table}[!h]
    \centering
    \caption{The statistics of the attributed and pure-structured graph datasets used in this work. Note that pure-structured graph datasets only involve graph topology without node features (In this table the mark ``---'' denotes ``no node features'').}\label{table:dataset-statistics}
    \resizebox{90mm}{!}{
    \begin{tabular}{l|cccccc}
        \toprule[1.2pt]
        \textbf{Dataset} & \textbf{Type} & \textbf{Nodes} & \textbf{Edges} & \textbf{Node Features} & \textbf{Classes} & \textbf{Density} \\
        \hline
        Cora & Attributed & 2,708 & 5,429 & 1,433 & 7 & 0.0014 \\
        Pubmed & Attributed & 19,217 & 44,338 & 500 & 3 & 0.0002 \\
        Citeseer & Attributed & 3,312 & 4,732 & 3,703 & 6 & 0.0008 \\
        Cora Full & Attributed & 20,095 & 62,421 & 8,710 & 67 & 0.0003 \\
        Amazon Photo & Attributed & 7,650 & 119,043 & 745 & 8 & 0.0040 \\
        Coauthor CS & Attributed & 18,333 & 81,894 & 6,805 & 15 & 0.0004 \\
        Coauthor Physics & Attributed & 34,493 & 247,962 & 8,415 & 5 & 0.0004 \\
        \hline
        Feather-Lastfm & Pure-Structured & 7,624 & 27,806 & --- & 18 & 0.0009 \\
        Musae-Github & Pure-Structured & 37,700 & 289,003 & --- & 2 & 0.0004 \\
        Feather-Deezer & Pure-Structured & 28,281 & 92,752 & --- & 2 & 0.0002 \\
        Twitch-Gamers & Pure-Structured & 168,113 & 1,048,575 & --- & 21 & 0.00007 \\
        Com-Amazon & Pure-Structured & 548,552 & 925,872 & --- & 24,104 & 0.000006 \\
        Com-DBLP & Pure-Structured & 425,957 & 1,049,865 & --- & 5,334 & 0.00001 \\
        Com-Youtube & Pure-Structured & 1,157,828 & 2,987,623 & --- & 2,883 & 0.000004 \\
        \bottomrule[1.2pt]
    \end{tabular}
    }
\end{table}

\textbf{Existing Methods.}~~We compare our proposed models to four types of existing methods, including (1) traditional GNNs such as GCN~\cite{kipf2017semi}, SGC~\cite{wu2019simplifying}, GraphSAGE~\cite{hamilton2017inductive}, GAT~\cite{velickovic2019graph}, SplineCNN~\cite{fey2018splinecnn}, and CS-GCN~\cite{hou2019measuring}; (2) traditional GSSL, including DeepWalk~\cite{perozzi2014deepwalk} and Node2Vec~\cite{grover2016node2vec}; (3) Graph Contrastive Learning (GCL) with negative pairs, including DGI~\cite{velivckovic2018deep}, GMI~\cite{peng2020graph}, MVGRL~\cite{hassani2020contrastive}, InfoGraph~\cite{sun2020infograph}, GCA~\cite{zhu2021graph}, GRACE~\cite{zhu2020deep}, AF-GCL~\cite{wang2022augmentation}, and Graph-MLP~\cite{hu2021graph}; (4) GSSL without negative pairs, including BGRL~\cite{thakoor2021bootstrapped},  SelfGNN~\cite{kefato2021self}, and AFGRL~\cite{lee2021augmentation}. Note that only Graph-MLP and our models use MLP as node encoder and other methods employ GNN as node encoder. SCGC~\cite{liu2022simple} has not been introduced into our experiments because it is designed for node clustering tasks.

\textbf{Our N2N Models.}~~We denote N2N-TAPS-$x$ as our model sampling top-$x$ positive neighbours based on TAPS, \textit{e.g.}, N2N-TAPS-1 samples the neighbour with maximum dependency value as positive. We evaluate our methods with \numrange{1}{5} positives to inspect how the positive size influences the node classification performance. We use N2N-random-$1$ as our baseline where one positive is sampled randomly from the neighbourhood of a node. By default, we aggregate all the neighbours of a node using \emph{mean} aggregator as positive, which is denoted as N2N.

\textbf{Baselines and Our NF-N2N Models.}~~We firstly design two baseline models: one is the combination of MLP and MSE (MLP-MSE in Table~\ref{table:overall}) without using the proposed decorrelation strategy; another is a standard node-node contrastive learning model (Contrastive in Table~\ref{table:overall}) employing the InfoNCE loss function performed on our augmentation-free framework. We then propose NF-N2N models with three GSD: first model incorporates \textbf{W}hitening normalization into MLP-MSE baseline to form NF-N2N (\textbf{W}); second model integrates MLP-MSE with \textbf{W}hitening normalization and \textbf{A}uto-correlation regularization as NF-N2N (\textbf{WA}); the last model combines MLP with \textbf{W}hitening normalization and \textbf{C}ross-correlation regularization to form NF-N2N (\textbf{WC}).

\textbf{Implementation Details.}~~For fair comparison, we follow the common practice to fix the number of hidden layers in our method and the compared GNNs and GCL encoders to be 2. For all datasets, we set the dimensionality of the hidden layer to be $512$. Some other important hyper-parameters include: dropout ratio is $0.6$ for Cora, Citeseer, and Coauthor CS, $0.2$ for Pubmed, $0.4$ for Amazon Photo, $0.5$ for Coauthor Physics, and $0.5$ for all pure-structured graphs; L2-regularization is $0.01$ for Cora, Citeseer, and Coauthor CS, $0.001$ for Pubmed and Amazon Photo, $0.05$ for Coauthor Physics, and $0.001$ for all pure-structured graphs; training epochs are $2000$ for all datasets; learning rate is $0.01$ for Pubmed and $0.001$ for the other datasets; nonlinear activation is ${\rm ReLU}$ function. For N2N-TAPS-$x$ (JL), $\alpha$ is set to $0.9$ for Cora, Pubmed, and Citeseer, $0.99$ for Amazon Photo and Coauthor CS, and $0.999$ for Coauthor Physics; temperature $\tau$ is $5$ for Cora, Pubmed, Citeseer, and Amazon Photo, $100$ for Coauthor CS, and $30$ for Coauthor Physics. The temperature $\tau$ of N2N-TAPS-$x$ (URL) is $5$ for all datasets. For all baselines and proposed NF-N2N methods, batch size is 2048; the iteration $P$ of whitening normalization and the positive constant $\beta$ for auto- and cross-correlation regularizations are tuned via cross-validation. These hyper-parameters are determined via cross-validation. We implement our models by Tensorflow 2.6. All of the experiments are performed on a machine with Intel CPU 12th Gen Intel Core i7-12700F, 32GB CPU memory, and GeForce RTX 3090 (24GB memory). We run all models five times on each dataset, and the mean and standard deviation of the micro-f1 score are used as the evaluation metric.

\vspace{-1em}
\subsection{Overall Results}\label{sec:overall}

\begin{table*}[!h]
    \caption{Performance comparison between existing methods and our methods on seven attributed graph datasets. Mean and standard deviation of 5-fold Micro-f1 scores are reported as evaluation metric. The best results for each dataset are highlighted in \textbf{bold}. The abbreviation OOM denotes the out-of-memory issue based on our hardware budget.}\label{table:overall}
    \centering
    \resizebox{180mm}{!}{
    \begin{tabular}{l|l|ccccccc}
    \toprule[1.2pt]
    \multirow{2}{*}{\textbf{Model Type}} & \multirow{2}{*}{\textbf{Model}} & \multirow{2}{*}{\textbf{Cora}} & \multirow{2}{*}{\textbf{Pubmed}} & \multirow{2}{*}{\textbf{Citeseer}} & \textbf{Cora} & \textbf{Amazon} & \textbf{Coauthor} & \textbf{Coauthor} \\
     & & & & & \textbf{Full} & \textbf{Photo} & \textbf{CS} & \textbf{Physics} \\
    \hline
     & GCN~\cite{kipf2017semi} & 84.95$\pm$0.12 & 87.22$\pm$0.10 & 78.00$\pm$0.10 & 60.20$\pm$0.08 & 89.96$\pm$0.12 & 90.20$\pm$0.12 & 91.76$\pm$0.08 \\
     & SGC~\cite{wu2019simplifying} & 84.22$\pm$0.10 & 86.52$\pm$0.12 & 77.68$\pm$0.20 & 57.36$\pm$0.25 & 88.25$\pm$0.28 & 90.16$\pm$0.18 & 91.26$\pm$0.08 \\
     & GraphSAGE~\cite{hamilton2017inductive} & 85.10$\pm$0.14 & 87.02$\pm$0.10 & 77.42$\pm$0.12 & 57.26$\pm$0.10 & 89.83$\pm$0.05 & 90.22$\pm$0.04 & 91.46$\pm$0.12 \\
    Traditional & GAT~\cite{velickovic2019graph} & 86.82$\pm$0.10 & 87.45$\pm$0.12 & 78.35$\pm$0.20 & 50.08$\pm$0.26 & 84.08$\pm$0.35 & 91.18$\pm$0.14 & 91.10$\pm$0.12 \\
    GNNs & SplineCNN~\cite{fey2018splinecnn} & 85.95$\pm$0.12 & 87.80$\pm$0.14 & 78.26$\pm$0.16 & 56.65$\pm$0.18 & 88.85$\pm$0.28 & 91.06$\pm$0.18 & 90.24$\pm$0.10 \\
     & CS-GCN~\cite{hou2019measuring} & 86.18$\pm$0.08 & 87.28$\pm$0.08 & 78.45$\pm$0.18 & 54.43$\pm$0.20 & 89.96$\pm$0.14 & 91.10$\pm$0.10 & 91.18$\pm$0.08 \\
    \hline
    Traditional & DeepWalk~\cite{perozzi2014deepwalk} & 80.16$\pm$0.18 & 86.42$\pm$0.16 & 63.50$\pm$0.22 & 50.82$\pm$0.13 & 82.70$\pm$0.24 & 82.76$\pm$0.30 & 86.15$\pm$0.32 \\
    GSSL & Node2Vec~\cite{grover2016node2vec} & 77.08$\pm$0.14 & 85.42$\pm$0.18 & 64.58$\pm$0.26 & 51.18$\pm$0.22 & 83.86$\pm$0.25 & 85.40$\pm$0.10 & 84.46$\pm$0.26 \\
    \hline
     & DGI~\cite{velivckovic2018deep} & 85.38$\pm$0.14 & 87.13$\pm$0.18 & 78.60$\pm$0.16 & 58.28$\pm$0.14 & 90.34$\pm$0.15 & 90.26$\pm$0.09 & 90.46$\pm$0.25 \\
     & GMI~\cite{peng2020graph} & 85.75$\pm$0.06 & 87.18$\pm$0.24 & 78.47$\pm$0.14 & 57.72$\pm$0.18 & 88.95$\pm$0.14 & 90.42$\pm$0.24 & 90.17$\pm$0.10 \\
     & MVGRL~\cite{hassani2020contrastive} & 85.16$\pm$0.18 & 86.96$\pm$0.15 & 77.99$\pm$0.14 & 58.26$\pm$0.12 & 88.20$\pm$0.15 & 90.36$\pm$0.14 & 89.24$\pm$0.15 \\
     & InfoGraph~\cite{sun2020infograph} & 84.68$\pm$0.15 & 87.46$\pm$0.06 & 78.55$\pm$0.10 & 58.83$\pm$0.20 & 90.14$\pm$0.28 & 90.06$\pm$0.14 & 90.26$\pm$0.18 \\
    GCL & GCA~\cite{zhu2021graph} & 86.28$\pm$0.28 & 86.56$\pm$0.18 & 77.26$\pm$0.18 & 55.28$\pm$0.18 & 89.20$\pm$0.14 & 91.10$\pm$0.25 & OOM \\
     & GRACE~\cite{zhu2020deep} & 85.25$\pm$0.28 & 87.26$\pm$0.34 & 77.62$\pm$0.28 & 55.96$\pm$0.18 & 90.15$\pm$0.22 & 91.12$\pm$0.24 & OOM \\
     & AF-GCL~\cite{wang2022augmentation} & 86.85$\pm$0.08 & 86.26$\pm$0.18 & 78.85$\pm$0.12 & 54.28$\pm$0.18 & 90.18$\pm$0.15 & 91.20$\pm$0.16 & 91.56$\pm$0.10 \\
     & Graph-MLP~\cite{hu2021graph} & 83.57$\pm$0.10 & 87.48$\pm$0.18 & 78.54$\pm$0.10 & 55.29$\pm$0.12 & 89.10$\pm$0.18 & 90.20$\pm$0.26 & 89.06$\pm$0.16 \\
    \hline
    GCL without & BGRL~\cite{thakoor2021bootstrapped} & 85.62$\pm$0.22 & 86.28$\pm$0.13 & 77.62$\pm$0.15 & 56.25$\pm$0.18 & 90.20$\pm$0.18 & 90.85$\pm$0.25 & 91.08$\pm$0.18 \\
    negative pairs & SelfGNN~\cite{kefato2021self} & 86.25$\pm$0.15 & 87.23$\pm$0.28 & 77.96$\pm$0.25 & 57.72$\pm$0.12 & 90.10$\pm$0.15 & 91.18$\pm$0.28 & 91.02$\pm$0.22 \\
     & AFGRL~\cite{lee2021augmentation} & 86.48$\pm$0.25 & 87.02$\pm$0.16 & 78.26$\pm$0.18 & 55.28$\pm$0.10 & 89.62$\pm$0.18 & 91.10$\pm$0.12 & 91.26$\pm$0.10 \\
    \hline
     & N2N-Random-1 (JL) & 83.46$\pm$0.18 & 86.20$\pm$0.08 & 76.85$\pm$0.30 & 53.28$\pm$0.26 & 86.25$\pm$0.15 & 89.65$\pm$0.06 & 88.75$\pm$0.12 \\
     & N2N-TAPS-1 (JL) & 85.46$\pm$0.08 & \textbf{91.08$\pm$0.16} & 80.24$\pm$0.13 & 57.56$\pm$0.12 & 89.90$\pm$0.08 & 92.07$\pm$0.06 & 90.70$\pm$0.20 \\
     & N2N-TAPS-2 (JL) & 86.36$\pm$0.16 & 90.76$\pm$0.14 & 80.34$\pm$0.06 & 58.06$\pm$0.18 & 89.77$\pm$0.07 & 92.11$\pm$0.08 & 90.28$\pm$0.24 \\
    Our N2N (JL) & N2N-TAPS-3 (JL) & 86.74$\pm$0.14 & 90.74$\pm$0.05 & 80.64$\pm$0.14 & 58.25$\pm$0.14 & 89.71$\pm$0.10 & 92.21$\pm$0.07 & 90.81$\pm$0.17 \\
     & N2N-TAPS-4 (JL) & 86.52$\pm$0.15 & 90.78$\pm$0.07 & \textbf{81.06$\pm$0.11} & 58.66$\pm$0.15 & 89.89$\pm$0.08 & 92.27$\pm$0.07 & 90.42$\pm$0.22 \\
     & N2N-TAPS-5 (JL) & 87.10$\pm$0.08 & 90.78$\pm$0.20 & 80.84$\pm$0.11 & 58.82$\pm$0.08 & 89.99$\pm$0.06 & 92.37$\pm$0.05 & 91.38$\pm$0.12 \\
     & N2N (JL) & 87.52$\pm$0.20 & 90.92$\pm$0.09 & 80.90$\pm$0.21 & 59.60$\pm$0.10 & 90.12$\pm$0.26 & 93.06$\pm$0.07 & 91.41$\pm$0.09 \\
    \hline
     & N2N-Random-1 (URL) & 82.50$\pm$0.10 & 84.08$\pm$0.12 & 75.06$\pm$0.12 & 53.15$\pm$0.14 & 86.07$\pm$0.12 & 88.23$\pm$0.08 & 86.56$\pm$0.04 \\
     & N2N-TAPS-1 (URL) & 84.66$\pm$0.11 & 88.42$\pm$0.07 & 77.44$\pm$0.07 & 57.10$\pm$0.16 & 89.40$\pm$0.05 & 91.72$\pm$0.03 & 90.35$\pm$0.06 \\
     & N2N-TAPS-2 (URL) & 85.60$\pm$0.08 & 89.26$\pm$0.08 & 78.56$\pm$0.08 & 58.24$\pm$0.12 & 90.49$\pm$0.04 & 91.56$\pm$0.03 & 90.68$\pm$0.02 \\
    Our N2N (URL) & N2N-TAPS-3 (URL) & 87.96$\pm$0.08 & 89.24$\pm$0.05 & 78.36$\pm$0.04 & 58.35$\pm$0.18 & 90.61$\pm$0.05 & 91.53$\pm$0.08 & 91.20$\pm$0.08 \\
     & N2N-TAPS-4 (URL) & 88.04$\pm$0.11 & 89.32$\pm$0.09 & 78.54$\pm$0.07 & 58.48$\pm$0.12 & 90.35$\pm$0.07 & 92.03$\pm$0.06 & 90.89$\pm$0.05 \\
     & N2N-TAPS-5 (URL) & 87.84$\pm$0.09 & 89.88$\pm$0.05 & 79.08$\pm$0.07 & 58.82$\pm$0.12 & 90.65$\pm$0.07 & 91.99$\pm$0.04 & 91.52$\pm$0.02 \\
     & N2N (URL) & 88.20$\pm$0.05 & 89.30$\pm$0.04 & 79.54$\pm$0.02 & 59.35$\pm$0.20 & \textbf{91.08$\pm$0.11} & 92.31$\pm$0.05 & 91.77$\pm$0.08 \\
    \hline
    \multirow{2}{*}{Baselines} & MLP-MSE & 36.30$\pm$0.18 & 49.00$\pm$0.09 & 69.90$\pm$0.14 & 4.46$\pm$0.34 & 22.57$\pm$0.12 & 25.47$\pm$0.08 & 50.87$\pm$0.12 \\
     & Contrastive & 86.30$\pm$0.12 & 87.00$\pm$0.09 & 78.30$\pm$0.05 & 51.29$\pm$0.16 & 88.43$\pm$0.18 & 90.63$\pm$0.08 & 91.25$\pm$0.10 \\
    \hline
     & NF-N2N (W) & \textbf{89.00$\pm$0.16} & 87.90$\pm$0.10 & 80.30$\pm$0.14 & \textbf{61.76$\pm$0.04} & 89.86$\pm$0.12 & 93.10$\pm$0.08 & 92.55$\pm$0.12 \\
    Our NF-N2N & NF-N2N (WA) & 88.80$\pm$0.13 & 88.10$\pm$0.05 & 79.50$\pm$0.08 & 60.96$\pm$0.16 & 90.23$\pm$0.08 & \textbf{93.35$\pm$0.12} & \textbf{93.62$\pm$0.15} \\
     & NF-N2N (WC) & 88.80$\pm$0.06 & 88.70$\pm$0.10 & 80.50$\pm$0.06 & 61.54$\pm$0.08 & 90.52$\pm$0.18 & 93.25$\pm$0.15 & 93.46$\pm$0.16 \\
    \bottomrule[1.2pt]
    \end{tabular}
    }
\end{table*}

In this section, we demonstrate our overall experimental results. Table~\ref{table:overall} shows the micro-f1 performance comparison between the proposed methods and existing methods on seven attributed graph datasets. From the results we obtain the following observations: (1)~The proposed N2N and NF-N2N models consistently outperform the comparing methods on all these seven datasets. The improvement can go up to 2\% on Cora, Citeseer, Coauthor CS, and Coauthor Physics, and 3\% on Pubmed. This shows the competitiveness of our N2N mutual information maximization strategy over GNNs and other GCL based solutions for node representation learning and indicates that high-quality node representations can be learned through our NF-N2N with GSD based constraints even using none of the feature aggregation, graph augmentation or negative pairs. At the same time, since our method is free from the aforementioned expensive operations, our methods enjoy better efficiency in terms of training and inference. (2)~Within the N2N family, we generally observe improvement when sampling more positives based on TAPS but the improvements are marginal. This demonstrates the potential of N2N-TAPS-1 as it avoids neighbourhood aggregation operation which is known to be expensive. However, when the single positive is sampled randomly from the neighbourhood, the performance drops significantly. This result shows the proposed TAPS strategy indeed can sample topologically meaningful positives. (3)~Comparing to MLP-MSE baseline, we observe significant improvement from NF-N2N. This is easy to understand as directly minimizing MSE can lead to trivial solution and consequently representation collapse. (4)~Comparing to Contrastive baseline, our NF-N2N methods achieve superior performance and the margins can go up to 3\% on Cora and Coauthor CS, 10\% on Cora Full, and 2\% on Citeseer and Coauthor Physics. The potential solution to improve the performance of Contrastive baseline is to increase the negative size, which, however, can lead to higher GPU memory consumption. (5)~Within the existing methods, the GCL solutions have comparable performance or even slightly better performance comparing to the fully supervised GNN variants. This observation shows SSL can be a promising alternative in graph-based representation learning.

\begin{table}[!h]
    \caption{Comparison of the GPU memory (MBs) usage (GPU memory usage during training) among existing methods, Contrastive baseline, and the proposed NF-N2N methods.}\label{table:memory}
    \centering
    \resizebox{90mm}{!}{
    \begin{tabular}{l|ccccccc}
    \toprule[1.2pt]
    \multirow{2}{*}{\textbf{Model}} & \multirow{2}{*}{\textbf{Cora}} & \multirow{2}{*}{\textbf{Pubmed}} & \multirow{2}{*}{\textbf{Citeseer}} & \textbf{Cora} & \textbf{Amazon} &  \textbf{Coauthor} & \textbf{Coauthor} \\
     & & & & \textbf{Full} & \textbf{Photo} &  \textbf{CS} & \textbf{Physics} \\
    \hline
    GCN~\cite{kipf2017semi} & 1,166 & 1,332 & 1,234 & 2,500 & 1,209 & 2,118 & 3,995 \\
    GCA~\cite{zhu2021graph} & 1,257 & 11,614 & 1,508 & 14,302 & 4,136 & 13,908 & OOM \\
    Graph-MLP~\cite{hu2021graph} & 1,237 & 4,018 & 1,395 & 5,459 & 1,803 & 3,624 & 7,216 \\
    BGRL~\cite{thakoor2021bootstrapped} & 1,412 & 1,705 & 1,668 & 3,645 & 1,509 & 3,164 & 5,858 \\
    Contrastive & 1,058 & 1,280 & 1,157 & 2,756 & 1,233 & 2,868 & 3,292 \\
    \hline
    NF-N2N (W) & 746 & 1,056 & 1,056 & 2,492 & 1,026 & 2,089 & 2,552 \\
    NF-N2N (WA) & 1,020 & 1,170 & 1,054 & 2,526 & 1,035 & 2,098 & 2,582 \\
    NF-N2N (WC) & 800 & 1,020 & 1,016 & 2,492 & 1,000 & 2,073 & 2,568 \\
    \bottomrule[1.2pt]
    \end{tabular}
    }
\end{table}

Our NF-N2N methods are expected to be more memory efficient comparing to existing work and baseline models as our work adopts simple MLP as node encoder and thus fully gets rid of the expensive node aggregation operation. Table~\ref{table:memory} shows the GPU memory comparison among representative methods and baselines, such as traditional GCN~\cite{kipf2017semi}, GCA~\cite{zhu2021graph} and Graph-MLP~\cite{hu2021graph} in GCL, BGRL~\cite{thakoor2021bootstrapped} in GCL without negative pairs, and Contrastive baseline. From the results we can see our NF-N2N methods consume the least GPU memory usage.

\begin{table*}[!h]
    \caption{Performance comparison between existing methods and our NF-N2N methods on seven pure-structured graph datasets.
    In SGC~\cite{wu2019simplifying}, $x$-hop denotes the aggregation operations with $x$ times.}\label{table:overall-large}
    \centering
    \resizebox{180mm}{!}{
    \begin{tabular}{l|ccccccc}
    \toprule[1.2pt]
    \multirow{2}{*}{\textbf{Model}} & \textbf{Feather-} & \textbf{Musae-} & \textbf{Feather-} & \textbf{Twitch-} & \textbf{Com-} & \textbf{Com-} & \textbf{Com-} \\
     & \textbf{Lastfm} & \textbf{Github} & \textbf{Deezer} & \textbf{Gamers} & \textbf{Amazon} & \textbf{DBLP} & \textbf{Youtube} \\
    \hline
    MLP & 82.21$\pm$0.28 & 81.42$\pm$0.16 & 52.56$\pm$0.30 & 73.28$\pm$0.42 & 84.42$\pm$0.25 & 86.47$\pm$0.10 & 38.25$\pm$0.26 \\
    MLP-MSE & 84.37$\pm$0.18 & 80.52$\pm$0.14 & 50.28$\pm$0.25 & 71.52$\pm$0.20 & 44.25$\pm$0.36 & 28.21$\pm$0.48 & 30.58$\pm$0.64 \\
    SGC~(1-hop)~\cite{wu2019simplifying} & 85.62$\pm$0.32 & 82.82$\pm$0.20 & 56.27$\pm$0.14 & OOM & 86.20$\pm$0.18 & 89.28$\pm$0.15 & OOM \\
    SGC~(2-hop)~\cite{wu2019simplifying} & 86.24$\pm$0.15 & OOM & 55.57$\pm$0.18 & OOM & OOM & OOM & OOM \\
    \hline
    NF-N2N (W) & 86.70$\pm$0.12 & \textbf{84.59$\pm$0.18} & \textbf{56.63$\pm$0.16} & \textbf{76.12$\pm$0.20} & \textbf{87.28$\pm$0.14} & \textbf{90.87$\pm$0.24} & \textbf{46.87$\pm$0.16} \\
    NF-N2N (WA) & 87.10$\pm$0.12 & 83.90$\pm$0.18 & 55.46$\pm$0.25 & 75.52$\pm$0.26 & 86.54$\pm$0.26 & 89.62$\pm$0.16 & 44.25$\pm$0.28 \\
    NF-N2N (WC) & \textbf{87.18$\pm$0.18} & 84.52$\pm$0.24 & 55.81$\pm$0.16 & 75.86$\pm$0.30 & 86.35$\pm$0.18 & 89.65$\pm$0.46 & 44.26$\pm$0.10 \\
    \bottomrule[1.2pt]
    \end{tabular}
    }
\end{table*}

Table~\ref{table:overall-large} demonstrates the micro-f1 performance comparison between the proposed NF-N2N models and existing methods on seven pure-structured graph datasets. In this table, only supervised-based SGC model and our NF-N2N models can run on such large graph datasets. Most of other existing models and the N2N method suffer from out-of-GPU-memory, since such models use the GNN encoder in their training phase, employ the contrastive negative pairs, or adopt sophisticated augmentation strategies in their framework. Even the TAPS strategy encounters GPU memory issue on most of such datasets. Hence, we remove these out-of-memory~(OOM) models from Table~\ref{table:overall-large}. From this table we observe the following conclusions: (1) The proposed NF-N2N models achieve promising results on all these seven datasets. The advantage can go up to 2\% on Musae-Github, 4\% on Twitch-Gamers, and 8\% on Com-Youtube. This shows our proposed NF-N2N models can effectively enrich the node representations based on the adjacency information. (2) The SGC method with $2$-hop aggregations is prone to have OOM issue based on its hardware budget in spite of the aggregation performed in advance via using cheap and sufficient CPU memory. This issue indicates that the standard aggregation of GNNs is invalid when the number of nodes and feature dimensions (represented by the size of adjacency matrix) explode. (3) Comparing to MLP-MSE baseline, our NF-N2N observes significant  improvement again, which validates the effectiveness of the GSD constraint in terms of preventing representation collapse.

\vspace{-1em}
\subsection{Ablation Studies for N2N}\label{sec:ablation-n2n}

In this section, we conduct additional ablation studies to reveal other appealing properties of the proposed N2N methods.

\begin{table}[!h]
    \caption{Performance of N2N (JL) model with random positive sampling on Amazon Photo and Coauthor Physics. We vary the positive sampling size from 1 to 5 and for each sampling size we run the experiments three times with different random seeds.}\label{table:random}
    \centering
    \resizebox{90mm}{!}{
    \begin{tabular}{l|ccc|ccc}
    \toprule[1.2pt]
    \textbf{Model} & \multicolumn{3}{c}{\textbf{Amazon Photo}} & \multicolumn{3}{|c}{\textbf{Coauthor Physics}} \\
    \hline
    Random Seeds & 1 & 2 & 3 & 1 & 2 & 3 \\
    \hline
    N2N-Random-1 (JL) & 86.25$\pm$0.15 & 85.52$\pm$0.14 & 84.08$\pm$0.20 & 88.75$\pm$0.12 & 87.25$\pm$0.10 & 86.20$\pm$0.28 \\
    N2N-Random-2 (JL) & 87.68$\pm$0.28 & 85.74$\pm$0.18 & 85.62$\pm$0.38 & 88.62$\pm$0.24 & 87.28$\pm$0.14 & 86.02$\pm$0.10 \\
    N2N-Random-3 (JL) & 87.95$\pm$0.10 & 86.28$\pm$0.08 & 85.20$\pm$0.16 & 89.24$\pm$0.16 & 87.64$\pm$0.10 & 86.52$\pm$0.14 \\
    N2N-Random-4 (JL) & 88.25$\pm$0.26 & 86.06$\pm$0.20 & 86.21$\pm$0.42 & 88.82$\pm$0.10 & 86.52$\pm$0.20 & 87.30$\pm$0.10 \\
    N2N-Random-5 (JL) & 88.30$\pm$0.12 & 85.52$\pm$0.14 & 86.56$\pm$0.30 & 89.65$\pm$0.20 & 87.82$\pm$0.15 & 87.22$\pm$0.14 \\
    \bottomrule[1.2pt]
    \end{tabular}
    }
\end{table}

\textbf{N2N (JL) based on Random Positive Sampling}.~~To further justify the necessity and advantage of our TAPS strategy, we run experiments on random positive sampling by varying the sampling size from 1 to 5. We choose two datasets, \textit{i.e.}, Amazon Photo and Coauthor Physics, for this experiment because their average node degree $> 5$. For each sampling size, we run the experiments three times with different random seeds. The results are shown in Table~\ref{table:random}. From the table we can clearly observe that random positive sampling results in large performance variance which means random sampling fails to identify consistent and informative neighbours.

\begin{table}[!h]
    \caption{The time cost comparison among typical GNN/GCL methods and our N2N-TAPS-$x$ models. The number indicates the time consumption for training a method on a dataset for one epoch. The symbol $\dagger$ behind N2N-TAPS-$x$ models indicates inference time. The unit of time is milliseconds (ms).}\label{table:time-consumption}
    \centering
    \resizebox{90mm}{!}{
    \begin{tabular}{l|cccccc}
    \toprule[1.2pt]
    \multirow{2}{*}{\textbf{Model}} & \multirow{2}{*}{\textbf{Cora}} & \multirow{2}{*}{\textbf{Pubmed}} & \multirow{2}{*}{\textbf{Citeseer}} & \textbf{Amazon} & \textbf{Coauthor} & \textbf{Coauthor} \\
    & & & & \textbf{Photo} & \textbf{CS} & \textbf{Physics} \\
    \hline
    GCN~\cite{kipf2017semi} & 1.68 & 5.88 & 6.64 & 6.08 & 10.06 & 11.31 \\
    GraphSAGE-Mean~\cite{hamilton2017inductive} & 1.56 & 6.52 & 6.46 & 6.28 & 9.65 & 12.06 \\
    FastGCN~\cite{chen2018fastgcn} & 1.21 & 5.73 & 6.89 & 5.24 & 9.06 & 10.04 \\
    DGI~\cite{velivckovic2018deep} & 1.82 & 7.20 & 8.56 & 8.25 & 12.00 & 15.68 \\
    GMI~\cite{peng2020graph} & 2.08 & 7.93 & 9.62 & 10.18 & 15.54 & 21.16 \\
    MVGRL~\cite{hassani2020contrastive} & 1.95 & 7.64 & 9.06 & 9.30 & 13.11 & 16.92 \\
    InfoGraph~\cite{sun2020infograph} & 1.86 & 8.65 & 11.16 & 11.42 & 12.82 & 15.20 \\
    Graph-MLP~\cite{hu2021graph} & 0.45 & 1.08 & 0.56 & 0.50 & 1.82 & 3.64 \\
    \hline
    N2N-TAPS-1 (JL) & 0.09 & 0.12 & 0.11 & 0.12 & 0.22 & 0.24 \\
    N2N-TAPS-5 (JL) & 0.15 & 0.14 & 0.18 & 0.19 & 0.46 & 0.63 \\
    N2N (JL)        & 0.30 & 0.19 & 0.19 & 0.20 & 1.19 & 1.68 \\
    \hline
    N2N-TAPS-1 (JL)$\dagger$ & 0.05 & 0.07 & 0.13 & 0.08 & 0.06 & 0.07 \\
    N2N-TAPS-5 (JL)$\dagger$ & 0.06 & 0.09 & 0.16 & 0.09 & 0.08 & 0.10 \\
    N2N (JL)$\dagger$        & 0.08 & 0.11 & 0.19 & 0.09 & 0.12 & 0.13 \\
    \hline
    N2N-TAPS-1 (URL) & 0.10 & 0.15 & 0.13 & 0.16 & 0.30 & 0.28 \\
    N2N-TAPS-5 (URL) & 0.18 & 0.18 & 0.16 & 0.25 & 0.42 & 0.54 \\
    N2N (URL)        & 0.22 & 0.25 & 0.18 & 0.38 & 1.25 & 1.80 \\
    \hline
    N2N-TAPS-1 (URL)$\dagger$ & 0.06 & 0.04 & 0.15 & 0.09 & 0.08 & 0.10 \\
    N2N-TAPS-5 (URL)$\dagger$ & 0.05 & 0.06 & 0.16 & 0.10 & 0.10 & 0.13 \\
    N2N (URL)$\dagger$        & 0.06 & 0.06 & 0.16 & 0.14 & 0.13 & 0.15 \\
    \bottomrule[1.2pt]
    \end{tabular}
    }
\end{table}

\textbf{Time Consumption.}~~Our methods are expected to be more efficient comparing to existing work. On one hand, our work adopts MLP as node encoder and thus avoids the expensive node aggregation in the encoding phase. On the other hand, TAPS enables us to sample limited high-quality positives upfront. Especially, when one positive is selected, we fully get rid of the aggregation operation.

Table~\ref{table:time-consumption} shows the time consumption comparison. From the results we can see our methods can be orders of faster than the typical GNN and GCL based methods. Graph-MLP~\cite{hu2021graph} also adopts MLP as encoder but it aligns a node to all the nodes that can be reached from this node. This explains its slowness on large datasets such as CS and Physics.

\begin{figure}[!h]
    \centering
    \includegraphics[scale=0.3]{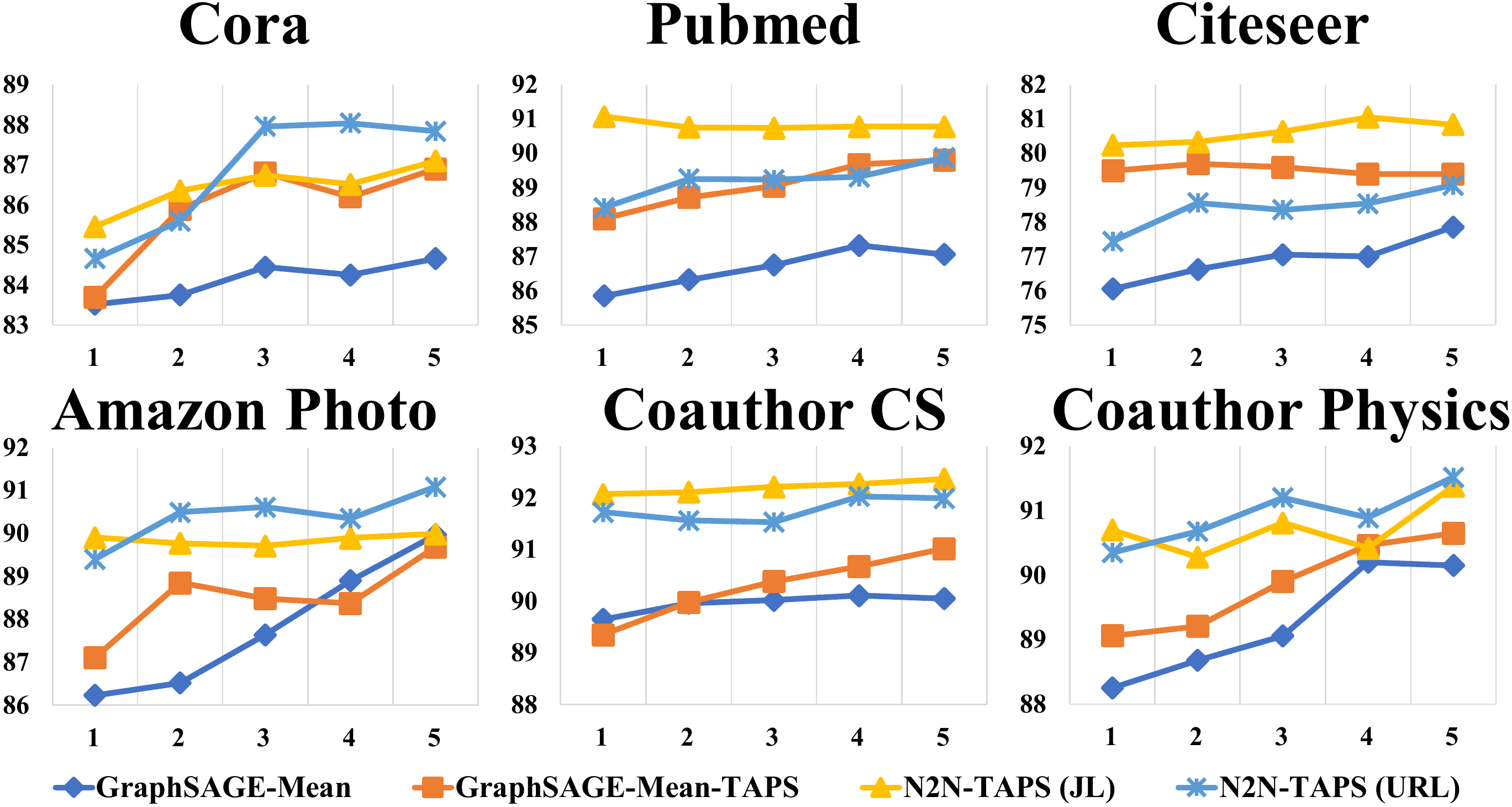}
    \caption{The performance comparison among GraphSAGE-Mean, GraphSAGE-Mean-TAPS, and N2N-TAPS-$x$ (JL and URL) based on varying number of sampled neighbours. Random neighbour sampling is used for GraphSAGE-Mean and TAPS is used for GraphSAGE-Mean-TAPS.}\label{fig:sampling}
\end{figure}

\textbf{Evaluation of TAPS Strategy.}~~TAPS is an important component in our framework to ensure the quality and efficiency of positive sampling. In Table~\ref{table:overall} we have shown the advantage of TAPS over random sampling on our N2N-TAPS-1 model. In this section, we apply TAPS sampling to another sampling based GNN baseline, GraphSAGE-Mean, to verify if TAPS can be used as a general neighbourhood sampling strategy to identify informative neighbours. Fig.~\ref{fig:sampling} shows the results. By default, GraphSAGE-Mean~\cite{hamilton2017inductive} uses random sampling to select neighbours for aggregation, which has the risk of absorbing noisy information. We replace random sampling in GraphSAGE-Mean with TAPS and leave all the other implementation intact. Its performance is obviously boosted and generally using more neighbours can benefit the performance more. This observation shows us again that it is important to consider the structural dependencies to select useful neighbours to enrich node representations.

\begin{figure}[!h]
    \centering
    \includegraphics[scale=0.27]{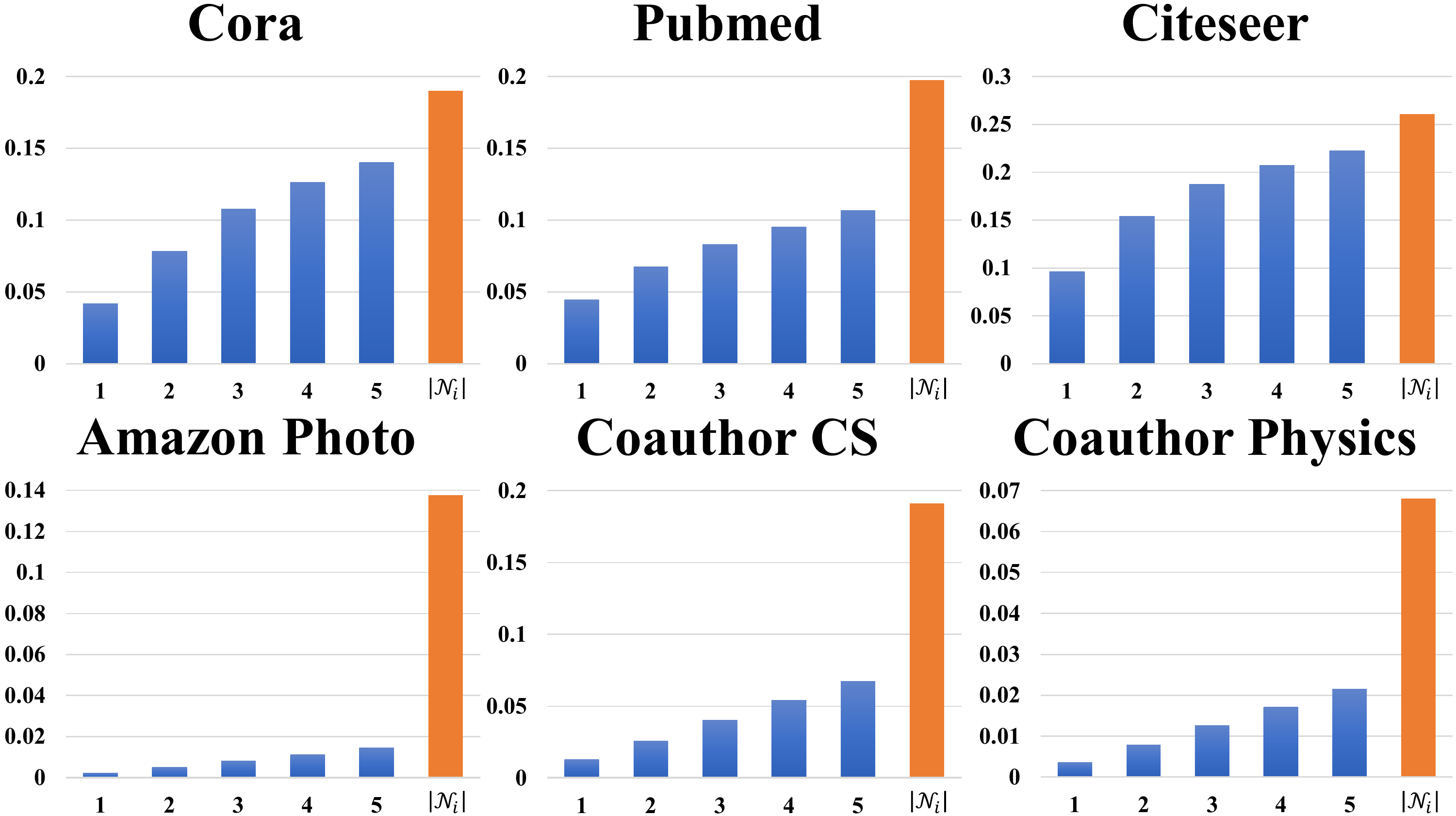}
    \caption{Label smoothness values obtained by using TAPS strategy to sample 1 to 5 neighbours (blue bars) and the label smoothness value obtained by considering all neighbours without any sampling strategy (orange bar).}\label{fig:label-smoothness}
\end{figure}
\begin{figure}[!h]
    \centering
    \includegraphics[scale=0.47]{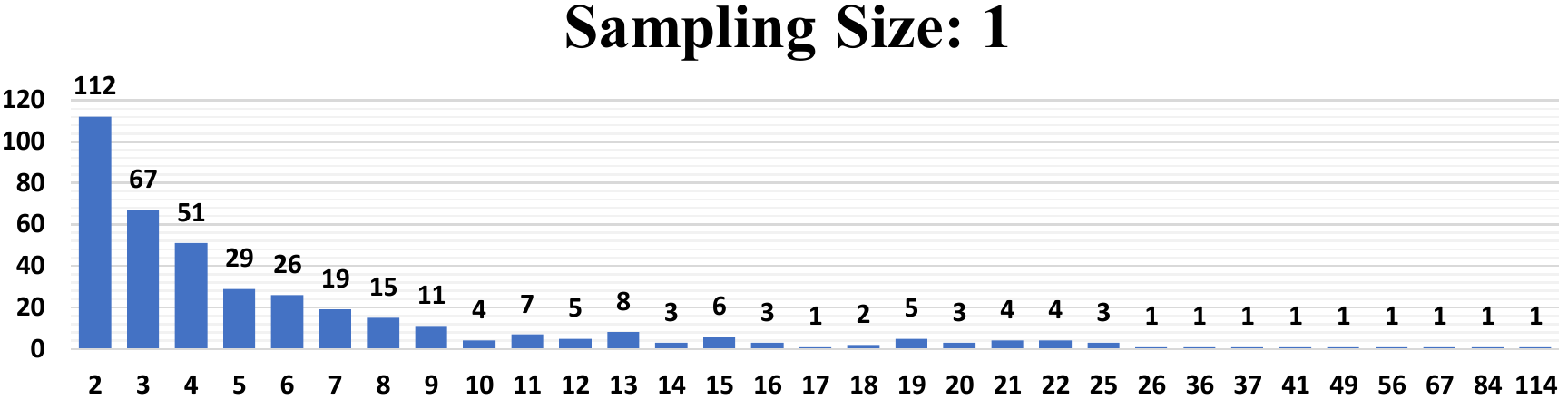}
    \caption{Subgraph statistics on Cora by using TAPS strategy to perform the subgraph partition. The horizontal axis indicates the number of nodes in the subgraph; the vertical axis implies the number of such subgraphs.}\label{fig:subgraph}
\end{figure}
\begin{figure}[!h]
    \centering
    \includegraphics[scale=0.21]{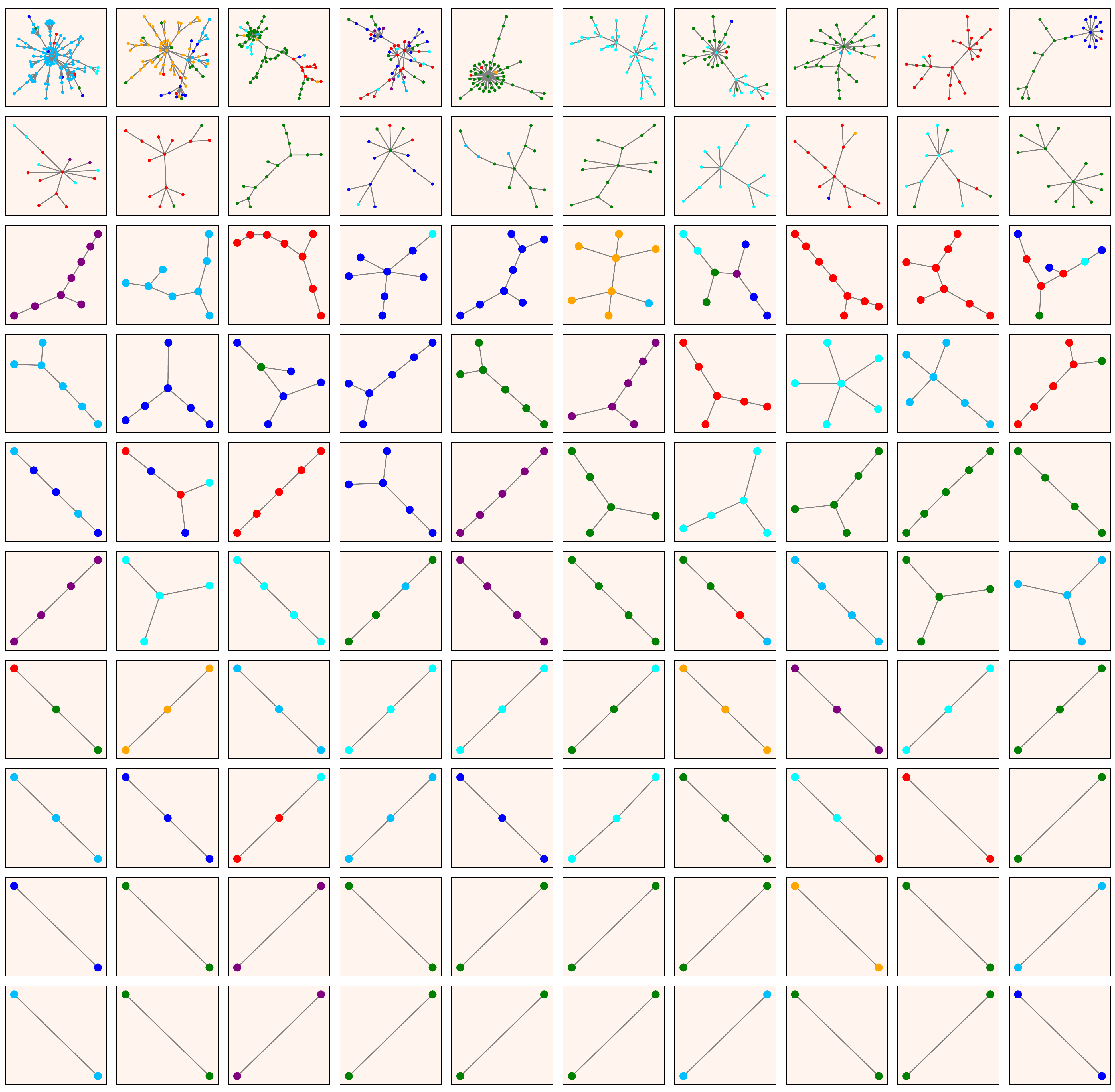}
    \caption{Visualization of part of the subgraphs derived from TAPS on Cora. We rank the sizes of the subgraphs and divide them into 10 intervals. For each row, we visualize the top-10 subgraphs within the corresponding interval from top (top $10\%$ interval) to bottom  ($80\% \sim 90\%$ interval). Different node colors represent different labels.}\label{fig:subgraph-shown}
\end{figure}

\textbf{Label Smoothness Analysis.}~To verify the quality of the neighbourhood sampling by using TAPS strategy, we introduce the label smoothness metric proposed in CS-GNN~\cite{hou2019measuring}: $\delta_{l}=\sum_{(v_{i},v_{j})\in \mathcal{E}}(1-\mathbb{I}(v_{i}\simeq v_{j}))/|\mathcal{E}|$, where $\mathbb{I}(\cdot)$ is an indicator function that has $\mathbb{I}(v_{i}\simeq v_{j})=1$ if $y_{v_{i}}=y_{v_{j}}$. Otherwise, $\mathbb{I}(v_{i}\simeq v_{j})=0$ when $y_{v_{i}}\neq y_{v_{j}}$. By virtue of the label smoothness, a large $\delta_{l}$ suggests that the nodes with different labels are regarded to be connected neighbours, while a small $\delta_{l}$ indicates a graph $\mathcal{G}$ possessing the higher-quality neighbourhood structure, \textit{i.e.}, most neighbours of a node have the same label as the node. Such high-quality neighbourhood can contribute homogeneous information gain to their corresponding central nodes~\cite{hou2019measuring}.

Fig.~\ref{fig:label-smoothness} shows that the label smoothness values gradually increases by expanding the sampling size from 1 to 5 with our TAPS strategy. Without any sampling strategy, the label smoothness value of the whole graph is the highest. This phenomenon suggests that our TAPS strategy can promote the neighbourhood sampling quality as the sampling size decreases, explaining why the proposed N2N-TAPS-1 model has competitive performance on some datasets.

TAPS strategy is essentially a subgraph partition scheme. A good partition is expected to lead to subgraphs faithful to the node labels. Fig.~\ref{fig:subgraph} shows the statistical distribution in terms of subgraph size (nodes in a subgraph) and the number of such subgraphs derived from TAPS. The details of the subgraph partition on Cora are visualized in Fig.~\ref{fig:subgraph-shown}, where different node colors represent different labels. In each of the subgraph, most of the nodes have the same color (same label), even in some large subgraphs, implying TPAS generates high-quality neighbourhood. This visualization also reveals that our TAPS strategy is able to model multi-hop contextual information in graph although we do not explicitly do so. The details of the statistical distribution and the subgraph partition for other datasets can be found in Appendix~\ref{appe:ss} and~\ref{appe:spv}.

\vspace{-1em}
\subsection{Ablation Studies for NF-N2N}\label{sec:ablation-nf-n2n}

In this section, we perform ablation studies to explore other interesting properties of the proposed NF-N2N methods.

\textbf{Combination of Different Constraints.}~~In this section, we evaluate different combinations of strategies proven effective to avoid representation collapse and see if such combinations can further boost the performance. The concrete combinations and the corresponding results are reported in Table~\ref{table:pluggable}. From the results we can observe that: (1)~The solutions with WN involved can generally achieve promising performance. (2)~When applying BN together with MSE minimization, trivial solution can also be effectively avoided, where similar conclusions are also discovered in~\cite{grill2020bootstrap}. (3)~When auto- or cross-correlation regularization is used alone, we can still get very competitive node classification performance. (4)~Adding BN on top of our proposed regularizations does not observe consistent improvement.

\begin{table*}[!h]
    \caption{Effect of incorporating Batch Normalization (BN), Whitening Normalization (WN), MSE, Auto-correlation Regularization (AR), Cross-correlation Regularization (CR) and their combinations on node classification performance.}\label{table:pluggable}
    \centering
    \resizebox{180mm}{!}{
    \begin{tabular}{ccccc|ccccccc}
    \toprule[1.2pt]
    \multirow{2}{*}{\textbf{BN}} & \multirow{2}{*}{\textbf{WN}} & \multirow{2}{*}{\textbf{MSE}} & \multirow{2}{*}{\textbf{AR}} & \multirow{2}{*}{\textbf{CR}} & \multirow{2}{*}{\textbf{Cora}} & \multirow{2}{*}{\textbf{Pubmed}} & \multirow{2}{*}{\textbf{Citeseer}} & \textbf{Cora} & \textbf{Amazon} & \textbf{Coauthor} & \textbf{Coauthor} \\
     & & & & & & & & \textbf{Full} & \textbf{Photo} & \textbf{CS} & \textbf{Physics} \\
    \hline
    \color{red}{\XSolidBold} & \color{red}{\XSolidBold} & \color{green}{\CheckmarkBold} & \color{red}{\XSolidBold} & \color{red}{\XSolidBold} & 36.30$\pm$0.18 & 49.00$\pm$0.09 & 69.90$\pm$0.14 & 4.46$\pm$0.34 & 22.57$\pm$0.12 & 25.47$\pm$0.08 & 50.87$\pm$0.12 \\
    \color{red}{\XSolidBold} & \color{red}{\XSolidBold} & \color{green}{\CheckmarkBold} & \color{green}{\CheckmarkBold} & \color{red}{\XSolidBold} & 87.40$\pm$0.06 & 87.20$\pm$0.14 & 79.00$\pm$0.10 & 55.91$\pm$0.16 & 89.54$\pm$0.12 & 91.03$\pm$0.05 & 94.27$\pm$0.02 \\
    \color{red}{\XSolidBold} & \color{red}{\XSolidBold} & \color{red}{\XSolidBold} & \color{red}{\XSolidBold} & \color{green}{\CheckmarkBold} & 87.80$\pm$0.10 & 87.80$\pm$0.15 & 78.80$\pm$0.02 & 53.46$\pm$0.10 & 89.49$\pm$0.05 & 89.05$\pm$0.08 & 93.90$\pm$0.05 \\
    \hline
    \color{green}{\CheckmarkBold} & \color{red}{\XSolidBold} & \color{green}{\CheckmarkBold} & \color{red}{\XSolidBold} & \color{red}{\XSolidBold} & 88.20$\pm$0.06 & 75.80$\pm$0.08 & 79.50$\pm$0.03 & 47.45$\pm$0.09 & 83.99$\pm$0.05 & 80.44$\pm$0.08 & 87.16$\pm$0.14 \\
    \color{green}{\CheckmarkBold} & \color{red}{\XSolidBold} & \color{green}{\CheckmarkBold} & \color{green}{\CheckmarkBold} & \color{red}{\XSolidBold} & 88.70$\pm$0.05 & 86.80$\pm$0.04 & 79.30$\pm$0.08 & 57.54$\pm$0.06 & 88.79$\pm$0.06 & 91.40$\pm$0.10 & 93.53$\pm$0.06 \\
    \color{green}{\CheckmarkBold} & \color{red}{\XSolidBold} & \color{red}{\XSolidBold} & \color{red}{\XSolidBold} & \color{green}{\CheckmarkBold} & 88.60$\pm$0.02 & 86.00$\pm$0.26 & 79.20$\pm$0.08 & 55.02$\pm$0.02 & 87.99$\pm$0.02 & 89.82$\pm$0.05 & 93.77$\pm$0.14 \\
    \hline
    \color{red}{\XSolidBold} & \color{green}{\CheckmarkBold} & \color{green}{\CheckmarkBold} & \color{red}{\XSolidBold} & \color{red}{\XSolidBold} & 89.00$\pm$0.16 & 87.90$\pm$0.10 & 80.30$\pm$0.14 & 61.76$\pm$0.04 & 89.86$\pm$0.12 & 93.10$\pm$0.08 & 92.55$\pm$0.12 \\
    \color{red}{\XSolidBold} & \color{green}{\CheckmarkBold} & \color{green}{\CheckmarkBold} & \color{green}{\CheckmarkBold} & \color{red}{\XSolidBold} & 88.80$\pm$0.13 & 88.10$\pm$0.05 & 79.50$\pm$0.08 & 60.96$\pm$0.16 & 90.23$\pm$0.08 & 93.35$\pm$0.12 & 93.62$\pm$0.15 \\
    \color{red}{\XSolidBold} & \color{green}{\CheckmarkBold} & \color{red}{\XSolidBold} & \color{red}{\XSolidBold} & \color{green}{\CheckmarkBold} & 88.80$\pm$0.06 & 88.70$\pm$0.10 & 80.50$\pm$0.06 & 61.54$\pm$0.08 & 90.52$\pm$0.18 & 93.25$\pm$0.15 & 93.46$\pm$0.16 \\
    \bottomrule[1.2pt]
    \end{tabular}
    }
\end{table*}

\begin{figure}[!h]
    \centering
    \includegraphics[scale=0.20]{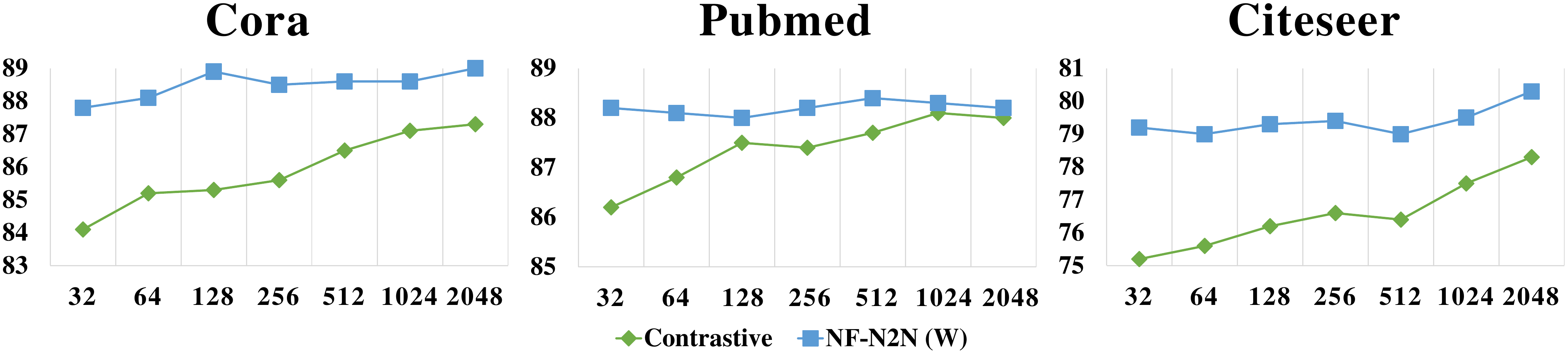}
    \caption{The performance comparison between Contrastive baseline and NF-N2N (W) based on varying number of batch sizes.}\label{fig:batch-size}
\end{figure}

\textbf{Batch Size.}~~In GCL, small batch size negatively influences the quality of the representations and thus deteriorates the classification performance on downstream tasks. To justify the advantage and robustness of our NF-N2N methods on varying batch sizes, we run experiments to compare Contrastive baseline and NF-N2N (W). The results are shown in Fig.~\ref{fig:batch-size}. From the table we can clearly observe that the performance of Contrastive baseline deteriorates when batch size becomes small, but our NF-N2N (W) can still maintain competitive performance under small batch size. This observation reassures us to use small batch size to alleviate GPU memory bottleneck.

\begin{figure}[!h]
    \centering
    \includegraphics[scale=0.20]{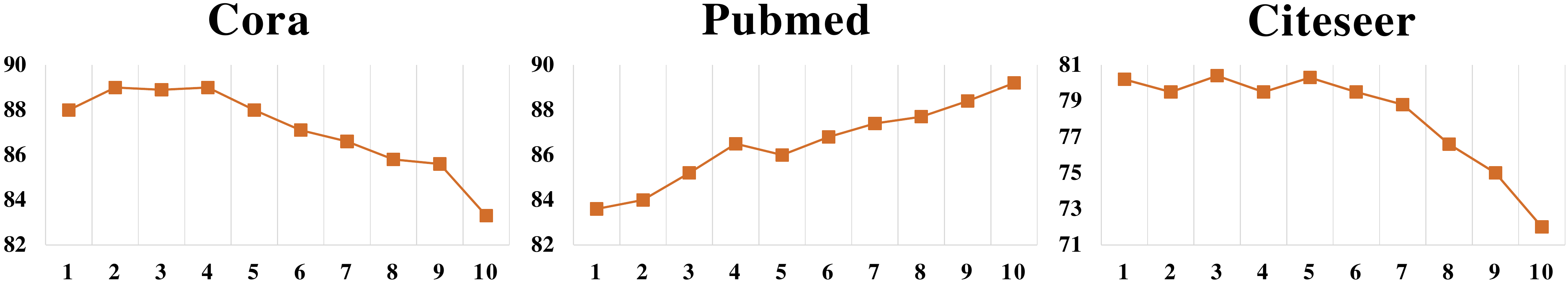}
    \caption{The node classification performance under the variation of the whitening iteration $P$ in NF-N2N (W) from $1$ to $10$.}\label{fig:iteration}
\end{figure}

\begin{figure}[!h]
    \centering
    \includegraphics[scale=0.20]{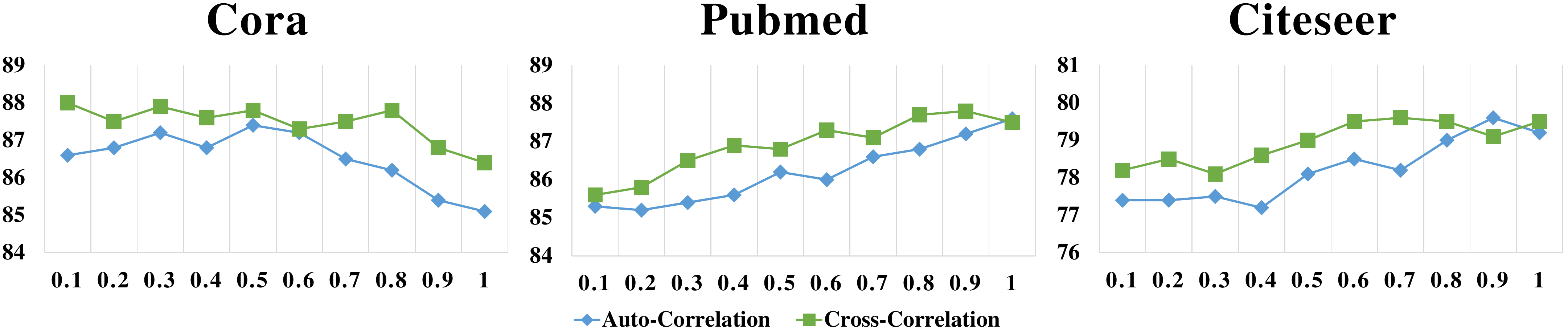}
    \caption{The node classification performance under the variation of the positive constant $\beta$ in auto- and cross-correlation regularizations from $0.1$ to $1.0$.}\label{fig:alpha}
\end{figure}

\textbf{Iteration $P$ in Whitening Normalization and Positive Constant $\beta$ in Auto- or Cross-Correlation Regularization.}~~Iteration $P$ in whitening normalization and positive constant $\beta$ in auto- or cross-correlation regularization are important hyper-parameters in the proposed decorrelation constraints because they control the graph signal decorrelation effect. Empirically, strong decorrelation leads to under-smoothing and weak decorrelation may fail to prevent over-smoothing. Fig.~\ref{fig:iteration} and Fig.~\ref{fig:alpha} show the trend of node classification performance under the variations of the three hyper-parameters. Generally, the trend of the curves are not consistent across different datasets. Even for the same curve, the trend is not monotonic. One possible explanation may be that the sweet spot in terms of graph smoothness can vary with datasets.

\vspace{-1em}
\section{Conclusions}\label{sec:conclusions}

This work presented two simple-yet-effective self-supervised node representation learning strategies (1) by directly optimizing the alignment between hidden representations of nodes and their neighbourhood through mutual information maximization; (2) by directly maximizing the similarities between representations of adjacency neighbouring nodes and proposed a Graph Signal Decorrelation (GSD) principle to avoid representation collapse. Theoretically, our formulation encourages graph smoothing. We also proposed a TAPS strategy to identify informative neighbours and improve the efficiency of our framework. It is worth mentioning when only one positive is selected, our model can fully avoid neighbourhood aggregation but still maintain promising node classification performance. One of the major contributions of this work is GSD principle. Under the umbrella thereof, different constraint implementations can be derived. We developed three versions of GSD implementations, termed whitening normalization, auto-, and cross-correlation regularization. Experiments on fourteen graph-based node classification datasets show the advantage of our methods.
\bibliographystyle{IEEEtran}
\vspace{-1em}
\bibliography{IEEEabrv, paper}

\begin{thebibliography}{10}
\providecommand{\url}[1]{#1}
\csname url@samestyle\endcsname
\providecommand{\newblock}{\relax}
\providecommand{\bibinfo}[2]{#2}
\providecommand{\BIBentrySTDinterwordspacing}{\spaceskip=0pt\relax}
\providecommand{\BIBentryALTinterwordstretchfactor}{4}
\providecommand{\BIBentryALTinterwordspacing}{\spaceskip=\fontdimen2\font plus
\BIBentryALTinterwordstretchfactor\fontdimen3\font minus
  \fontdimen4\font\relax}
\providecommand{\BIBforeignlanguage}[2]{{%
\expandafter\ifx\csname l@#1\endcsname\relax
\typeout{** WARNING: IEEEtran.bst: No hyphenation pattern has been}%
\typeout{** loaded for the language `#1'. Using the pattern for}%
\typeout{** the default language instead.}%
\else
\language=\csname l@#1\endcsname
\fi
#2}}
\providecommand{\BIBdecl}{\relax}
\BIBdecl

\bibitem{zhao2019multi}
Z.-M. Chen, X.-S. Wei, P.~Wang, and Y.~Guo, ``{Multi-label image recognition
  with graph convolutional networks},'' in \emph{Proceedings of the IEEE
  Conference on Computer Vision and Pattern Recognition}, 2019.

\bibitem{wu2021graph}
L.~Wu, Y.~Chen, K.~Shen, X.~Guo, H.~Gao, S.~Li, J.~Pei, and B.~Long, ``Graph
  neural networks for natural language processing: A survey,'' \emph{arXiv
  preprint arXiv:2106.06090}, 2021.

\bibitem{cui2018survey}
P.~Cui, X.~Wang, J.~Pei, and W.~Zhu, ``A survey on network embedding,''
  \emph{IEEE Transactions on Knowledge and Data Engineering}, vol.~31, no.~5,
  pp. 833--852, 2018.

\bibitem{li2020graph}
M.~Li, S.~Chen, Y.~Zhang, and I.~Tsang, ``Graph cross networks with vertex
  infomax pooling,'' in \emph{Advances in Neural Information Processing
  Systems}, vol.~33, 2020, pp. 14\,093--14\,105.

\bibitem{wieder2020compact}
O.~Wieder, S.~Kohlbacher, M.~Kuenemann, A.~Garon, P.~Ducrot, T.~Seidel, and
  T.~Langer, ``A compact review of molecular property prediction with graph
  neural networks,'' \emph{Drug Discovery Today: Technologies}, 2020.

\bibitem{hamilton2017inductive}
W.~L. Hamilton, R.~Ying, and J.~Leskovec, ``Inductive representation learning
  on large graphs,'' in \emph{Proceedings of the 31st International Conference
  on Neural Information Processing Systems}, 2017, pp. 1025--1035.

\bibitem{velickovic2019graph}
P.~Veli{\v{c}}kovi{\'c}, G.~Cucurull, A.~Casanova, A.~Romero, P.~Lio, and
  Y.~Bengio, ``Graph attention networks,'' in \emph{International Conference on
  Learning Representations}, 2019.

\bibitem{corso2020principal}
G.~Corso, L.~Cavalleri, D.~Beaini, P.~Li{\`o}, and P.~Veli{\v{c}}kovi{\'c},
  ``Principal neighbourhood aggregation for graph nets,'' \emph{Advances in
  Neural Information Processing Systems}, vol.~33, 2020.

\bibitem{multiview}
Y.~Tian, D.~Krishnan, and P.~Isola, ``Contrastive multiview coding,'' in
  \emph{European Conference on Computer Vision}.\hskip 1em plus 0.5em minus
  0.4em\relax Springer, 2020, pp. 776--794.

\bibitem{chen2020simple}
T.~Chen, S.~Kornblith, M.~Norouzi, and G.~Hinton, ``A simple framework for
  contrastive learning of visual representations,'' in \emph{International
  conference on machine learning}.\hskip 1em plus 0.5em minus 0.4em\relax PMLR,
  2020, pp. 1597--1607.

\bibitem{devlin2019bert}
J.~Devlin, M.-W. Chang, K.~Lee, and K.~Toutanova, ``Bert: Pre-training of deep
  bidirectional transformers for language understanding,'' in \emph{Proceedings
  of the 2019 Conference of the North American Chapter of the Association for
  Computational Linguistics: Human Language Technologies, Volume 1 (Long and
  Short Papers)}, 2019, pp. 4171--4186.

\bibitem{liu2022graph}
Y.~Liu, M.~Jin, S.~Pan, C.~Zhou, Y.~Zheng, F.~Xia, and P.~Yu, ``Graph
  self-supervised learning: A survey,'' \emph{IEEE Transactions on Knowledge
  and Data Engineering}, 2022.

\bibitem{park2019relational}
W.~Park, D.~Kim, Y.~Lu, and M.~Cho, ``Relational knowledge distillation,'' in
  \emph{Proceedings of the IEEE/CVF Conference on Computer Vision and Pattern
  Recognition}, 2019, pp. 3967--3976.

\bibitem{cho2019efficacy}
J.~H. Cho and B.~Hariharan, ``On the efficacy of knowledge distillation,'' in
  \emph{Proceedings of the IEEE/CVF International Conference on Computer
  Vision}, 2019, pp. 4794--4802.

\bibitem{oord2018representation}
A.~v.~d. Oord, Y.~Li, and O.~Vinyals, ``Representation learning with
  contrastive predictive coding,'' \emph{arXiv preprint arXiv:1807.03748},
  2018.

\bibitem{zbontar2021barlow}
J.~Zbontar, L.~Jing, I.~Misra, Y.~LeCun, and S.~Deny, ``Barlow twins:
  Self-supervised learning via redundancy reduction,'' in \emph{International
  Conference on Machine Learning}.\hskip 1em plus 0.5em minus 0.4em\relax PMLR,
  2021, pp. 12\,310--12\,320.

\bibitem{kipf2017semi}
T.~Kipf and M.~Welling, ``Semi-supervised classification with graph
  convolutional networks,'' in \emph{International Conference on Learning
  Representations}, 2017.

\bibitem{xu2019how}
K.~Xu, W.~Hu, J.~Leskovec, and S.~Jegelka, ``How powerful are graph neural
  networks,'' in \emph{International Conference on Learning Representations},
  2019.

\bibitem{yue2022survey}
L.~Yue, X.~Jun, Z.~Sihang, W.~Siwei, G.~Xifeng, Y.~Xihong, L.~Ke, T.~Wenxuan,
  L.~X. Wang \emph{et~al.}, ``A survey of deep graph clustering: Taxonomy,
  challenge, and application,'' \emph{arXiv preprint arXiv:2211.12875}, 2022.

\bibitem{dong2022node}
W.~Dong, J.~Wu, Y.~Luo, Z.~Ge, and P.~Wang, ``Node representation learning in
  graph via node-to-neighbourhood mutual information maximization,'' in
  \emph{Proceedings of the IEEE/CVF Conference on Computer Vision and Pattern
  Recognition}, 2022, pp. 16\,620--16\,629.

\bibitem{wu2019comprehensive}
Z.~Wu, S.~Pan, F.~Chen, G.~Long, C.~Zhang, and S.~Y. Philip, ``A comprehensive
  survey on graph neural networks,'' \emph{IEEE Transactions on Neural Networks
  and Learning Systems}, 2020.

\bibitem{scarselli2008graph}
F.~Scarselli, M.~Gori, A.~C. Tsoi, M.~Hagenbuchner, and G.~Monfardini, ``The
  graph neural network model,'' \emph{IEEE transactions on neural networks},
  vol.~20, no.~1, pp. 61--80, 2008.

\bibitem{defferrard2016convolutional}
M.~Defferrard, X.~Bresson, and P.~Vandergheynst, ``Convolutional neural
  networks on graphs with fast localized spectral filtering,'' in
  \emph{Proceedings of the 30th International Conference on Neural Information
  Processing Systems}, 2016, pp. 3844--3852.

\bibitem{wu2019simplifying}
F.~Wu, A.~Souza, T.~Zhang, C.~Fifty, T.~Yu, and K.~Weinberger, ``Simplifying
  graph convolutional networks,'' in \emph{International Conference on Machine
  Learning}, 2019, pp. 6861--6871.

\bibitem{hou2019measuring}
Y.~Hou, J.~Zhang, J.~Cheng, K.~Ma, R.~T. Ma, H.~Chen, and M.-C. Yang,
  ``Measuring and improving the use of graph information in graph neural
  networks,'' in \emph{International Conference on Learning Representations},
  2019.

\bibitem{peng2020graph}
Z.~Peng, W.~Huang, M.~Luo, Q.~Zheng, Y.~Rong, T.~Xu, and J.~Huang, ``Graph
  representation learning via graphical mutual information maximization,'' in
  \emph{Proceedings of The Web Conference 2020}, 2020, pp. 259--270.

\bibitem{velivckovic2018deep}
P.~Veli{\v{c}}kovi{\'c}, W.~Fedus, W.~L. Hamilton, P.~Li{\`o}, Y.~Bengio, and
  R.~D. Hjelm, ``Deep graph infomax,'' in \emph{International Conference on
  Learning Representations}, 2018.

\bibitem{hjelm2018learning}
R.~D. Hjelm, A.~Fedorov, S.~Lavoie-Marchildon, K.~Grewal, P.~Bachman,
  A.~Trischler, and Y.~Bengio, ``Learning deep representations by mutual
  information estimation and maximization,'' in \emph{International Conference
  on Learning Representations}, 2018.

\bibitem{hassani2020contrastive}
K.~Hassani and A.~H. Khasahmadi, ``Contrastive multi-view representation
  learning on graphs,'' in \emph{International Conference on Machine
  Learning}.\hskip 1em plus 0.5em minus 0.4em\relax PMLR, 2020, pp. 4116--4126.

\bibitem{sun2020infograph}
F.-Y. Sun, J.~Hoffman, V.~Verma, and J.~Tang, ``Infograph: Unsupervised and
  semi-supervised graph-level representation learning via mutual information
  maximization,'' in \emph{International Conference on Learning
  Representations}, 2020.

\bibitem{zhu2021graph}
Y.~Zhu, Y.~Xu, F.~Yu, Q.~Liu, S.~Wu, and L.~Wang, ``Graph contrastive learning
  with adaptive augmentation,'' in \emph{Proceedings of the Web Conference
  2021}, 2021, pp. 2069--2080.

\bibitem{zhu2020deep}
Y.~Zhu, Y.~Xu, F.~Yu, Q.~Liu, and S.~Wu, ``Deep graph contrastive
  representation learning,'' \emph{arXiv preprint arXiv:2006.04131}, 2020.

\bibitem{wang2022augmentation}
H.~Wang, J.~Zhang, Q.~Zhu, and W.~Huang, ``Augmentation-free graph contrastive
  learning,'' \emph{arXiv preprint arXiv:2204.04874}, 2022.

\bibitem{hu2021graph}
Y.~Hu, H.~You, Z.~Wang, Z.~Wang, E.~Zhou, and Y.~Gao, ``Graph-mlp: Node
  classification without message passing in graph,'' \emph{arXiv preprint
  arXiv:2106.04051}, 2021.

\bibitem{liu2022simple}
Y.~Liu, X.~Yang, S.~Zhou, and X.~Liu, ``Simple contrastive graph clustering,''
  \emph{arXiv preprint arXiv:2205.07865}, 2022.

\bibitem{grill2020bootstrap}
J.-B. Grill, F.~Strub, F.~Altch{\'e}, C.~Tallec, P.~Richemond, E.~Buchatskaya,
  C.~Doersch, B.~Avila~Pires, Z.~Guo, M.~Gheshlaghi~Azar \emph{et~al.},
  ``Bootstrap your own latent-a new approach to self-supervised learning,''
  \emph{Advances in Neural Information Processing Systems}, vol.~33, pp.
  21\,271--21\,284, 2020.

\bibitem{he2020momentum}
K.~He, H.~Fan, Y.~Wu, S.~Xie, and R.~Girshick, ``Momentum contrast for
  unsupervised visual representation learning,'' in \emph{Proceedings of the
  IEEE/CVF Conference on Computer Vision and Pattern Recognition}, 2020, pp.
  9729--9738.

\bibitem{chen2021exploring}
X.~Chen and K.~He, ``Exploring simple siamese representation learning,'' in
  \emph{Proceedings of the IEEE/CVF Conference on Computer Vision and Pattern
  Recognition}, 2021, pp. 15\,750--15\,758.

\bibitem{bardes2022vicreg}
A.~Bardes, J.~Ponce, and Y.~Lecun, ``Vicreg: Variance-invariance-covariance
  regularization for self-supervised learning,'' in \emph{International
  Conference on Learning Representations}, 2022.

\bibitem{thakoor2021bootstrapped}
S.~Thakoor, C.~Tallec, M.~G. Azar, M.~Azabou, E.~L. Dyer, R.~Munos,
  P.~Veli{\v{c}}kovi{\'c}, and M.~Valko, ``Large-scale representation learning
  on graphs via bootstrapping,'' in \emph{International Conference on Learning
  Representations}, 2021.

\bibitem{kefato2021self}
Z.~T. Kefato and S.~Girdzijauskas, ``Self-supervised graph neural networks
  without explicit negative sampling,'' in \emph{The International Workshop on
  Self-Supervised Learning for the Web (SSL'21), at WWW'21}, 2021.

\bibitem{lee2021augmentation}
N.~Lee, J.~Lee, and C.~Park, ``Augmentation-free self-supervised learning on
  graphs,'' \emph{Association for the Advancement of Artificial Intelligence},
  2022.

\bibitem{shuman2013emerging}
D.~I. Shuman, S.~K. Narang, P.~Frossard, A.~Ortega, and P.~Vandergheynst, ``The
  emerging field of signal processing on graphs: Extending high-dimensional
  data analysis to networks and other irregular domains,'' \emph{IEEE signal
  processing magazine}, vol.~30, no.~3, pp. 83--98, 2013.

\bibitem{belghazi2018mutual}
M.~I. Belghazi, A.~Baratin, S.~Rajeshwar, S.~Ozair, Y.~Bengio, A.~Courville,
  and D.~Hjelm, ``Mutual information neural estimation,'' in
  \emph{International Conference on Machine Learning}.\hskip 1em plus 0.5em
  minus 0.4em\relax PMLR, 2018, pp. 531--540.

\bibitem{li2018deeper}
Q.~Li, Z.~Han, and X.-M. Wu, ``Deeper insights into graph convolutional
  networks for semi-supervised learning,'' in \emph{Thirty-Second AAAI
  Conference on Artificial Intelligence}, 2018.

\bibitem{xu2018representation}
K.~Xu, C.~Li, Y.~Tian, T.~Sonobe, K.-i. Kawarabayashi, and S.~Jegelka,
  ``Representation learning on graphs with jumping knowledge networks,'' in
  \emph{International Conference on Machine Learning}.\hskip 1em plus 0.5em
  minus 0.4em\relax PMLR, 2018, pp. 5453--5462.

\bibitem{ermolov2021whitening}
A.~Ermolov, A.~Siarohin, E.~Sangineto, and N.~Sebe, ``Whitening for
  self-supervised representation learning,'' in \emph{International Conference
  on Machine Learning}.\hskip 1em plus 0.5em minus 0.4em\relax PMLR, 2021, pp.
  3015--3024.

\bibitem{ioffe2015batch}
S.~Ioffe and C.~Szegedy, ``Batch normalization: Accelerating deep network
  training by reducing internal covariate shift,'' in \emph{International
  conference on machine learning}.\hskip 1em plus 0.5em minus 0.4em\relax PMLR,
  2015, pp. 448--456.

\bibitem{huang2018decorrelated}
L.~Huang, D.~Yang, B.~Lang, and J.~Deng, ``Decorrelated batch normalization,''
  in \emph{Proceedings of the IEEE Conference on Computer Vision and Pattern
  Recognition}, 2018, pp. 791--800.

\bibitem{huang2019iterative}
L.~Huang, Y.~Zhou, F.~Zhu, L.~Liu, and L.~Shao, ``Iterative normalization:
  Beyond standardization towards efficient whitening,'' in \emph{Proceedings of
  the IEEE/CVF Conference on Computer Vision and Pattern Recognition}, 2019,
  pp. 4874--4883.

\bibitem{guo2010newton}
C.-H. Guo, ``On newton’s method and halley’s method for the principal pth
  root of a matrix,'' \emph{Linear algebra and its applications}, vol. 432,
  no.~8, pp. 1905--1922, 2010.

\bibitem{fey2018splinecnn}
M.~Fey, J.~E. Lenssen, F.~Weichert, and H.~M{\"u}ller, ``Splinecnn: Fast
  geometric deep learning with continuous b-spline kernels,'' in
  \emph{Proceedings of the IEEE/CVF Conference on Computer Vision and Pattern
  Recognition}, 2018, pp. 869--877.

\bibitem{shchur2018pitfalls}
O.~Shchur, M.~Mumme, A.~Bojchevski, and S.~G{\"u}nnemann, ``Pitfalls of graph
  neural network evaluation,'' \emph{arXiv preprint arXiv:1811.05868}, 2018.

\bibitem{rozemberczki2020feather}
B.~Rozemberczki and R.~Sarkar, ``Characteristic functions on graphs: Birds of a
  feather, from statistical descriptors to parametric models,'' in
  \emph{Proceedings of the 29th ACM International Conference on Information and
  Knowledge Management (CIKM '20)}.\hskip 1em plus 0.5em minus 0.4em\relax ACM,
  2020, p. 1325–1334.

\bibitem{rozemberczki2019multiscale}
B.~Rozemberczki, C.~Allen, and R.~Sarkar, ``Multi-scale attributed node
  embedding,'' 2019.

\bibitem{rozemberczki2021twitch}
B.~Rozemberczki and R.~Sarkar, ``Twitch gamers: A dataset for evaluating
  proximity preserving and structural role-based node embeddings,'' 2021.

\bibitem{yang2012defining}
J.~Yang and J.~Leskovec, ``Defining and evaluating network communities based on
  ground-truth,'' in \emph{Proceedings of the ACM SIGKDD Workshop on Mining
  Data Semantics}, 2012, pp. 1--8.

\bibitem{perozzi2014deepwalk}
B.~Perozzi, R.~Al-Rfou, and S.~Skiena, ``Deepwalk: Online learning of social
  representations,'' in \emph{Proceedings of the 20th ACM SIGKDD international
  conference on Knowledge discovery and data mining}, 2014, pp. 701--710.

\bibitem{grover2016node2vec}
A.~Grover and J.~Leskovec, ``node2vec: Scalable feature learning for
  networks,'' in \emph{Proceedings of the 22nd ACM SIGKDD international
  conference on Knowledge discovery and data mining}, 2016, pp. 855--864.

\bibitem{chen2018fastgcn}
J.~Chen, T.~Ma, and C.~Xiao, ``Fastgcn: Fast learning with graph convolutional
  networks via importance sampling,'' in \emph{International Conference on
  Learning Representations}, 2018.

\bibitem{dong2022improving}
W.~Dong, J.~Wu, X.~Zhang, Z.~Bai, P.~Wang, and M.~Wo{\'z}niak, ``Improving
  performance and efficiency of graph neural networks by injective
  aggregation,'' \emph{Knowledge-Based Systems}, vol. 254, p. 109616, 2022.

\end{thebibliography}

\clearpage





\setcounter{page}{1}

\appendices

\section{Proof of Theorem~\ref{theo:kl-mut}}\label{appe:theo:kl-mut}

By virtue of the relationship between mutual information and information entropy, we obtain:

\begin{equation}\label{eq:mutual-entropy}
\begin{split}
    &I(S(\bm{x})^{(l)};H(\bm{x})^{(l)})= \\
    &{\rm H}(S(\bm{x})^{(l)})+{\rm H}(H(\bm{x})^{(l)})-{\rm H}(S(\bm{x})^{(l)},H(\bm{x})^{(l)}), \\
\end{split}
\end{equation}

\noindent where ${\rm H}(\cdot)$ is the information entropy and ${\rm H}(\cdot,\cdot)$ is the joint entropy. And the information gain or Kullback-Leibler divergence with information entropy is defined as:

\begin{equation}\label{eq:KL-entropy}
\begin{split}
    &D_{KL}(S(\bm{x})^{(l)}\|H(\bm{x})^{(l)})= \\
    &{\rm H}(S(\bm{x})^{(l)},H(\bm{x})^{(l)})-{\rm H}(S(\bm{x})^{(l)}). \\
\end{split}
\end{equation}

\noindent Combining Eq.\emph{~(\ref{eq:mutual-entropy})} with\emph{~(\ref{eq:KL-entropy})}, we get:

\begin{equation}\label{eq:mutual-KL-entropy}
\begin{split}
    &I(S(\bm{x})^{(l)};H(\bm{x})^{(l)})={\rm H}(S(\bm{x})^{(l)})+{\rm H}(H(\bm{x})^{(l)}) \\
    &-D_{KL}(S(\bm{x})^{(l)}\|H(\bm{x})^{(l)})-{\rm H}(S(\bm{x})^{(l)}) \\
    &={\rm H}(H(\bm{x})^{(l)})-D_{KL}(S(\bm{x})^{(l)}\|H(\bm{x})^{(l)}). \\
\end{split}
\end{equation}

\noindent In terms of Eq.\emph{~(\ref{eq:mutual-KL-entropy})}, we observe that $I(S(\bm{x})^{(l)};H(\bm{x})^{(l)})$ is negatively correlated with $D_{KL}(S(\bm{x})^{(l)}\|H(\bm{x})^{(l)})$. Associated with $D_{KL}(S(\bm{x})^{(l)}\|H(\bm{x})^{(l)})\sim \delta_{f}^{(l)}$\emph{~\cite{hou2019measuring}}, maximizing $I(S(\bm{x})^{(l)};H(\bm{x})^{(l)})$ essentially minimizes $D_{KL}(S(\bm{x})^{(l)}\|H(\bm{x})^{(l)})$ and $\delta_{f}^{(l)}$, which attains the goal of graph smoothing:

\begin{equation}
\begin{split}
    I(S(\bm{x})^{(l)};H(\bm{x})^{(l)})&\sim \frac{1}{D_{KL}(S(\bm{x})^{(l)}\|H(\bm{x})^{(l)})} \\
    &\sim \frac{1}{\delta_{f}^{(l)}}. \\
\end{split}
\end{equation}

\section{Proof of Theorem~\ref{theo:info-MSE}}\label{app:theo:info-MSE}

Minimizing the MSE loss function in Eq.~(\ref{eq:MSE-GTV}) requires to set derivative of $\vec{\boldsymbol s}_{d^{(l)}}^{{\rm T}} \cdot \mathbf{L} \cdot \vec{\boldsymbol s}_{d^{(l)}}$ with respect to $\vec{\boldsymbol s}_{d^{(l)}}$ to zero:

\begin{equation}\label{eq:derivative-MSE}
\begin{split}
    \frac{\partial \vec{\boldsymbol s}_{d^{(l)}}^{{\rm T}} \cdot \mathbf{L} \cdot \vec{\boldsymbol s}_{d^{(l)}}}{\partial \vec{\boldsymbol s}_{d^{(l)}}}&=0, \\
    2 \cdot \mathbf{L} \cdot \vec{\boldsymbol s}_{d^{(l)}}&=0, \\
    (\mathbf{D}-\mathbf{A}) \cdot \vec{\boldsymbol s}_{d^{(l)}}&=0, \\
    \mathbf{D} \cdot \vec{\boldsymbol s}_{d^{(l)}} &= \mathbf{A} \cdot \vec{\boldsymbol s}_{d^{(l)}}, \\
    \vec{\boldsymbol s}_{d^{(l)}} &= \mathbf{D}^{-1}\mathbf{A} \cdot \vec{\boldsymbol s}_{d^{(l)}}, \\
\end{split}
\end{equation}

\noindent in terms of the symmetric property of Graph Laplacian matrix $\mathbf{L}$. Note that $\mathbf{D}^{-1}\mathbf{A}$ is the \emph{mean} aggregator~\cite{hamilton2017inductive}. Eq.~(\ref{eq:derivative-MSE}) can be explained as an limit distribution where $\vec{\boldsymbol s}_{d^{(l)}}^{({\rm lim})} = \mathbf{D}^{-1}\mathbf{A} \cdot \vec{\boldsymbol s}_{d^{(l)}}^{({\rm lim})}$. Then we use the following iterative form to approximate the limit $\vec{\boldsymbol s}_{d^{(l)}}^{({\rm lim})}$ with $K \to +\infty$:

\begin{equation}\label{eq:iterative-derivative-MSE}
    \vec{\boldsymbol s}_{d^{(l)}}^{(K)} = \mathbf{D}^{-1}\mathbf{A} \cdot \vec{\boldsymbol s}_{d^{(l)}}^{(K-1)}.
\end{equation}

\noindent To satisfy Eq.~(\ref{eq:iterative-derivative-MSE}), the limit $\vec{\boldsymbol s}_{d^{(l)}}^{({\rm lim})}$ must be $\hat{\sigma} \cdot \vec{\boldsymbol 1}$ with a scalar $\hat{\sigma}$. The work~\cite{dong2022improving} proposed that the maximum eigenvalue of the right stochastic square matrix $\mathbf{D}^{-1}\mathbf{A}$ is $1$ with each row summing to $1$, implying the \emph{mean} aggregator as a low-pass filter that iteratively minimizes the feature smoothness metric $\delta_{f}^{(l)}$ by leveraging Eq.~(\ref{eq:iterative-derivative-MSE}). The work~\cite{hou2019measuring} has also proved running this aggregator on graphs satisfies $D_{KL}(S(\bm{x})^{(l)}\|H(\bm{x})^{(l)}) \sim \delta_{f}^{(l)}$, hence, minimizing $\mathcal{L}_{{\rm MSE}}^{(l)}$ also minimizes $D_{KL}(S(\bm{x})^{(l)}\|H(\bm{x})^{(l)})$, \emph{i.e.}:

\begin{equation}\label{eq:KL-smoothness-2}
    \mathcal{L}_{{\rm MSE}}^{(l)} \sim D_{KL}(S(\bm{x})^{(l)}\|H(\bm{x})^{(l)}) \sim \delta_{f}^{(l)}.
\end{equation}

\noindent Associated with Theorem~\ref{theo:kl-mut}, the MSE loss $\mathcal{L}_{{\rm MSE}}^{(l)}$ minimization essentially minimizes $D_{KL}(S(\bm{x})^{(l)}\|H(\bm{x})^{(l)})$ and maximizes $I(S(\bm{x})^{(l)};H(\bm{x})^{(l)})$, which attains the goal of graph smoothing:

\begin{equation}
\begin{split}
    \mathcal{L}_{{\rm MSE}}^{(l)}({\rm GTV}) &\sim D_{KL}(S(\bm{x})^{(l)}\|H(\bm{x})^{(l)}) \\
    &\sim \frac{1}{I(S(\bm{x})^{(l)};H(\bm{x})^{(l)})}. \\
\end{split}
\end{equation}

\section{Proof of Theorem~\ref{theo:graph-signal-decoorelation}}\label{app:theo:graph-signal-decoorelation}

We denote a graph signal $\vec{\boldsymbol s}_{i}$ represented by the linear combination of orthonormal eigenvectors $\mathbf{V}=[\vec{\boldsymbol v}_{1},\cdots,\vec{\boldsymbol v}_{N}]$ of Graph Laplacian matrix $\mathbf{L}$, \emph{i.e.}:

\begin{equation}\label{eq:graph-signal-eigenvector}
    \vec{\boldsymbol s}_{i}=\sigma_{1}^{i} \cdot \vec{\boldsymbol v}_{1} + \cdots + \sigma_{N}^{i} \cdot \vec{\boldsymbol v}_{N}=\mathbf{V} \cdot \vec{\boldsymbol \sigma}_{i},
\end{equation}

\noindent where $0\leq i \leq D^{(l)}$ and the coefficient vector is $\vec{\boldsymbol \sigma}_{i}=[\sigma_{1}^{i},\cdots,\sigma_{N}^{i}]^{{\rm T}}$. By virtue of Eq.~(\ref{eq:graph-signal-eigenvector}), GTV for graph signal $\vec{\boldsymbol s}_{i}$ is:

\begin{equation}\label{eq:GTV-graph-signal-eigenvector}
\begin{split}
    \vec{\boldsymbol s}_{i}^{{\rm T}} \cdot \mathbf{L} \cdot \vec{\boldsymbol s}_{i} =& \vec{\boldsymbol s}_{i}^{{\rm T}} \cdot \mathbf{L} \cdot (\sigma_{1}^{i} \cdot \vec{\boldsymbol v}_{1} + \cdots + \sigma_{N}^{i} \cdot \vec{\boldsymbol v}_{N}) \\
     =& \vec{\boldsymbol s}_{i}^{{\rm T}} \cdot (\sigma_{1}^{i} \cdot \mathbf{L} \cdot \vec{\boldsymbol v}_{1} + \cdots + \sigma_{N}^{i} \cdot \mathbf{L} \cdot \vec{\boldsymbol v}_{N}) \\
     =& \vec{\boldsymbol s}_{i}^{{\rm T}} \cdot (\sigma_{1}^{i} \cdot \lambda_{1} \cdot \vec{\boldsymbol v}_{1} + \cdots + \sigma_{N}^{i} \cdot \lambda_{N} \cdot \vec{\boldsymbol v}_{N}) \\
     =& (\sigma_{1}^{i} \cdot \vec{\boldsymbol v}_{1}^{{\rm T}} + \cdots + \sigma_{N}^{i} \cdot \vec{\boldsymbol v}_{N}^{{\rm T}}) \cdot \\
     &(\sigma_{1}^{i} \cdot \lambda_{1} \cdot \vec{\boldsymbol v}_{1} + \cdots + \sigma_{N}^{i} \cdot \lambda_{N} \cdot \vec{\boldsymbol v}_{N}) \\
     =& (\sigma_{1}^{i})^{2} \cdot \lambda_{1} \cdot \vec{\boldsymbol v}_{1}^{{\rm T}} \cdot \vec{\boldsymbol v}_{1} + \cdots \\
     &+ (\sigma_{N}^{i})^{2} \cdot \lambda_{N} \cdot \vec{\boldsymbol v}_{N}^{{\rm T}} \cdot \vec{\boldsymbol v}_{N} \\
     =& (\sigma_{1}^{i})^{2} \cdot \lambda_{1} + \cdots + (\sigma_{N}^{i})^{2} \cdot \lambda_{N}, \\
\end{split}
\end{equation}

\noindent where $\vec{\boldsymbol v}_{n}^{{\rm T}} \cdot \vec{\boldsymbol v}_{m}=1$ when $n=m$ and $\vec{\boldsymbol v}_{n}^{{\rm T}} \cdot \vec{\boldsymbol v}_{m}=0$ if $n \neq m$, with $1 \leq n \leq N$ and $1 \leq m \leq N$.

On the other hand, when any two graph signals are orthogonal to each other, \emph{i.e.}, $\vec{\boldsymbol s}_{i} \perp \vec{\boldsymbol s}_{j}\ s.t.\ i \neq j$ with $0 \leq i \leq D^{(l)}$ and $0 \leq j \leq D^{(l)}$, we obtain:

\begin{equation}\label{eq:orthogonal-coefficient}
\begin{split}
    \vec{\boldsymbol s}_{i}^{{\rm T}} \cdot \vec{\boldsymbol s}_{j} =& (\mathbf{V} \cdot \vec{\boldsymbol \sigma}_{i})^{{\rm T}} \mathbf{V} \cdot \vec{\boldsymbol \sigma}_{j} \\
    =& \vec{\boldsymbol \sigma}_{i}^{{\rm T}} \cdot \mathbf{V}^{{\rm T}}\mathbf{V} \cdot \vec{\boldsymbol \sigma}_{j} \\
    =& \vec{\boldsymbol \sigma}_{i}^{{\rm T}} \cdot \vec{\boldsymbol \sigma}_{j} \\
    =& 0,
\end{split}
\end{equation}

\noindent implying any two coefficient vectors being orthogonal to each other, \emph{i.e.}, $\vec{\boldsymbol \sigma}_{i} \perp \vec{\boldsymbol \sigma}_{j}\ s.t.\ i \neq j$.

When MSE minimization makes any graph signal, for example $\vec{\boldsymbol s}_{1}$, equal to $\sigma_{1} \cdot \vec{\boldsymbol v}_{1}$ resulting in $\vec{\boldsymbol s}_{1}^{{\rm T}} \cdot \mathbf{L} \cdot \vec{\boldsymbol s}_{1}=0$ in terms of Eq.~(\ref{eq:eigenvalues-eigenvectors}), the coefficient vector $\vec{\boldsymbol \sigma}_{1}=[\sigma_{1}^{1}, 0, \cdots, 0]$ makes $\vec{\boldsymbol s}_{1}^{{\rm T}} \cdot \mathbf{L} \cdot \vec{\boldsymbol s}_{1}=(\sigma_{1}^{1})^{2} \cdot \lambda_{1}$ with $\sigma_{1}^{1}\neq 0$. Assuming $i,j\in [1,\cdots, D^{(l)}]$ and $D^{(l)} < N$, we get:

\begin{equation}\label{eq:each-GTV}
\begin{split}
    \vec{\boldsymbol s}_{1}^{{\rm T}} \cdot \mathbf{L} \cdot \vec{\boldsymbol s}_{1}&=(\sigma_{1}^{1})^{2} \cdot \lambda_{1}, \\
    &\cdots, \\
    \vec{\boldsymbol s}_{D^{(l)}}^{{\rm T}} \cdot \mathbf{L} \cdot \vec{\boldsymbol s}_{D^{(l)}}&=(\sigma_{D^{(l)}}^{D^{(l)}})^{2} \cdot \lambda_{D^{(l)}}, \\
\end{split}
\end{equation}

\noindent under $\vec{\boldsymbol s}_{i} \perp \vec{\boldsymbol s}_{j}$ and $\vec{\boldsymbol \sigma}_{i} \perp \vec{\boldsymbol \sigma}_{j}\ s.t.\ i \neq j$. Hence, the GTV is:

\begin{equation}\label{eq:decorrelation-GTV-proof}
\begin{split}
    \sum_{i=1}^{D^{(l)}}\vec{\boldsymbol s}_{i}^{{\rm T}} \cdot \mathbf{L} \cdot \vec{\boldsymbol s}_{i}&=(\sigma_{1}^{1})^{2} \cdot \lambda_{1} + \cdots + (\sigma_{D^{(l)}}^{D^{(l)}})^{2} \cdot \lambda_{D^{(l)}} \\
    &=\sigma_{1}^{2} \cdot \lambda_{1} + \cdots + \sigma_{D^{(l)}}^{2} \cdot \lambda_{D^{(l)}} > 0. \\
\end{split}
\end{equation}

\section{Pseudocodes of Three Graph Signal Decorrelations}\label{app:pseudocodes}

\begin{algorithm}[!h]
    \caption{The pseudocode for whitening normalization.}
    \LinesNumbered
    \KwIn{Node representations $\mathbf{H}^{(l)}$, iteration $P$.}
    \KwOut{ZCA node representations $\mathbf{H}_{{\rm ZCA}}^{(l)}$.}
    $\vec{\boldsymbol \mu}^{{\rm T}} \gets \frac{1}{N}\cdot \vec{\boldsymbol 1}^{{\rm T}} \cdot \mathbf{H}^{(l)}$\;
    $\widehat{\mathbf{H}}^{(l)} \gets \mathbf{H}^{(l)}-\vec{\boldsymbol 1} \cdot \vec{\boldsymbol \mu}^{{\rm T}}$\;
    $\mathbf{C}_{{\rm auto}}^{(l)} \gets \frac{1}{N}\widehat{\mathbf{H}}^{(l){\rm T}}\widehat{\mathbf{H}}^{(l)} + \epsilon \cdot \mathbf{I}$\;
    $\mathbf{P}_{0} \gets \mathbf{I}$, $\mathbf{N}_{0} \gets \frac{\mathbf{C}_{{\rm auto}}^{(l)}}{{\rm tr}(\mathbf{C}_{{\rm auto}}^{(l)})}$\;
    \For{$1\leq p \leq P$}{
        $\mathbf{P}_{p} \gets \mathbf{P}_{p-1}\left(\frac{3\cdot\mathbf{I}-\mathbf{N}_{p-1}}{2}\right)$\;
        $\mathbf{N}_{p} \gets \left(\frac{3\cdot\mathbf{I}-\mathbf{N}_{p-1}}{2}\right)^{2}\mathbf{N}_{p-1}$\;
    }
    $\mathbf{H}_{{\rm ZCA}}^{(l)} \gets \frac{\mathbf{P}_{p}}{{\rm tr}(\mathbf{C}_{{\rm auto}}^{(l)})}$\;
    \Return $\mathbf{H}_{{\rm ZCA}}^{(l)}$\;
\end{algorithm}

\begin{algorithm}[!h]
    \caption{The pseudocode for auto-correlation regularization.}
    \LinesNumbered
    \KwIn{Node representations $\mathbf{H}^{(l)}$, positive constant $\beta$.}
    \KwOut{Auto-correlation regularization $\mathcal{L}_{{\rm auto-reg}}^{(l)}$.}
    $\vec{\boldsymbol \mu}^{{\rm T}} \gets \frac{1}{N}\cdot \vec{\boldsymbol 1}^{{\rm T}} \cdot \mathbf{H}^{(l)}$\;
    $\widehat{\mathbf{H}}^{(l)} \gets \mathbf{H}^{(l)}-\vec{\boldsymbol 1} \cdot \vec{\boldsymbol \mu}^{{\rm T}}$\;
    $\mathbf{C}_{{\rm auto}}^{(l)} \gets \frac{1}{N}\widehat{\mathbf{H}}^{(l){\rm T}}\widehat{\mathbf{H}}^{(l)} + \epsilon \cdot \mathbf{I}$\;
    $\mathcal{L}_{{\rm auto-reg}}^{(l)} \gets \sum_{i}\left(1-\mathbf{C}_{{\rm auto}[i,i]}^{(l)}\right)^{2} + \beta \cdot \sum_{i}\sum_{j \neq i}\mathbf{C}_{{\rm auto}[i,j]}^{(l)2}$\;
    \Return $\mathcal{L}_{{\rm auto-reg}}^{(l)}$\;
\end{algorithm}

\begin{algorithm}[!h]
    \caption{The pseudocode for cross-correlation regularization.}
    \LinesNumbered
    \KwIn{Anchor node representations $\mathbf{H}^{{\rm anchor}(l)}$, view node representations $\mathbf{H}^{{\rm view}(l)}$, positive constant $\beta$.}
    \KwOut{Cross-correlation regularization $\mathcal{L}_{{\rm cross-reg}}^{(l)}$.}
    $\vec{\boldsymbol \mu}^{{\rm anchor}\;{\rm T}} \gets \frac{1}{N}\cdot \vec{\boldsymbol 1}^{{\rm T}} \cdot \mathbf{H}^{{\rm anchor}(l)}$\;
    $\widehat{\mathbf{H}}^{{\rm anchor}(l)} \gets \mathbf{H}^{{\rm anchor}(l)}-\vec{\boldsymbol 1} \cdot \vec{\boldsymbol \mu}^{{\rm anchor}\;{\rm T}}$\;
    $\vec{\boldsymbol \mu}^{{\rm view}\;{\rm T}} \gets \frac{1}{N}\cdot \vec{\boldsymbol 1}^{{\rm T}} \cdot \mathbf{H}^{{\rm view}(l)}$\;
    $\widehat{\mathbf{H}}^{{\rm view}(l)} \gets \mathbf{H}^{{\rm view}(l)}-\vec{\boldsymbol 1} \cdot \vec{\boldsymbol \mu}^{{\rm view}\;{\rm T}}$\;
    $\mathbf{C}_{{\rm cross}}^{(l)} \gets \frac{1}{N}\widehat{\mathbf{H}}^{{\rm anchor}(l){\rm T}}\widehat{\mathbf{H}}^{{\rm view}(l)} + \epsilon \cdot \mathbf{I}$\;
    $\mathcal{L}_{{\rm cross-reg}}^{(l)} \gets \sum_{i}\left(1-\mathbf{C}_{{\rm cross}[i,i]}^{(l)}\right)^{2} + \beta \cdot \sum_{i}\sum_{j \neq i}\mathbf{C}_{{\rm cross}[i,j]}^{(l)2}$\;
    \Return $\mathcal{L}_{{\rm cross-reg}}^{(l)}$\;
\end{algorithm}

\section{Subgraph Statistics}\label{appe:ss}
\begin{figure}[!h]
    \centering
    \includegraphics[scale=0.32]{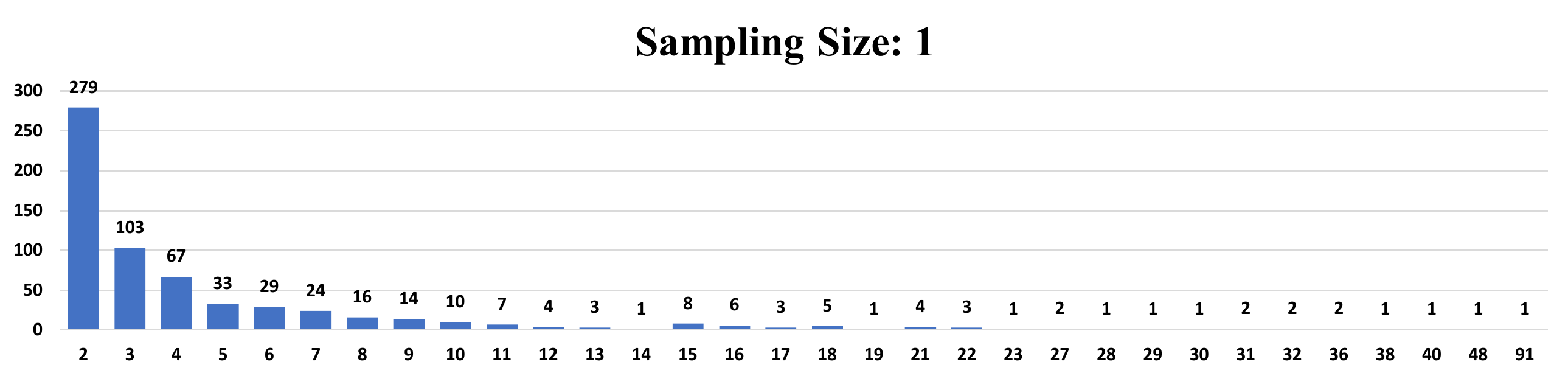}
    \caption{Subgraph statistics on Citeseer by using TAPS strategy to perform the subgraph partition with sampling size 1.}\label{fig:subgraph-citeseer}
\end{figure}
\begin{figure}[!h]
    \centering
    \includegraphics[scale=0.32]{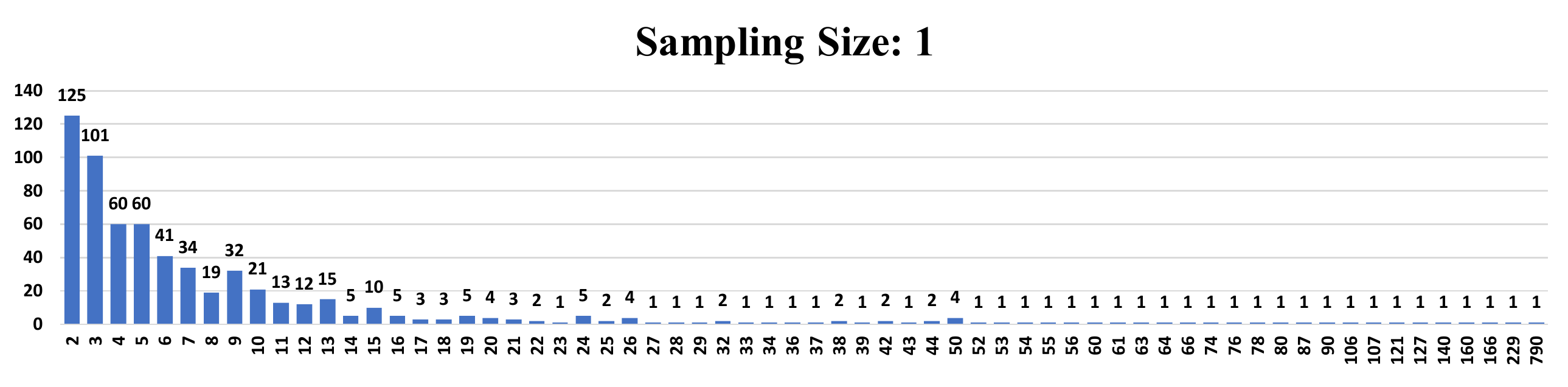}
    \caption{Subgraph statistics on Amazon Photo by using TAPS strategy to perform the subgraph partition with sampling size 1.}\label{fig:subgraph-photo}
\end{figure}
\begin{figure}[!h]
    \centering
    \includegraphics[scale=0.32]{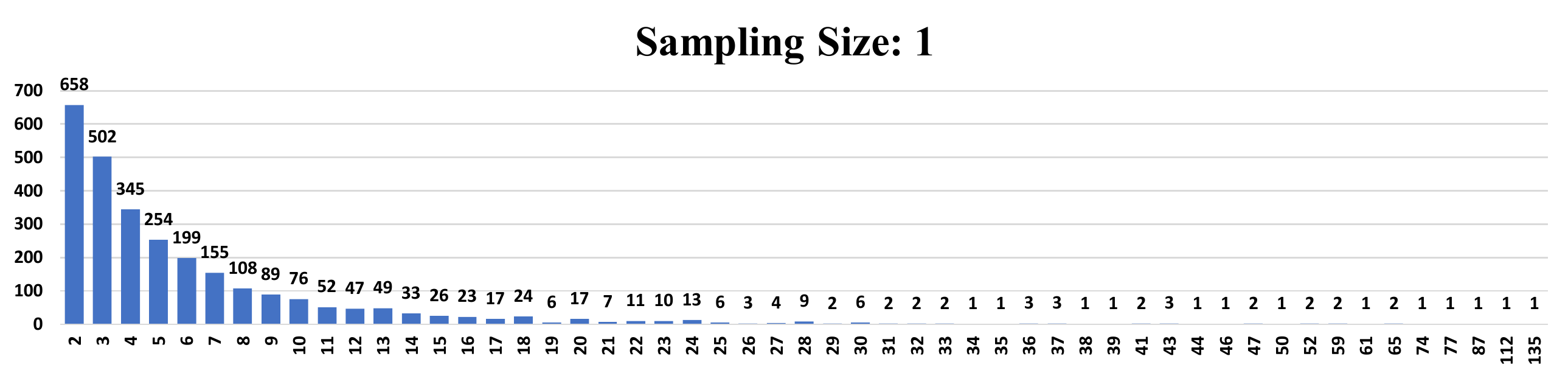}
    \caption{Subgraph statistics on Coauthor CS by using TAPS strategy to perform the subgraph partition with sampling size 1.}\label{fig:subgraph-CS}
\end{figure}
\begin{figure}[!h]
    \centering
    \includegraphics[scale=0.32]{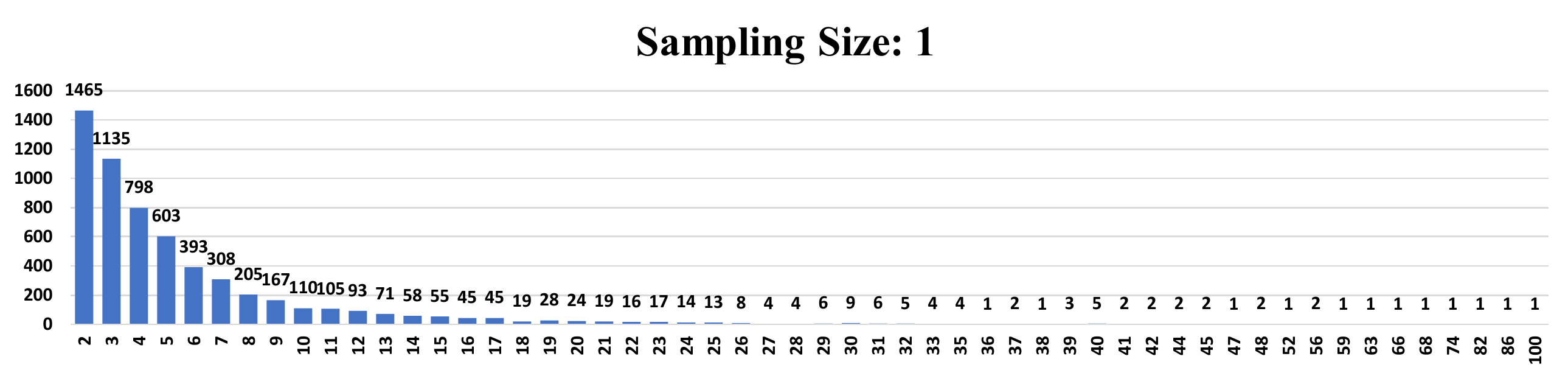}
    \caption{Subgraph statistics on Coauthor Physics by using TAPS strategy to perform the subgraph partition with sampling size 1.}\label{fig:subgraph-phy}
\end{figure}
\begin{figure*}[!t]
    \centering
    \includegraphics[scale=0.34]{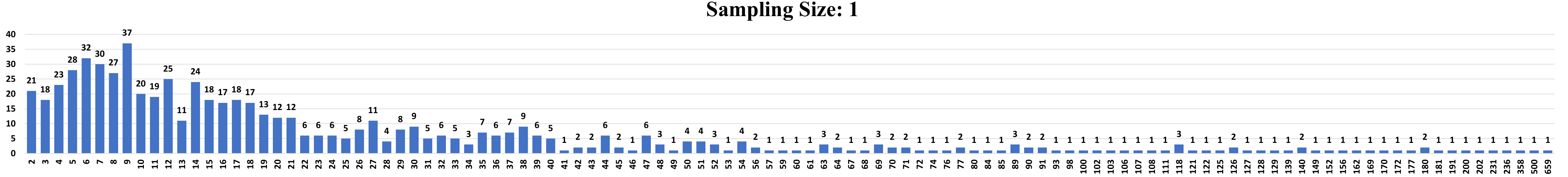}
    \caption{Subgraph statistics on Pubmed by using TAPS strategy to perform the subgraph partition with sampling size 1.}\label{fig:subgraph-pubmed}
\end{figure*}
Subgraph statistics on Citeseer (Fig.~\ref{fig:subgraph-citeseer}), Amazon Photo (Fig.~\ref{fig:subgraph-photo}), Coauthor CS (Fig.~\ref{fig:subgraph-CS}), Coauthor Physics (Fig.~\ref{fig:subgraph-phy}), and Pubmed (Fig.~\ref{fig:subgraph-pubmed}) by using TAPS strategy to perform the subgraph partition with sampling size 1.

\section{Subgraph Partition Visualization}\label{appe:spv}
\begin{figure}[!h]
    \centering
    \includegraphics[scale=0.21]{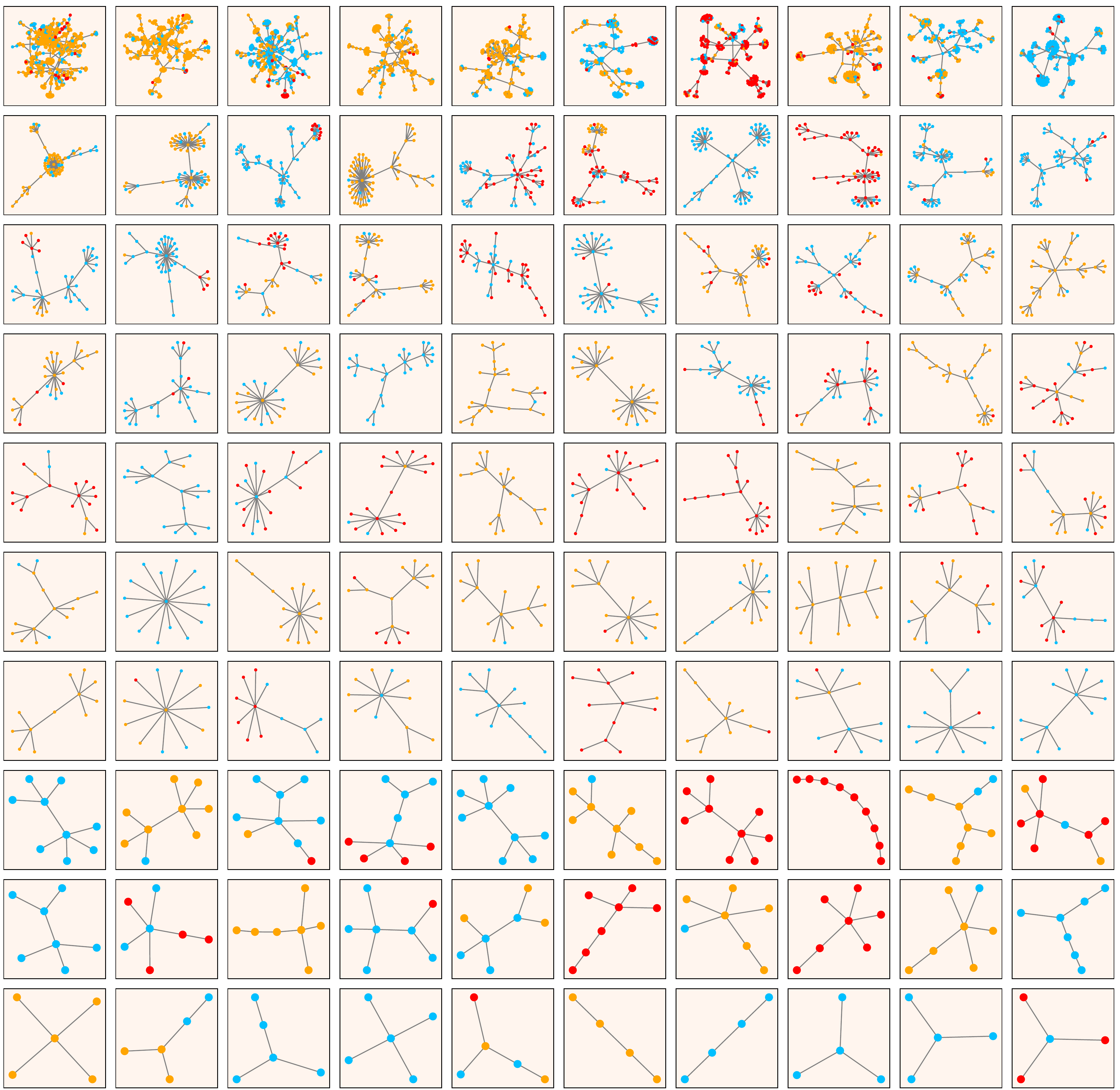}
    \caption{Visualization of part of the subgraphs derived from TAPS with sampling size 1 on Pubmed.}
    \label{fig:subgraph-shown-pubmed}
\end{figure}
\begin{figure}[!h]
    \centering
    \includegraphics[scale=0.21]{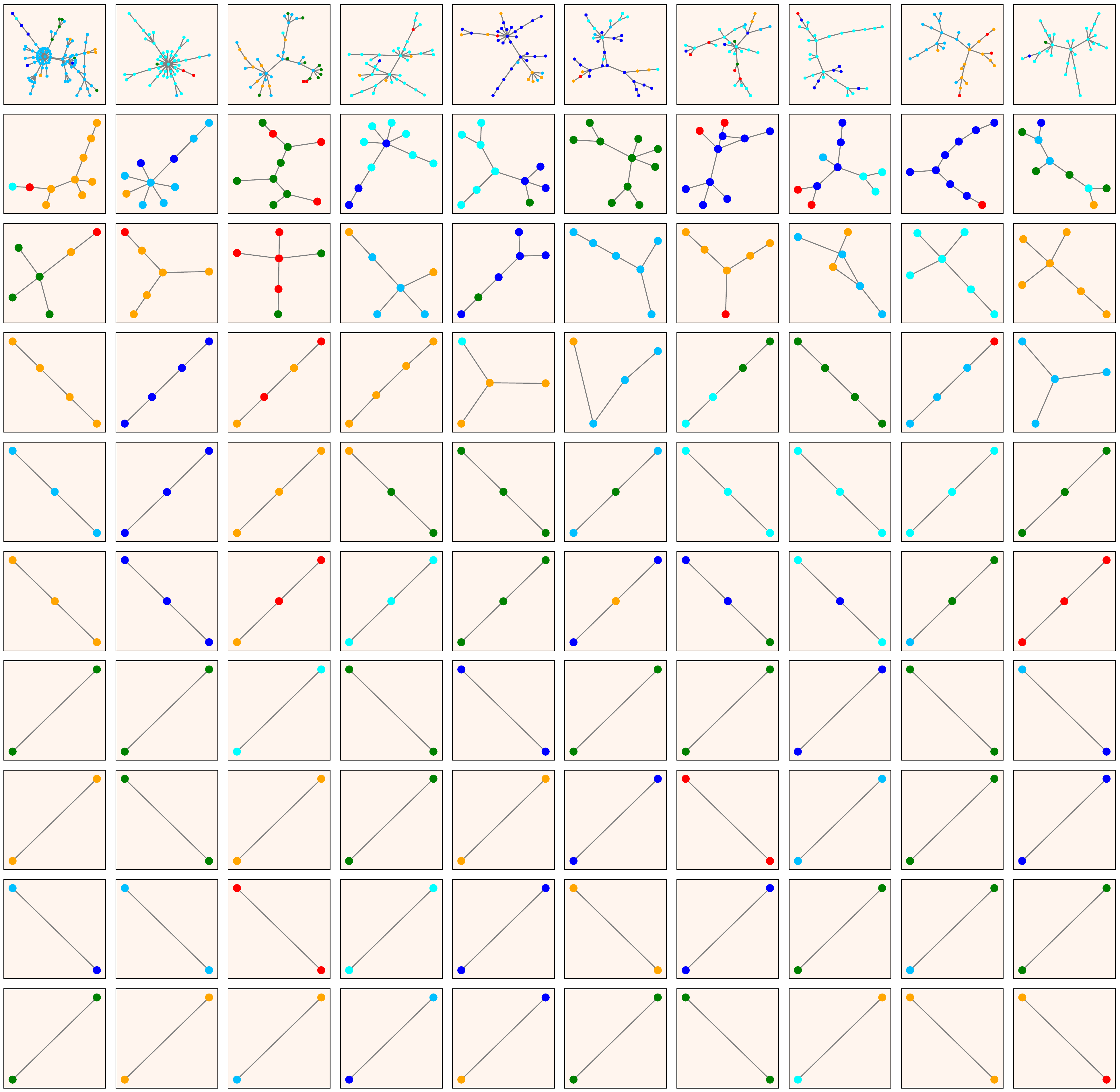}
    \caption{Visualization of part of the subgraphs derived from TAPS with sampling size 1 on Citeseer.}
    \label{fig:subgraph-shown-citeseer}
\end{figure}
\begin{figure}[!h]
    \centering
    \includegraphics[scale=0.21]{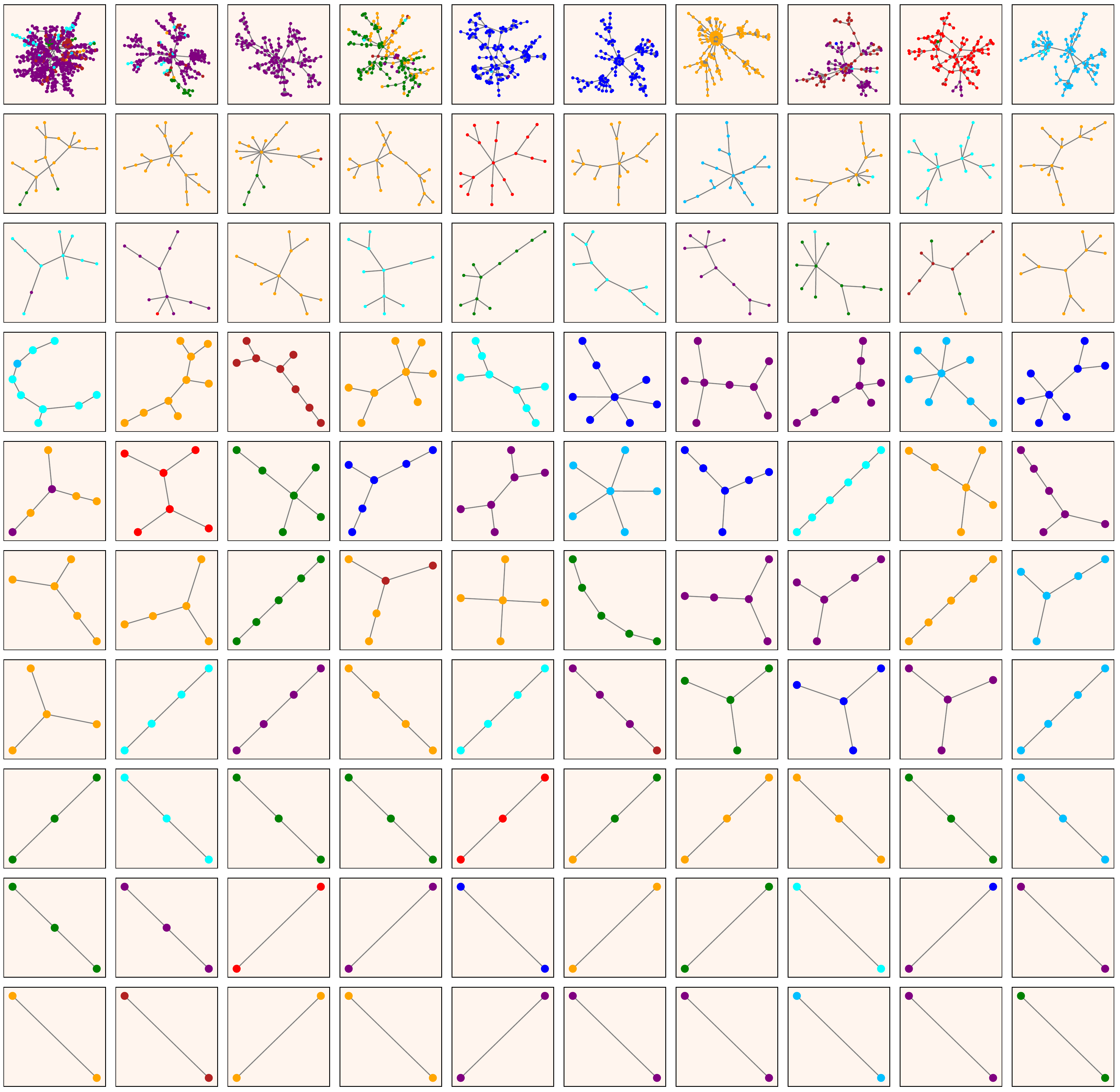}
    \caption{Visualization of part of the subgraphs derived from TAPS with sampling size 1 on Amazon Photo.}
    \label{fig:subgraph-shown-photo}
\end{figure}
\begin{figure}[!h]
    \centering
    \includegraphics[scale=0.21]{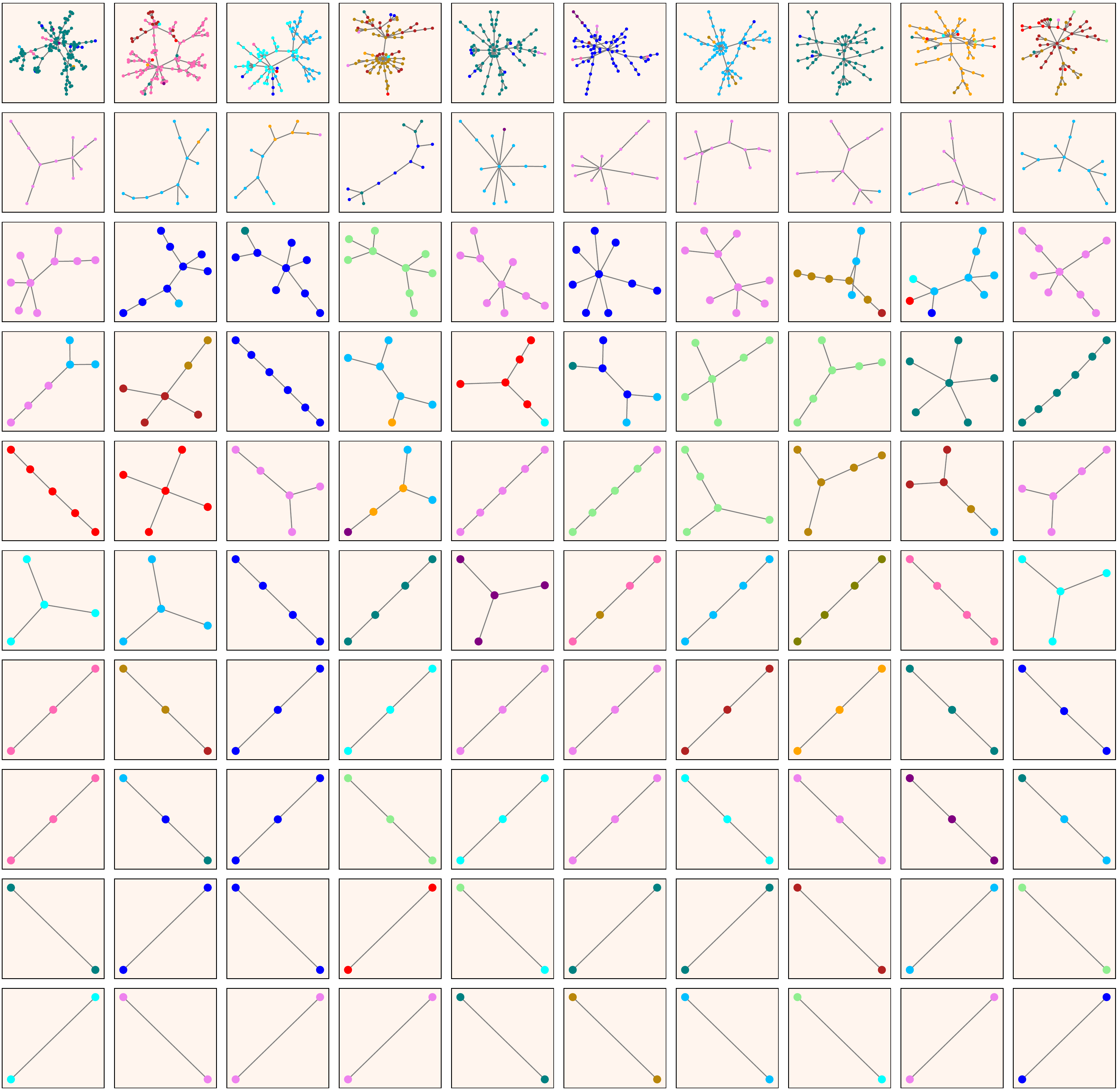}
    \caption{Visualization of part of the subgraphs derived from TAPS with sampling size 1 on Coauthor CS.}
    \label{fig:subgraph-shown-CS}
\end{figure}
\begin{figure}[!h]
    \centering
    \includegraphics[scale=0.21]{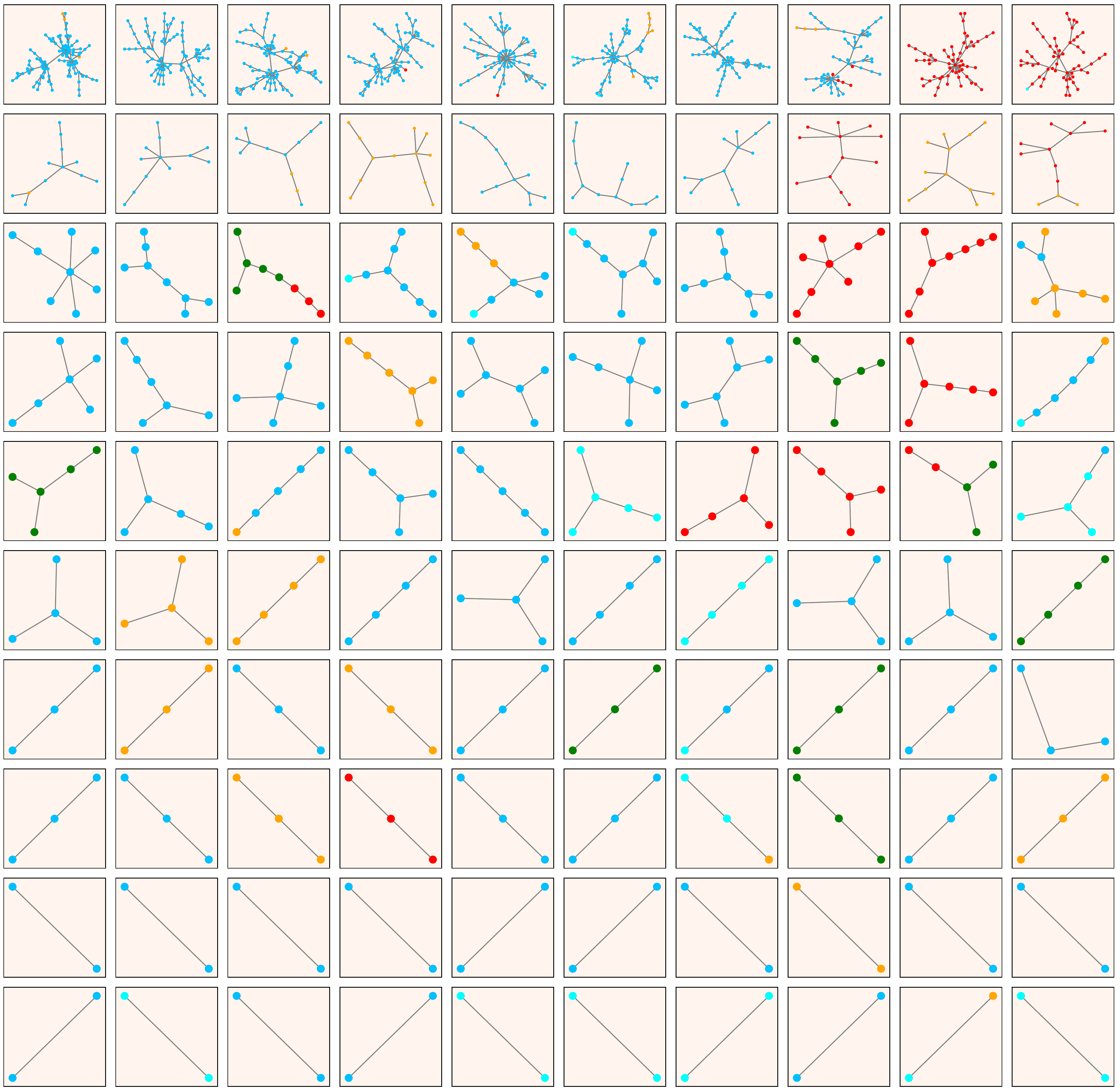}
    \caption{Visualization of part of the subgraphs derived from TAPS with sampling size 1 on Coauthor Physics.}
    \label{fig:subgraph-shown-phy}
\end{figure}
Visualization of part of the subgraphs derived from TAPS with sampling size 1 on Pubmed (Fig.~\ref{fig:subgraph-shown-pubmed}), Citeseer (Fig.~\ref{fig:subgraph-shown-citeseer}), Amazon Photo (Fig.~\ref{fig:subgraph-shown-photo}), Coauthor CS (Fig.~\ref{fig:subgraph-shown-CS}), and Coauthor Physics (Fig.~\ref{fig:subgraph-shown-phy}).


\ifCLASSOPTIONcaptionsoff
  \newpage
\fi



\end{document}